\documentclass[journal]{IEEEtran}
\usepackage{graphicx}
\usepackage{subfigure}
\usepackage{amsmath}
\usepackage{amssymb}
\usepackage{tabularx}
\usepackage{color}
\usepackage{bm}
\usepackage{algorithm}
\usepackage{algorithmic}
\usepackage{multirow}
\usepackage{subfigure}
\usepackage{amsthm}
\usepackage{booktabs}
\usepackage{graphicx}
\usepackage{subeqnarray}
\usepackage{cases}
\usepackage{mathrsfs}
\ifCLASSINFOpdf
\else
\fi
\begin{document}
%
\title{{Task-Oriented Convex Bilevel Optimization with Latent Feasibility}}
%
%
%

	\author{Risheng Liu,~\IEEEmembership{Member,~IEEE,}
	Long Ma,
	Xiaoming Yuan,
	Shangzhi Zeng,
	and Jin Zhang
	\thanks{This work is partially supported by the National Key R\&D Program of China (2020YFB1313503), the National Natural Science Foundation of China (Nos. 61922019 and 11971220), Shenzhen Science and Technology Program (No. RCYX20200714114700072), the Stable Support Plan Program of Shenzhen Natural Science Fund (No. 20200925152128002), Guangdong Basic and Applied Basic Research Foundation (2019A1515011152), the Pacific Institute for the Mathematical Sciences (PIMS), and the Fundamental Research Funds for the Central Universities.}
	\IEEEcompsocitemizethanks{\IEEEcompsocthanksitem R. Liu is with DUT-RU International School of Information Science \& Engineering, Dalian University of Technology, Dalian, 116024, China. E-mail: rsliu@dlut.edu.cn.\protect\
		\IEEEcompsocthanksitem L. Ma is with the School of Software Technology, Dalian University of Technology, Dalian, 116024, China. He is also with the Peng Cheng Laboratory, Shenzhen, 518052, China. E-mail: malone94319@gmail.com. \protect\
		\IEEEcompsocthanksitem X. Yuan is with the Department of Mathematics, The University of Hong Kong, Hong Kong. E-mail: xmyuan@hku.hk. \protect\
		\IEEEcompsocthanksitem S. Zeng is with the Department of Mathematics and Statistics, University of Victoria, Canada. E-mail: zengshangzhi@uvic.ca.
		\IEEEcompsocthanksitem J. Zhang is with the Department of Mathematics, SUSTech International Center for Mathematics, Southern University of Science and Technology, National Center for Applied Mathematics Shenzhen, Shenzhen, 518055, China. ({Corresponding author}, E-mail: zhangj9@sustech.edu.cn.)  \protect\\
	}
	\thanks{Manuscript received April 19, 2005; revised August 26, 2015.}}

%
%

\newtheorem{thm}{Theorem}
\newtheorem{prop}{Proposition}
\newtheorem{cor}{Corollary}
\newtheorem{assumption}{Assumption}
\newtheorem{lemma}{Lemma}
\newtheorem{defi}{Definition}
\newtheorem{remark}{Remark}
\newtheorem{condition}{Condition}

\markboth{Journal of \LaTeX\ Class Files,~Vol.~14, No.~8, August~2015}%
{Shell \MakeLowercase{\textit{et al.}}: Bare Demo of IEEEtran.cls for IEEE Journals}
%



\maketitle

\begin{abstract}
	This paper firstly proposes a convex bilevel optimization paradigm to formulate and optimize popular learning and vision problems in real-world scenarios. Different from conventional approaches, which directly design their iteration schemes based on given problem formulation, we introduce a task-oriented energy as our latent constraint which integrates richer task information. By explicitly re-characterizing the feasibility, we establish an efficient and flexible algorithmic framework to tackle convex models with both shrunken solution space and powerful auxiliary (based on domain knowledge and data distribution of the task). In theory, we present the convergence analysis of our latent feasibility re-characterization based numerical strategy. We also analyze the stability of the theoretical convergence under computational error perturbation. Extensive numerical experiments are conducted to verify our theoretical findings and evaluate the practical performance of our method on different applications.
\end{abstract}

\begin{IEEEkeywords}
	Convex optimization, latent constraint, global convergence, image processing.
\end{IEEEkeywords}

%
\IEEEpeerreviewmaketitle

\section{Introduction}
\begin{figure}[t]
	\centering
	\begin{tabular}{c}
		\includegraphics[width=0.97\linewidth]{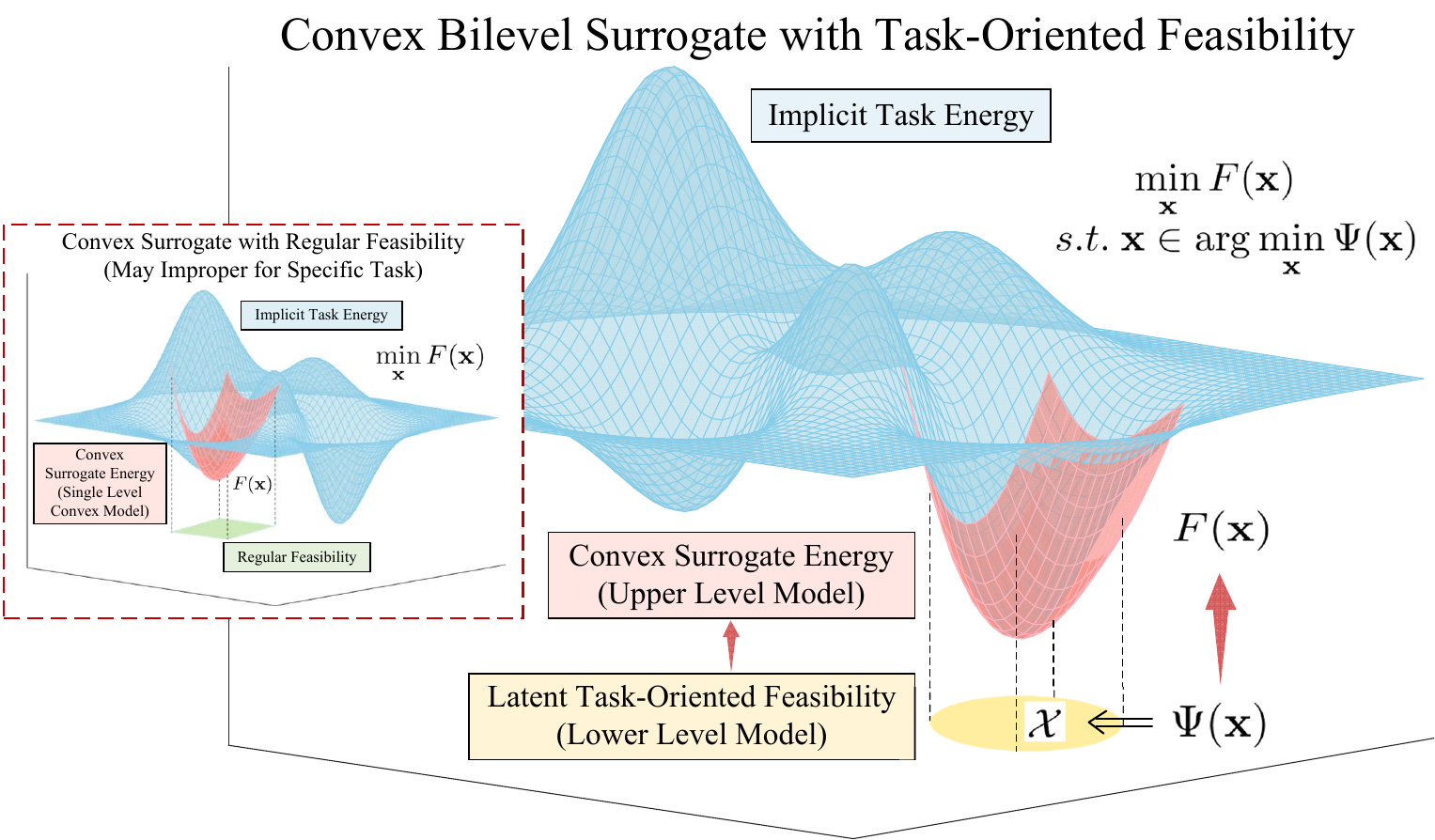}
	\end{tabular}
	\caption{{Illustrating the mechanism of TOLF. The blue and red surfaces represent the implicit nonconvex energy and convex surrogate, respectively. It can be seen that with improper feasibility (e.g., the green region in the red dashed rectangle), standard convex optimization methods (e.g., ~\cite{beck2009fast,bubeck2015convex,li2018structure}) may not obtain the desired solutions for the specific task. In contrast, TOLF aims to introduce latent feasibility to collect rich task information to narrow down the solution space (illustrated as the yellow region) and then solve a convex bilevel formulation to obtain the task-specific optimal solution.}}
	\label{fig:Motivation}
\end{figure}

Over the past decades, convex optimization techniques have been widely used to address machine learning and computer vision problems~\cite{beck2009fast,Li2013An,peng2018structured,huang2020partially}. The main idea behind these approaches is to approximate the implicit task energy (possibly non-convex) by a convex surrogate. Then numerical solvers can be adopted to obtain desirable global solutions. However, due to the complexity of tasks and data distributions, it is usually challenging to exactly obtain the task-desired optimal solutions only based on these simple convex optimization formulations. For example, as in Fig.~\ref{fig:Motivation} (the red dashed rectangle), a convex optimization model with improper feasibility may directly lead to incorrect solutions for the given task. 

Non-convex optimization techniques are usually suggested to encode complex prior information for the task solution. However, in theory, finding global minimizers to non-convex problems is too ambitious. Even worse, according to prevailing non-convex optimization theories, iterative sequences generated by non-convex optimization solvers converge to superficial critical points, which might be saddle points or local maximizers~\cite{bolte2014proximal}. In recent years, a variety of plug-and-play iterative modules have been introduced to perform task-specific optimization. The idea is to unroll an existing optimization process and replace the explicit iterative updating rule with hand-designed operators and/or learned architectures~\cite{Zhang2017Learning,liu2019convergence,liu2020investigating}. Unfortunately, due to the uncontrolled inexact computational modules, it is hard to theoretically guarantee the convergence of these methods. The work in~\cite{Liu2019TGDOF} tried to introduce error control rules to correct improper modules and thus ensure the convergence of their trained iterations.
However, these additional error checking process will slow down the particular computation when handling challenging problems. Moreover, solid theoretical investigations (e.g., the stability of the iterations) are still missing.

In this work, we shall propose a new framework, named Task-Oriented Latent Feasibility (TOLF), to construct convex bilevel optimization models with the support of task information (e.g., principled knowledge and training data) for complex learning and vision problems. Specifically, as illustrated in Fig.~\ref{fig:Motivation}, instead of directly optimizing the convex surrogate on the entire feature space or with explicitly defined constraints, we introduce a task-oriented energy (e.g., the lower-level model in Fig.~\ref{fig:Motivation}) to narrow down the solution space. The convex objective (e.g., the upper-level model in Fig.~\ref{fig:Motivation}) is then optimized upon the optimal solution set of the lower-level model\footnote{Since we always consider the convex (but not strongly convex) lower-level energy, the feasibly solution set of our model may not be a singleton.}, resulting in a convex bilevel optimization problem~\cite{dempe2015bilevel}. Along this direction, we actually suggest a convex optimization model with hierarchies. Two subproblems with hierarchical structures are incorporated to simultaneously formulate the objective and constraint of the given tasks (possibly with non-convex implicit energy).
Since in TOLF we consider general convex composite (not necessarily smooth) energies in both upper and lower subproblems, to our best knowledge, no efficient bilevel optimization algorithms can handle the resulted model.
Fortunately, by re-characterizing the latent feasibility as explicit set constraints, an efficient bilevel iteration scheme is developed to tackle TOLF. 
Theoretically, we prove that our iteration scheme can strictly converge to the global optimal solution of the established convex bilevel model. We also show that the computational errors caused by lower-level solution set re-characterization can be successfully dominated, and thus the proposed method is sufficiently stable.
Extensive experiments on real-world image processing task demonstrate the efficiency and effectiveness of the proposed framework.
In summary, our contributions mainly include:
\begin{itemize}
	\item TOLF provides a flexible framework from a new bilevel perspective. In particular, TOLF incorporates task information to narrow down the solution space and improve the iteration behaviors, leading to a powerful convex bilevel scheme for challenging problems (possibly with unknown nonconvex energies, illustrated in Fig.~\ref{fig:Motivation}). 
	\item By investigating the convergence as well as its stability, our latent feasibility re-characterization based solution strategy for solving nested energy model in TOLF is strictly convinced. Thus we obtain solid theoretical guarantees for the proposed task-oriented convex bilevel optimization paradigm.
	\item 
	We develop a practical method based on TOLF to exploit both the domain knowledge and data-driven engines for real-world applications (e.g., image processing). Our method successfully addresses the issues in standard numerical schemes (hard to leverage data information) and plug-and-play iterations (lack of theoretical guarantees).
\end{itemize}

\section{Related Works}
In this section, we briefly review popular convex, non-convex and plug-and-play optimization techniques in learning and vision fields. 

\textbf{Convex Optimization.}
Recognizing or formulating a given problem as a convex optimization model has been a prevailing manner in a wide range of application areas, e.g., LASSO~\cite{tibshirani1996regression}, matrix completion~\cite{candes2009exact}. Algorithms for solving convex optimization models are rich, efficient and reliable. They can be easily embedded in analysis tools or control systems~\cite{boyd2004convex}. More importantly, stationary solutions to convex models are globally optimal with solid theoretical guarantee~\cite{bubeck2015convex}. Unfortunately, the real-world tasks~\cite{zoran2011learning, li2018structure} are with large complexity and diversity. The advantageous convexity is usually missing, thus it is too ambitious to accurately realize the ideal task-desired optimal solution for a complex task in terms of convex surrogates.

\textbf{Non-convex Optimization.} 
To ameliorate better modeling power, various non-convex prior regularizations~\cite{krishnan2009fast,xu2012structure,pan2016l0} have been derived from given tasks. 
Unfortunately, the non-convexity creates difficulties in designing effective solution schemes~\cite{li2015accelerated,jain2017non}. 
A popular workaround is the relaxation from non-convexity to convexity. However, this step involves losses of accuracy, which sometimes is fateful in applications~\cite{negahban2012unified}. Therefore, solving the non-convex optimizations directly surpasses relaxation-based techniques in some senses, and tremendous successes have been witnessed~\cite{chen2015fast}.
However, general non-convex optimization strategies can hardly guarantee desirable solution qualities. 

\textbf{Plug-and-Play Optimization.}
Recently proposed plug-and-play strategies ~\cite{chan2016plug,romano2017little,liu2019convergence} incorporate some task-related computational modules (e.g., handcrafted operations or trained network architecture) into certain optimization procedures. Substituting subproblems in an optimization process with data-driven networks gains popularity among the plug-and-play literature~\cite{Zhang2017Learning,dong2018denoising}. The essential idea behind the plug-and-play schemes is the derivation of learned architectures from optimization models to incorporate data priors. Unfortunately, despite the observable high-quality performance in practice, the lack of theoretical convergence limits its scientific contributions. 
	To address this issue, the work in~\cite{liu2019convergence} has presented the convergence analysis after introducing an error control rule with a loop body, but this step brings extra computational loads and the final theoretical results which just reach the critical points are still not fully satisfied. In~\cite{Ryu2019ICML}, by designing spectral normalization to train a denoiser, the theoretical convergence analysis of the plug-and-play method is explored based on the forward-backward splitting algorithm. 
	{Moreover, a series of works~\cite{sreehari2016plug, teodoro2018convergent, sun2019online,sun2021scalable} proved the convergence under a set of explicit assumptions, e.g., strongly convex, Lipschitz continuous gradient. There was another approach to prove the convergence which introduced the additional assumptions on the plug-and-play denoisers (e.g., non-expansive,  symmetric)~\cite{reehorst2018regularization,nair2021fixed,cohen2020regularization,gavaskar2021plug,xu2020provable}.
	 Different from them, our TOLF is established on a new bilevel optimization paradigm, which allows us to narrow down the solution space by regarding task information as the lower-level subproblem, rather than focusing on defining the regularization like works mentioned. On the other hand, the iterative process of TOLF we designed is derived based on re-characterization of the latent feasibility, which can bring solid theoretical guarantees and don't need for any additional assumptions. More importantly, our work is flexible enough with a newly-designed bilevel model and we demonstrates the superiority in multiple challenging image-processing applications (e.g., low-light image enhancement) but some have not been completed in other works mentioned.
}

\section{Task-Oriented Convex Bilevel Optimization}
Similar to standard convex optimization paradigms, we also consider the following composite optimization model:
\begin{equation}
\min\limits_{\mathbf{x}}F(\mathbf{x}):=f(\mathbf{x})+g(\mathbf{x}), \label{eq:upper}
\end{equation}	
where both $f$, $g$ are extended-valued convex functions $\mathbb{R}^n\to(-\infty,\infty]$ and $g$ is possibly nonsmooth. But instead of directly optimizing Eq.~\eqref{eq:upper} on the entire feature space or enforcing explicit constraints in existing approaches~\cite{beck2009fast}, we aim to establish a new method, named Task-Oriented Latent Feasibility (TOLF), to collect task information to assist in solving Eq.~\eqref{eq:upper} in challenging real-world scenarios. 

\subsection{Energy-based Latent Feasibility}
The latent feasible set of Eq.~\eqref{eq:upper} can be formulated as the minimizers of another optimization model with the task-oriented composite energy:
\begin{equation}
\min\limits_{\mathbf{x}}\Psi(\mathbf{x})=\psi(\mathbf{x}) + \varphi(\mathbf{x}),\label{eq:lower}
\end{equation}
where $\psi$, $\varphi$ are also extended-valued convex functions $\mathbb{R}^n\to(-\infty,\infty]$ and $\varphi$ is possibly nonsmooth. We will demonstrate in Sec.~\ref{sec:app} how to define $\psi$ and $\varphi$ based on the principled knowledge and/or collected training data for complex tasks in learning and vision communities.
After this process, we amount to solving the following convex bilevel optimization model (i.e., TOLF):
\begin{equation}
\min\limits_{\mathbf{x}}F(\mathbf{x}) \;\; s.t. \ \mathbf{x}
\in \arg\min\limits_{\mathbf{x}}\Psi(\mathbf{x}),\label{eq:bp}
\end{equation} 
which implicitly integrate two different hierarchies of task information (i.e., $F$ and $\Psi$).
From an optimization perspective, we actually utilize the lower-level subproblem Eq.~\eqref{eq:lower} to characterize the feasible region of Eq.~\eqref{eq:bp}. The main benefit of such strategy is that we can take full advantage of our domain knowledge of the task. Indeed, Eq.~\eqref{eq:bp} can be recognized as a specific convex bilevel model. However, both upper and lower levels are in general in lack of smoothness and strong convexity, and thus
the existing solution schemes~\cite{solodov2007explicit,beck2014first,sabach2017first} no longer admit any theoretical validity. Specifically, when the upper-level subproblem is in the absence of strong convexity, directly solving Eq.~\eqref{eq:bp} with such latent feasibility is extremely challenging.

\subsection{Feasibility Re-characterization and Optimization}
Motivated by the observation that in applications, the latent feasibility in Eq.~\eqref{eq:lower} usually possesses some underlying structures, in this paper, we develop a new optimization strategy with solid theoretical guarantees. In particular, we re-characterize the latent feasibility in Eq.~\eqref{eq:lower}, and by doing this, we shall reformulate Eq.~\eqref{eq:bp} in terms of the re-characterization of Eq.~\eqref{eq:lower} into a standard optimization problem which is numerically tractable. To this end,
we first list some structural assumptions, which are necessary for our following analysis and are easy to be satisfied in applications of practical interests. Specifically, throughout this paper, we suppose that $\psi$ 
has the following structural properties.\footnote{Some commonly used functions in learning and vision (e.g., linear regression $h(\mathbf{z})=\frac{1}{2}\|\mathbf{z} - \mathbf{b}_1\|^2$, logistic regression $h(\mathbf{z})=\sum_{i=1}^m \log(1 + e^{\mathbf{z}_i})-\langle \mathbf{b}_2, \mathbf{z}\rangle$ and likelihood estimation under Poisson noise $h(\mathbf{z}) =-\sum_{i=1}^m \log(\mathbf{z}_i)+ \langle \mathbf{b}_3, \mathbf{z}\rangle$) automatically satisfy these assumptions, where $\mathbf{b}_1$, $\mathbf{b}_2$ and $\mathbf{b}_3$ are parameters.}
\begin{assumption}\label{strongconASS}
	$\psi(\mathbf{x})=h(\mathcal{A}(\mathbf{x}))$, where $\mathcal{A}$ is some given linear operator and function $h$ is closed, proper, convex and admits the properties that (i) $h$ is continuously differentiable on $\mathtt{dom}h$, assumed to be open, and (ii) $h$ is local strongly convex on  $\mathtt{dom}h$. 
\end{assumption}

We are now ready to state the following theorem to investigate the feasibility of our problem.
\begin{thm}\label{pro:solution}({\rm Latent feasibility re-characterization})
	Let $\mathcal{X}$ be the solution set of Eq.~\eqref{eq:lower} (i.e., $\mathcal{X}:=\arg\min_{\mathbf{x}}\Psi(\mathbf{x})$), then $\mathcal{A}$ is invariant on $\mathcal{X}$. That is, given any $\bar{\mathbf{x}}\in \mathcal{X}$, $\mathcal{X}$ can be explicitly characterized as
	\begin{equation}\label{rechare}
	\mathcal{X}=\left\{\mathbf{x}| \mathcal{A}(\mathbf{x})=\mathcal{A}(\bar{\mathbf{x}}), \varphi(\mathbf{x}) \leq \varphi(\bar{\mathbf{x}}) \right\}.
	\end{equation}
\end{thm}


Following Theorem~\ref{pro:solution}, we define $\mathcal{X}_{\varphi}:=\{\mathbf{x}|\varphi(\mathbf{x}) \leq \varphi(\bar{\mathbf{x}})\}$ and $\bar{\mathbf{y}}=\mathcal{A}(\bar{\mathbf{x}})$. Then the original bilevel model in Eq.~\eqref{eq:bp} can be equivalently reformulated as the following single-level constrained optimization problem:
\begin{equation}
\min\limits_{\mathbf{x}} f(\mathbf{x}) + g(\mathbf{x}) + \iota_{\mathcal{X}_{\varphi}}(\mathbf{x}), \ s.t. \ \mathcal{A}(\mathbf{x}) = \bar{\mathbf{y}},\label{eq:bp-linear}
\end{equation}
where $\iota_{\mathcal{X}_{\varphi}}$ denotes the indicator of $\mathcal{X}_{\varphi}$. Now our model (i.e., Eq.~\eqref{eq:bp}) is reformulated to the single level standard optimization Eq.~\eqref{eq:bp-linear}. To solve the single level reformulation, we introduce the following augmented Lagrangian function with auxiliary variables $\mathbf{s}$, $\mathbf{z}$
$$
\begin{array}{l}
\mathcal{L}_\beta(\mathbf{x},\mathbf{z},\mathbf{s},\{\bm{\lambda}_i\}_{i=1}^3):=f(\mathbf{x}) + g(\mathbf{z})+ \iota_{\mathcal{X}_{\varphi}}(\mathbf{s})\\
\qquad\quad + \langle \bm{\lambda}_1,\mathbf{z}-\mathbf{x}\rangle + \langle \bm{\lambda}_2,\mathbf{s}-\mathbf{x}\rangle + \langle \bm{\lambda}_3,\mathcal{A}(\mathbf{x})-\bar{\mathbf{y}}\rangle\\
\qquad\quad + \frac{\beta}{2}(\|\mathbf{z}-\mathbf{x}\|^2 +\|\mathbf{s}-\mathbf{x}\|^2+ \|\mathcal{A}(\mathbf{x})-\bar{\mathbf{y}}\|^2),
\end{array}$$
where $\{\bm{\lambda}_i\}_{i=1}^3$ denote the dual multipliers and $\beta> 0$ denotes the penalty parameter. Then the proximal ADMM (with $\tau > 0$) reads as follows:
\begin{equation}\label{eq:admm}
\left\{\begin{array}{l}
\mathbf{z}^{k+1}\in\arg\min\limits_{\mathbf{z}}\mathcal{L}_\beta(\mathbf{x}^k,\mathbf{z},\mathbf{s}^k,\{\bm{\lambda}_i^k\}_{i=1}^3) \\\qquad\quad + \frac{\tau}{2}\|\mathbf{z}-\mathbf{z}^k\|^2,\\
\mathbf{s}^{k+1}\in\arg\min\limits_{\mathbf{s}}\mathcal{L}_\beta(\mathbf{x}^k,\mathbf{z}^k,\mathbf{s},\{\bm{\lambda}_i^k\}_{i=1}^3) \\\qquad\quad + \frac{\tau}{2}\|\mathbf{s}-\mathbf{s}^k\|^2,\\
\mathbf{x}^{k+1}\in\arg\min\limits_{\mathbf{x}}\mathcal{L}_\beta(\mathbf{x},\mathbf{z}^{k+1},\mathbf{s}^{k+1},\{\bm{\lambda}_i^k\}_{i=1}^3)\\\qquad\quad + \frac{\tau}{2}\|\mathbf{x}-\mathbf{x}^k\|^2,\\		
\bm{\lambda}_1^{k+1}=\bm{\lambda}_1^{k} + \beta(\mathbf{z}^{k+1}-\mathbf{x}^{k+1}),\\
\bm{\lambda}_2^{k+1}=\bm{\lambda}_2^{k} + \beta(\mathbf{s}^{k+1}-\mathbf{x}^{k+1}),\\
\bm{\lambda}_3^{k+1}=\bm{\lambda}_3^{k} + \beta(\mathcal{A}(\mathbf{x}^{k+1})-\bar{\mathbf{y}}).
\end{array}\right.
\end{equation}

Thanks to Theorem~\ref{pro:solution}, we directly have a corollary to guarantee the convergence of Eq.~\eqref{eq:admm} toward the global optimal solutions of Eq.~\eqref{eq:bp}.

\begin{cor}
	Suppose that the problem in Eq.~\eqref{eq:bp-linear} has KKT solutions.
	Let $\{(\mathbf{x}^k,\mathbf{z}^k,\mathbf{s}^k,\bm{\lambda}^k)\}$ be the sequence generated by Eq.~\eqref{eq:admm} on problem \eqref{eq:bp-linear}, then $\{(\mathbf{x}^k,\mathbf{z}^k,\mathbf{s}^k,\bm{\lambda}^k)\}$ converges to the KKT point set of Eq.~\eqref{eq:bp-linear}. In particular, the sequence $\{\mathbf{x}^k\}$ converges to the global optimal solutions of Eq.~\eqref{eq:bp}.
\end{cor}
\begin{remark}
	Indeed, we can further estimate a nice linear convergence rate of Eq.~\eqref{eq:admm} for particular models. That is, if $f$ takes the form that $f(\mathbf{x}) = h(\tilde{\mathcal{A}} \mathbf{x})$ where $\tilde{\mathcal{A}}$ is some given linear operator, $h$ satisfies Assumption \ref{strongconASS}, $g$ represents (1) convex polyhedral regularizer; (2) group-lasso regularizer; (3) sparse group lasso regularizer, and $\psi$ is a convex polyhedral function, then $\{(\mathbf{x}^k,\mathbf{z}^k,\mathbf{s}^k,\bm{\lambda}^k)\}$ converges linearly to the KKT point set of the problem in Eq.~\eqref{eq:bp-linear}. In particular, the sequence $\{\mathbf{x}^k\}$ converges linearly to the global optimal solutions of Eq.~\eqref{eq:bp}.
\end{remark}

\section{Theoretical Investigations}
Thanks to Theorem \ref{pro:solution}, the solutions returned by Eq.~\eqref{eq:bp-linear} can exactly optimize the bilevel problem in Eq.~\eqref{eq:bp}. Our single-level reformulation based optimization scheme relies on the re-characterization of the solution set. To be specific, we require one solution of the constraint subproblem (i.e., $\bar{\mathbf{x}} \in \arg\min_{\mathbf{x}}\psi(\mathbf{x}) + \varphi(\mathbf{x})$) to construct the solution set $\mathcal{X}$.

In general, obtaining such a solution exactly is intractable. That is, we in practice can only calculate a solution with computation errors for the constraint subproblem in Eq.~\eqref{eq:lower}, i.e., obtain a point $\bar{\mathbf{x}}_{\delta}$ satisfying $d(\bar{\mathbf{x}}_{\delta},\mathcal{X}) \leq \delta$, where $d$ is the distance mapping, and $\delta\geq 0$ measures the computational errors.
As a consequence, we consider the practical optimization process of Eq.~\eqref{eq:bp} as solving an approximation of Eq.~\eqref{eq:bp-linear}, which can be formulated as follows\footnote{Please notice that Eq.~\eqref{appro_prob} is only used for our theoretical analysis, but not practical computation.}:
\begin{equation}\label{appro_prob}
\min\limits_{\mathbf{x}} F(\mathbf{x}), \ s.t. \ \mathcal{A}(\mathbf{x}) = \mathcal{A}(\bar{\mathbf{x}}_{\delta}), \ \varphi(\mathbf{x})\leq \varphi(\bar{\mathbf{x}}_{\delta}).
\end{equation}

In the following, we shall analyze the convergence behaviors and stability properties of our practical computation (can be abstractly formulated as Eq.~\eqref{appro_prob}) from the perturbation analysis perspective. Specifically, we consider the errors for solving $\bar{\mathbf{x}}$ as the perturbation of optimizing Eq.~\eqref{eq:bp-linear} and obtain the following constructive results:
\begin{itemize}
	\item Convergence (Theorem~\ref{th1}): As the error $\delta$ decreases to $0$ in Eq.~\eqref{appro_prob}, the solution sequence strictly converges to our desired solution of the bilevel problem in Eq.~\eqref{eq:bp}.
	\item Stability (Theorem~\ref{thm:stability}): The proximity from the optimal solution of Eq.~\eqref{appro_prob} to the solution set of the bilevel problem in Eq.~\eqref{eq:bp} can be strictly dominated in terms of $\delta$.
\end{itemize}

{
	\subsection{Convergence Analysis}
	Before proving our formal convergence result, we first introduce some necessary notations. By respectively considering $\mathcal{A}(\bar{\mathbf{x}}_{\delta})$ and $\varphi(\bar{\mathbf{x}}_{\delta})$ in Eq.~\eqref{appro_prob} as perturbed $\bar{\mathbf{y}}$ and $\bar{\mathbf{s}}$, we are now aiming to investigate the stability of the following parameterized optimization problem
	\begin{equation}
	(P_{\mathbf{p}}) \quad \min\limits_{{\mathbf{x}}} F(\mathbf{x}), \ s.t. \ \left\{\begin{array}{l}
	\mathcal{A}(\mathbf{x}) - \bar{\mathbf{y}} = \mathbf{p}_1, \\
	\varphi(\mathbf{x})\leq \bar{\mathbf{s}}+\mathbf{p}_2,
	\end{array} \right.
	\end{equation}
	where $\mathbf{p}=\{\mathbf{p}_1,\mathbf{p}_2\}$, $\bar{\mathbf{y}} = \mathcal{A}(\bar{\mathbf{x}})$ and $\bar{\mathbf{s}} = \varphi(\bar{\mathbf{x}})$ for any given $\bar{\mathbf{x}} \in \mathcal{X}$. Moreover, we shall need the following notations.
	\begin{itemize}
		\item The feasible set mapping of $P_{\mathbf{p}}$: $\mathcal{S}_{feas}(\mathbf{p}):= \{\mathbf{x} | \mathcal{A}(\mathbf{x}) - \bar{\mathbf{y}} = \mathbf{p}_1, \varphi(\mathbf{x}) \le \bar{\mathbf{s}} + \mathbf{p}_2\}$.
		\item The optimal value mapping of $P_{\mathbf{p}}$: $\mathcal{S}_{val}(\mathbf{p}):= \inf_{\mathbf{x}}\{ F(\mathbf{x}) ~|~ \mathcal{A}(\mathbf{x}) - \bar{\mathbf{y}} = \mathbf{p}_1, ~ \varphi(\mathbf{x}) \le \bar{\mathbf{s}} + \mathbf{p}_2\}$.
		\item The solution set mapping of $P_{\mathbf{p}}$: $\mathcal{S}_{sol}(\mathbf{p}) := \{ \mathbf{x} \in \mathcal{S}_{feas}(\mathbf{p})~|~ F(\mathbf{x}) = \mathcal{S}_{val}(\mathbf{p})\}$.
	\end{itemize}
	
	Continuity properties of set-valued mapping $\mathcal{S} : \mathbb{R}^m \rightrightarrows \mathbb{R}^n$ is developed in terms of outer and inner limits:
	{\small \[
		\begin{aligned}
		\limsup_{\mathbf{p} \rightarrow \bar{\mathbf{p}}}\mathcal{S}(\mathbf{p}) &:= \{\mathbf{x} ~|~ \exists \mathbf{p}^{\nu} \rightarrow \bar{\mathbf{p}}, \exists \mathbf{x}^{\nu} \rightarrow \mathbf{x} ~\text{with}~ \mathbf{x}^{\nu} \in \mathcal{S}(\mathbf{p}^{\nu}) \}, \\
		\liminf_{\mathbf{p} \rightarrow \bar{\mathbf{p}}}\mathcal{S}(\mathbf{p}) &:= \{\mathbf{x} ~|~ \forall \mathbf{p}^{\nu} \rightarrow \bar{\mathbf{p}}, \exists \mathbf{x}^{\nu} \rightarrow \mathbf{x} ~\text{with}~ \mathbf{x}^{\nu} \in \mathcal{S}(\mathbf{p}^{\nu}) \}. \\
		\end{aligned}
		\] }
	\begin{defi}\label{defnitionconti}
		A set-valued mapping $\mathcal{S} : \mathbb{R}^m \rightrightarrows \mathbb{R}^n$ is outer semicontinuous (OSC) at $\bar{\mathbf{p}}$ when
		$
		\limsup_{\mathbf{p} \rightarrow \bar{\mathbf{p}}}\mathcal{S}(\mathbf{p}) \subseteq \mathcal{S}(\bar{\mathbf{p}})
		$
		and inner semicontinuous (ISC) at $\bar{\mathbf{p}}$ when
		$
		\liminf_{\mathbf{p} \rightarrow \bar{\mathbf{p}}}\mathcal{S}(\mathbf{p}) \supseteq \mathcal{S}(\bar{\mathbf{p}}).
		$
		It is called continuous at $\bar{\mathbf{p}}$ when it is both OSC and ISC at $\bar{\mathbf{p}}$, as expressed by
		$
		\lim_{\mathbf{p} \rightarrow \bar{\mathbf{p}}}\mathcal{S}(\mathbf{p}) = \mathcal{S}(\bar{\mathbf{p}}).
		$
	\end{defi}
	
	\begin{lemma}\label{prop2}
		Suppose that $F$ is a continuous function. If $\mathcal{S}_{feas}(\mathbf{p})$ is continuous at ${0}$ and $\mathcal{S}_{sol}({0}) \neq \varnothing$, then $\mathcal{S}_{sol}(\mathbf{p})$ is outer semicontinuous at ${0}$.
	\end{lemma}
	
	\begin{remark}
		We shall clarify the continuity assumption regarding $\mathcal{S}_{feas}$ in Lemma \ref{prop2}.
		In fact, when $\varphi$ is a convex polyhedral function, $\mathcal{S}_{feas}$ is a closed polyhedral convex mapping. Then according to Theorem 3C.3 in \cite{dontchev2009implicit}, we know that $\mathcal{S}_{feas}$ is Lipschitz continuous, i.e. there exists $\kappa \ge 0$ such that for all $\mathbf{p}_1, \mathbf{p}_2 \in \mathtt{dom}\mathcal{S}_{feas}$,
		\[
		h(\mathcal{S}_{feas}(\mathbf{p}_1),\mathcal{S}_{feas}(\mathbf{p}_2)) \le \kappa\|\mathbf{p}_1-\mathbf{p}_2\|,
		\]
		where for any nonempty sets $\mathcal{E}$ and $\mathcal{F}$, $h(\mathcal{E},\mathcal{F})$ is given by
		$
		h(\mathcal{E},\mathcal{F}) = \max\{e(\mathcal{E},\mathcal{F}),e(\mathcal{F},\mathcal{E})\}$,
		and	
		$e(\mathcal{E},\mathcal{F}) = \sup_{\mathbf{x} \in \mathcal{E}}d(\mathbf{x},\mathcal{F})
		$.
		Therefore, when $\varphi$ is a convex polyhedral function, all the assumptions about $\mathcal{S}_{feas}$ in the lemma above are satisfied.	
	\end{remark}
	
	Now we are ready to induce the main result to guarantee the convergence of our proposed optimization scheme.
	\begin{thm}\label{th1}
		Suppose that $\varphi$ is a convex polyhedral function and let $\{{\mathbf{x}}_{\delta_k}^*\}$ be the solution returned by solving Eq.~\eqref{appro_prob} with errors $\{\delta_k\}$.
		If $\delta_k \rightarrow 0$, then
		\begin{enumerate}
			\item For any accumulation point ${\mathbf{x}}^*$ of the sequence $\{{\mathbf{x}}_{\delta_k}^*\}$, we have that ${\mathbf{x}}^* \in \mathcal{S}_{sol}({0})$. That is, ${\mathbf{x}}^*$ solves bilevel problem Eq.~\eqref{eq:bp}.
			\item If $F$ is coercive, then the sequence $\{{\mathbf{x}}_{\delta_k}^*\}$ is bounded and hence admits at least one accumulation point.
		\end{enumerate}
	\end{thm}
	
	\subsection{Stability Analysis}
	
	Before we establish the desired stability result for Eq.~\eqref{appro_prob}, we need the stability analysis as preliminaries.
	\begin{prop}\label{prop3}
		Suppose that there exists neighborhood $\mathcal{N}$ of some point $\bar{\mathbf{x}} \in \mathcal{S}_{sol}({0})$ such that
		$
		F(\mathbf{x}) \ge F(\bar{\mathbf{x}}) + \frac{c}{2} d(\mathbf{x},\mathcal{S}_{sol}({0}))^2, \ \forall \mathbf{x} \in \mathcal{N} \cap \mathcal{S}_{feas}({0}),
		$
		where $F$ is Lipschitz continuous with modulus $L$ on $\mathcal{N}$, and there exist $\kappa_1, \kappa_2$ such that 
		$
		\mathcal{S}_{feas}(\mathbf{p}) \cap \mathcal{N} \subseteq \mathcal{S}_{feas}({0}) + \kappa_1\|\mathbf{p}\|\mathbb{B},
		$
		and
		$
		\mathcal{S}_{feas}(0) \cap \mathcal{N} \subseteq \mathcal{S}_{feas}(\mathbf{p}) + \kappa_2\|\mathbf{p}\|\mathbb{B}.
		$
		Then for any $\mathbf{x}_p \in \mathcal{S}_{sol}(\mathbf{p}) \cap \mathcal{N}$, we have
		$
		d(\mathbf{x}_p, \mathcal{S}_{sol}({0})) \le \kappa_1\|\mathbf{p}\| + \left(\frac{(\kappa_1+\kappa_2)L}{c}\right)^{\frac{1}{2}}\|\mathbf{p}\|^{\frac{1}{2}}.
		$
	\end{prop}
	
	According to the results obtained above, together with the arguments given in the proof of Theorem \ref{th1}, we also have following stability guarantees as follows.
	\begin{thm}\label{thm:stability}
		For given $\delta > 0$, let $\mathbf{x}^*_{\delta}$ represents the solution returned by solving Eq.~\eqref{appro_prob}.
		Suppose that $\varphi$ is a convex polyhedral function. If there exists a neighborhood $\mathcal{N}$ of some point $\bar{\mathbf{x}} \in \mathcal{S}_{sol}(0)$ such that
		$
		F(\mathbf{x}) \ge F(\bar{\mathbf{x}}) + \frac{c}{2} d(\mathbf{x},\mathcal{S}_{opt}(0))^2, \ \forall \mathbf{x} \in \mathcal{N} \cap \mathcal{S}_{feas}(0),
		$
		and $F$ is Lipschitz continuous on $\mathcal{N}$, then there exist $c_1, c_2 > 0$ such that for any $\mathbf{x}^*_{\delta} \in \mathcal{N}$, we have
		$
		d(\mathbf{x}^*_{\delta}, \mathcal{S}_{sol}(0)) \le c_1\delta + c_2\delta^{1/2}.
		$
	\end{thm}
	
	\section{Applications}\label{sec:app}
This section shows how to apply TOLF to integrate task-oriented information to solve the classical $\ell_1$ regularized convex optimization model for a variety of challenging applications, such as Image Restoration (IR), Compressed Sensing MRI (CS-MRI) and Low-Light Image Enhancement (LLIE). Please notice that the last two tasks actually have never been addressed by such a simple convex model in previous works. 
Specifically, given the observed image $\mathbf{y}$, by representing the latent image as $\mathbf{F}\mathbf{x}$ (i.e., $\mathbf{x}$ represents the sparse code of the latent image on the inverse wavelet transform $\mathbf{F}$), we consider the upper-level objective in Eq.~\eqref{eq:bp} as the following $\ell_1$-regularized energy:
\begin{equation}
\min\limits_{\mathbf{x}}F(\mathbf{x})=\frac{1}{2}\|\mathbf{A}\mathbf{F}\mathbf{x}-\mathbf{y}\|^2+\eta\|\mathbf{x}\|_1,\label{eq:sc}
\end{equation}
where $\mathbf{A}$ denotes a task-specific matrix, and $\eta>0$ is a trade-off parameter. The model in Eq.~\eqref{eq:sc} is actually a classical convex formulation for image restoration~\cite{beck2009fast}, in which we consider $\mathbf{A}$ as the blur matrix.
In Table~\ref{tab:SpecificSettingofA}, we summarize the specific forms of $\mathbf{A}$ for three image processing applications including Image Restoration (IR), Compressed Sensing MRI (CS-MRI), and Low-Light Image Enhancement (LLIE).

\begin{table}[t]
	\centering
	\caption{The specific forms of the linear operator $\mathbf{A}$ in different applications. }
	\begin{tabular}{ccc}
		\toprule
		\footnotesize Application &\footnotesize Linear operator $\mathbf{A}$ &\footnotesize Remarks\\
		\midrule
		\footnotesize IR&\footnotesize $\mathbf{K}$&$\mathbf{K}$ denotes the blur matrix\\
		\midrule
		\multirow{2}{*}{\footnotesize CS-MRI}
		&\multirow{2}{*}{\footnotesize $\mathbf{P}\mathbf{H}$}&\footnotesize $\mathbf{P}$ is the under-sampling matrix \\
		&&\footnotesize $\mathbf{H}$ is the Fourier transform\\
		\midrule
		\footnotesize LLIE &\footnotesize $\mathbf{T}$&\footnotesize $\mathbf{T}$ denotes the illumination matrix\\
		\bottomrule
	\end{tabular}
	\label{tab:SpecificSettingofA}
\end{table}

\subsection{Latent Constraints}
Within TOLF, we introduce an energy-based latent constraint on $\mathbf{x}$  for Eq.~\eqref{eq:sc} as follows:
\begin{equation}\label{eq:lower-app}
\mathbf{x}\in\mathcal{X}:=\arg\min\limits_{\mathbf{x}}\Psi(\mathbf{x})=\frac{1}{2}\|\mathbf{D}(\mathbf{F}\mathbf{x}-\mathbf{u})\|^2 + \gamma\|\mathbf{D}\mathbf{F}\mathbf{x}\|_1,
\end{equation}
where $\mathbf{D}$ denotes the gradient operator, $\mathbf{u}=\mathcal{T}(\mathbf{y})$, and $\gamma$ is a balancing parameter. {As for the first term in Eq. (10), it is introduced to ensure the prominent structural similarity (extracted by the gradient map) between the warm-start $\mathbf{u}$ and the desired output $\mathbf{Fx}$. The second term is to enforce the sparsity of image gradients.}
In fact, $\Psi$ embeds the task information from two different perspectives on the image gradient domain. On the one hand, we utilize $\|\mathbf{D}\mathbf{F}\mathbf{x}\|_1$ to enforce the sparsity of image gradients (i.e., total variation prior). {On the other hand, we incorporate a task-specific operation $\mathcal{T}$\footnote{The specific form and analysis can be found in Sec.~\ref{sec:IterationBehaviors} and~\ref{sec:ParameterAnalysis}.} to generate warm-start to guide the optimization process. }

{
Here, We would like to clarify that we actually have not made such an assumption that the problem in Eq.~\eqref{eq:lower-app} has multiple solutions. Indeed, the multiple solutions property is completely due to the underlying structure of the latent constraint defined in Eq.~\eqref{eq:lower-app}. This also justifies our motivation to study the bilevel optimization paradigm. 
Particularly, the non-emptiness of the latent constraint defined by the solution set of the optimization problem in Eq.~\eqref{eq:lower-app} can be explicitly shown as following. We may first consider an auxiliary problem defined by
\[
\min_{\mathbf{y}} \tilde{\Psi}(\mathbf{y}) =  \frac{1}{2}\|\mathbf{y} - \mathbf{D}\mathbf{u}\|^2 + \gamma \|\mathbf{y}\|_1.
\] 
As $\tilde{\Psi}$ is strongly convex, the above optimization has a unique solution, that is there exists $\bar{\mathbf{y}}$ such that
\[
\bar{\mathbf{y}}=\underset{\mathbf{y}}{\mathrm{argmin}}~\tilde{\Psi}(\mathbf{y}).
\]
Then the solution set of the optimization problem in Eq.~\eqref{eq:lower-app} can be characterized by 
\[
\{ \mathbf{x} ~|~ \mathbf{D} \mathbf{F} \mathbf{x} = \bar{\mathbf{y}}\}=\underset{\mathbf{x}}{\mathrm{argmin}}~\Psi(\mathbf{x}).
\]
According to the definition of $\mathbf{D} $ and $\mathbf{F}$, the linear operator $\mathbf{D} \mathbf{F} $ is not injective, and thus the solution set of the optimization problem in Eq.~\eqref{eq:lower-app} has multiple solutions.
}

{	
	We also make detailed explanation about why using proximal ADMM. 	
	Applying the convergence result of primal variables for vanilla ADMM established in~\cite{eckstein1992douglas} to the optimization problem in Eq.~\eqref{eq:bp-linear} requires the linear operator $\mathcal{A}$ to be injective. However, when $\mathcal{A}$ is injective, the latent feasible set of Eq.~\eqref{eq:upper}  (i.e., the solution set of the lower-level problem in Eq.~\eqref{eq:lower}) is unique, this is not the case that we are interested in. Furthermore, for the latent feasible set in Eq.~\eqref{eq:lower-app}, the linear operator $\mathcal{A}$ is chosen as $\mathbf{D}\mathbf{F}$ and it is not injective.
	The added proximal term can make the subproblems in the proximal ADMM scheme be strongly convex, specially the $\mathbf{x}$-update subproblem. This can help the subproblems be more stable and easier to be solved.  	
}


We emphasize that $\mathcal{T}$ actually implements the mechanism similar to plug-and-play architecture~\cite{chan2016plug,liu2019convergence}, but we only need to calculate it once at the initial stage, and thus it reduces the computational burden than that in existing plug-and-play methods. More importantly, TOLF obtains much better theoretical properties than plug-and-play approaches~\cite{chan2016plug,liu2019convergence,romano2017little} in terms of both convergence and stability.

\begin{table}[t]
	\centering
	\caption{Quantitative results of different optimization mechanisms (i.e., FISTA, PP-ADMM and TOLF) on solving Eq.~\eqref{eq:sc} for the image restoration task.}
	\renewcommand\arraystretch{1.25}
	\begin{tabular}{p{0.95cm}<{\centering}p{0.95cm}<{\centering}p{0.95cm}<{\centering}p{0.95cm}<{\centering}p{0.95cm}<{\centering}p{0.95cm}<{\centering}}
		\toprule
		\multirow{2}*{Method}&	\multirow{2}{*}{\footnotesize FISTA}&\multicolumn{2}{c}{\footnotesize PP-ADMM}&\multicolumn{2}{c}{\footnotesize TOLF}\\
		\cline{3-6}
		~&~&\footnotesize (DC)&\footnotesize (RF)&\footnotesize (DC)&\footnotesize (RF)\\
		\midrule
		\footnotesize 	PSNR&\footnotesize 29.82&\footnotesize 27.69&\footnotesize 30.06&\footnotesize {31.64} &\footnotesize 30.95\\
		\footnotesize SSIM&\footnotesize 0.8818&\footnotesize 0.8037&\footnotesize 0.8735&\footnotesize {0.8997}&\footnotesize 0.8871\\
		\bottomrule
	\end{tabular}
	\label{tab:OptimizationMechanism}
\end{table}

\begin{figure}[t]
	\centering
	\begin{tabular}{c@{\extracolsep{0.4em}}c}
		\includegraphics[width=0.46\linewidth]{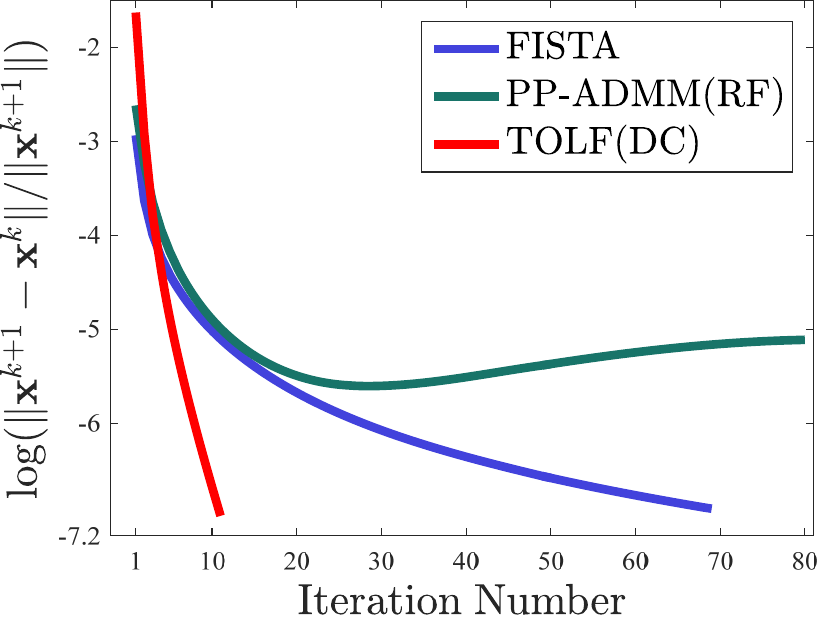}&
		\includegraphics[width=0.46\linewidth]{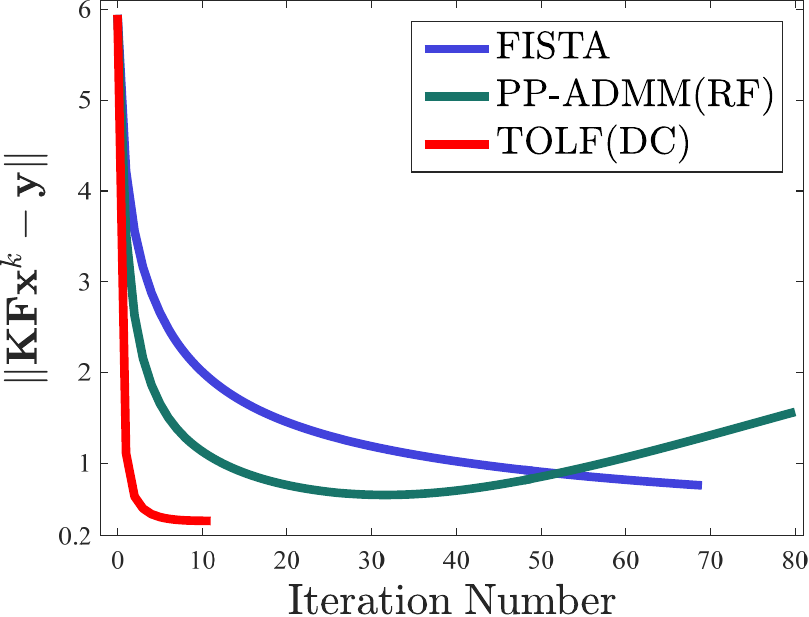}\\
		\footnotesize (a) Relative Error &\footnotesize (b) Reconstruction Error\\
	\end{tabular}
	\caption{Iteration behaviors of different optimization mechanisms (i.e., FISTA, PP-ADMM and TOLF) on Eq.~\eqref{eq:sc}. As for PP-ADMM and TOLF categories, we just adopt the settings with the best performance in Table~\ref{tab:OptimizationMechanism} (i.e., PP-ADMM (RF) and TOLF (DC)). }
	\label{fig:OptimizationMechanism}
\end{figure}

	\subsection{Iteration Scheme}\label{sec:IterationScheme}
	

		With $\Psi(\mathbf{x})$ defined in \eqref{eq:lower-app}, we have $\varphi(\mathbf{x}) = \gamma\|\mathbf{D}\mathbf{F}\mathbf{x}\|_1$ and $\psi(\mathbf{x})= \frac{1}{2}\|\mathbf{D}(\mathbf{F}\mathbf{x}-\mathbf{u})\|^2$ satisfying Assumption \ref{strongconASS} with $\mathcal{A}(\mathbf{x}) = \mathbf{D}\mathbf{F}\mathbf{x}$ and $h(\cdot) = \frac{1}{2}\| \cdot -\mathbf{D}\mathbf{u}\|^2$. With an (approximation) lower-level solution $\hat{\mathbf{x}}$ obtained from solving the optimization problem in Eq.~\eqref{eq:lower-app}, applying Theorem \ref{pro:solution} gives us the following (approximation) formula of the feasible set: 
			{
				\begin{equation*} \hat{\mathcal{X}}=\{\mathbf{x}|\mathbf{D}\mathbf{F}\mathbf{x}=\mathbf{D}{{\mathbf{o}}}, \|\mathbf{D}\mathbf{F}\mathbf{x}\|_1 \le \|\mathbf{D}\mathbf{F}{{\hat{\mathbf{x}}}}\|_1\},
			\end{equation*} }%
		where ${{\mathbf{o}}}=\mathbf{F}{{\hat{\mathbf{x}}}}$. 
		By introducing auxiliary variables $\mathbf{s}_1$, $\mathbf{s}_2$, $r$, $\mathbf{w}$, we can rewrite the feasible set as 			
		{
		\begin{equation*} 
			\begin{aligned}
			\hat{\mathcal{X}}=&\{\mathbf{x}|\mathbf{D}\mathbf{F}\mathbf{x}=\mathbf{D}{{\mathbf{o}}}, \\
			&\mathbf{D}\mathbf{F}\mathbf{x}+\mathbf{s}_{1}-\mathbf{w}=0, \\
			&\mathbf{D}\mathbf{F}\mathbf{x}-\mathbf{s}_{2}+\mathbf{w}=0,\\
			&\langle\mathbf{e},\mathbf{w}\rangle+r-\hat{t}=0,
			\mathbf{s}_1,\mathbf{s}_2,r \ge 0\}, \\
			\end{aligned}
	\end{equation*} }
		where ${\hat{t}} = \|\mathbf{D}\mathbf{F}{{\hat{\mathbf{x}}}}\|_1$ and $\mathbf{e}$ denotes the all one vector.

{
		Here we make the clarification about the reason of rewriting feasible set. Since, if $\mathbf{x}$ satifes $\| \mathbf{DFx}\|_1 <= \hat{t}$, by setting $\mathbf{w}_i = |(\mathbf{DFx})_i|$, we have $\mathbf{DFx} <= \mathbf{w}$, $\mathbf{-DFx} <= \mathbf{w}$ and $\langle\mathbf{e},\mathbf{w}\rangle <= \hat{t}$, which yields the existences of $\mathbf{s}_1,\mathbf{s}_2, r >=0$ such that $\mathbf{DFx} + \mathbf{s}_1-\mathbf{w} =0$, $\mathbf{DFx}-\mathbf{s}_2+\mathbf{w} = 0$, and $\langle\mathbf{e},\mathbf{w}\rangle = \hat{t}$. On the other hand, for any given $\mathbf{x}$, if there exist $\mathbf{w}, \mathbf{s}_1,\mathbf{s}_2, r$ satisfying $\mathbf{D}\mathbf{F}\mathbf{x}+\mathbf{s}_{1}-\mathbf{w}=0$,  $\mathbf{D}\mathbf{F}\mathbf{x}-\mathbf{s}_{2}+\mathbf{w}=0$,  $\langle\mathbf{e},\mathbf{w}\rangle+r-\hat{t}=0$ and
		$\mathbf{s}_1,\mathbf{s}_2,r \ge 0$, we can obtain from $\mathbf{DFx} <= \mathbf{w}$, $\mathbf{-DFx} <= \mathbf{w}$ that $|(\mathbf{DFx})_i| <= \mathbf{w}_i$ and thus $\langle\mathbf{e},\mathbf{w}\rangle <= \hat{t}$ implies that $\|\mathbf{DFx}\|_1 <= \hat{t}$.
	}
		
		With an additional auxiliary variable $\mathbf{z}$ and equality constraint $\mathbf{x} - \mathbf{z} = 0$, the convex bilevel optimization model can be reformulated into a single level constrained optimization problem with objective $\frac{1}{2}\|\mathbf{A}\mathbf{F}\mathbf{x}-\mathbf{y}\|^2+\eta\|\mathbf{z}\|_1 + \delta_{\ge 0}(\mathbf{s}_1,\mathbf{s}_2,r)$ and linear equality constraints 
		{
		\begin{equation*} 
		\left\{
		\begin{aligned}
		&\mathbf{x} - \mathbf{z} = 0,\\
		&\mathbf{D}\mathbf{F}\mathbf{x}=\mathbf{D}{{\mathbf{o}}}, \\
		&\mathbf{D}\mathbf{F}\mathbf{x}+\mathbf{s}_{1}-\mathbf{w}=0, \\
		&\mathbf{D}\mathbf{F}\mathbf{x}-\mathbf{s}_{2}+\mathbf{w}=0,\\
		&\langle\mathbf{e},\mathbf{w}\rangle+r-\hat{t}=0. \\
	\end{aligned}\right.
\end{equation*} 
}
	Based on this reformulation, the {proximal ADMM~\cite{boyd2011distributed}} scheme with $(\mathbf{z},\mathbf{s}_{1},\mathbf{s}_{2},r)$ as the first block and $(\mathbf{w},\mathbf{x})$ as the second block gives following updating rule:
\begin{equation*} 
	\left\{
	\begin{aligned}
	&[\mathbf{z}^{k+1}]_{i} = \mathtt{sign}([\mathbf{v}^{k}]_{i})\max\left\{0,\left|[\mathbf{v}^{k}]_{i}\right|-\zeta\right\},\\
	&[\mathbf{s}_{1}^{k+1}]_{i} = \max\left\{0,\frac{\beta([\mathbf{w}^{k}]_{i}- [\mathbf{D}\mathbf{F}\mathbf{x}^{k}]_{i})-[\bm{\lambda}_{\mathbf{s}_{1}}^{k}]_i+\tau[\mathbf{s}_{1}^{k}]_{i}}{\beta+\tau}\right\},\\
	&[\mathbf{s}_{2}^{k+1}]_{i} = \max\left\{0,\frac{\beta([\mathbf{w}^{k}]_{i}+[\mathbf{D}\mathbf{F}\mathbf{x}^{k}]_{i})+[\bm{\lambda}_{\mathbf{s}_{2}}^{k}]_i+\tau[\mathbf{s}_{2}^{k}]_i}{\beta+\tau}\right\},\\
	&r^{k+1} = \max\left\{0,\frac{\beta(\bar{t}-\langle\mathbf{e},\mathbf{w}^{k}\rangle)-{\lambda}_r^{k}+\tau r^{k}}{\beta+\tau}\right\},\\
	&\mathbf{w}^{k+1} =((2\beta +\tau)\mathbf{I}+\beta\mathbf{e}\mathbf{e}^{\top})^{-1}\mathbf{b}^{k+1},\\
	&\mathbf{x}^{k+1} = ({\mathbf{c}^{k+1}})^{-1}{\mathbf{d}^{k+1}},\\
	&\bm{\lambda}_\mathbf{x}^{k+1} = \bm{\lambda}_\mathbf{x}^{k}+\beta(\mathbf{D}\mathbf{F}\mathbf{x}^{k+1}-\mathbf{D}{{\mathbf{o}}}),\\
	&\bm{\lambda}_\mathbf{z}^{k+1} = \bm{\lambda}_\mathbf{z}^{k}+\beta(\mathbf{z}^{k+1}-\mathbf{x}^{k+1}),\\
	&\bm{\lambda}_{\mathbf{s}_{1}}^{k+1} = \bm{\lambda}_{\mathbf{s}_{1}}^{k}+\beta(\mathbf{D}\mathbf{F}\mathbf{x}^{k+1} + \mathbf{s}_{1}^{k+1}- \mathbf{w}^{k+1}),\\
	&\bm{\lambda}_{\mathbf{s}_{2}}^{k+1} = \bm{\lambda}_{\mathbf{s}_{2}}^{k}+\beta(\mathbf{D}\mathbf{F}\mathbf{x}^{k+1} - \mathbf{s}_{2}^{k+1}+ \mathbf{w}^{k+1}),\\
	&{\lambda}_{r}^{k+1} = {\lambda}_{r}^{k}+\beta(\langle\mathbf{e},\mathbf{w}^{k+1}\rangle+r^{k+1}-{\hat{t}}),\\
	\end{aligned}
	\right.
	\end{equation*} 
	where $\beta, \tau, \zeta=\frac{\eta}{\beta+\tau} > 0,$ are respectively the penalty and regularization parameters, $[\cdot]_i$ denotes the $i$-th element of the given vector, and $\mathbf{I}$ is the identity matrix. The symbol $\mathtt{sign}(\cdot)$ denotes the sign function. The other variables are presented as
	{
	\begin{equation*}
		\left\{
		\begin{aligned}
			\mathbf{v}^{k}&=\frac{\beta\mathbf{x}^{k}-\bm{\lambda}_\mathbf{z}^{k}+\tau\mathbf{z}^{k}}{\beta+\tau},\\
			{\mathbf{b}}^{k+1}&=\beta\mathbf{s}_{1}^{k+1}+\bm{\lambda}_{\mathbf{s}_{1}}^{k}+\beta\mathbf{s}_{2}^{k+1}-\bm{\lambda}_{\mathbf{s}_{2}}^{k}+\varrho\mathbf{e}+\tau\mathbf{w}^{k},\\
			{\mathbf{c}}^{k+1}&={\mathbf{F}^{\top}\mathbf{A}^{\top}\mathbf{A}\mathbf{F}+3\beta\mathbf{F}^{\top}\mathbf{D}^{\top}\mathbf{D}\mathbf{F}+(\beta+\tau)\mathbf{I}},\\
			{\mathbf{d}}^{k+1}&=\mathbf{F}^{\top}\mathbf{A}^{\top}\mathbf{y}+\beta\mathbf{F}^{\top}\mathbf{D}^{\top}\mathbf{D}{{\mathbf{o}}} -\mathbf{F}^{\top}\mathbf{D}^{\top}\bm{\lambda}_{\mathbf{x}}^{k}+\beta\mathbf{z}^{k+1}+\bm{\lambda}_\mathbf{z}^{k} \\
			&+\mathbf{F}^{\top}\mathbf{D}^{\top}\left(\beta(\mathbf{s}_{2}^{k+1}-\bm{\lambda}_{\mathbf{s}_{2}}^{k})-\beta(\mathbf{s}_{1}^{k+1}+\bm{\lambda}_{{\mathbf{s}_{1}}}^{k})\right)+\tau\mathbf{x}^{k},\\
			\varrho&=\beta\hat{t}-\beta{r}^{k+1}-{\lambda}_r^{k}.\\
		\end{aligned}\right.
	\end{equation*} 
}

	\begin{figure}[t]
		\centering
		\begin{tabular}{c@{\extracolsep{0.4em}}c}
			\includegraphics[width=0.46\linewidth]{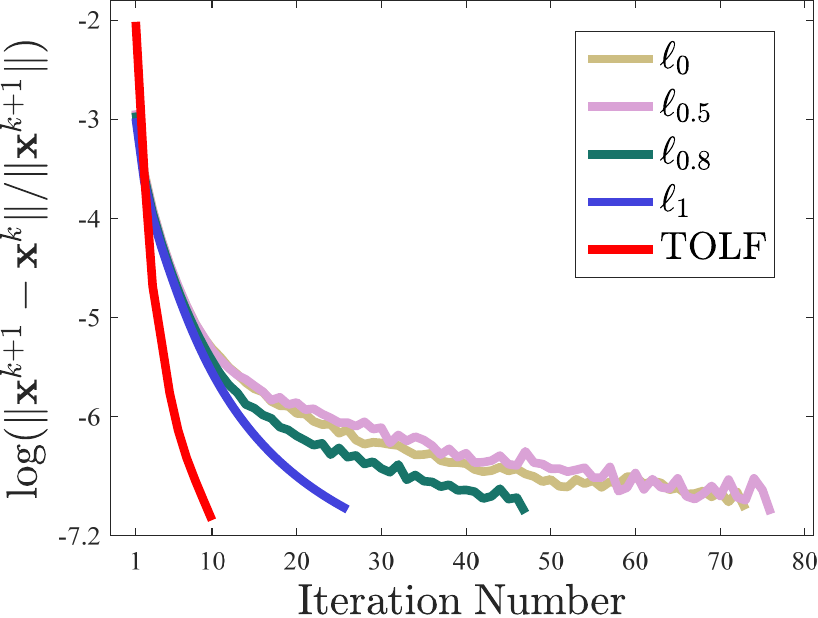}&
			\includegraphics[width=0.46\linewidth]{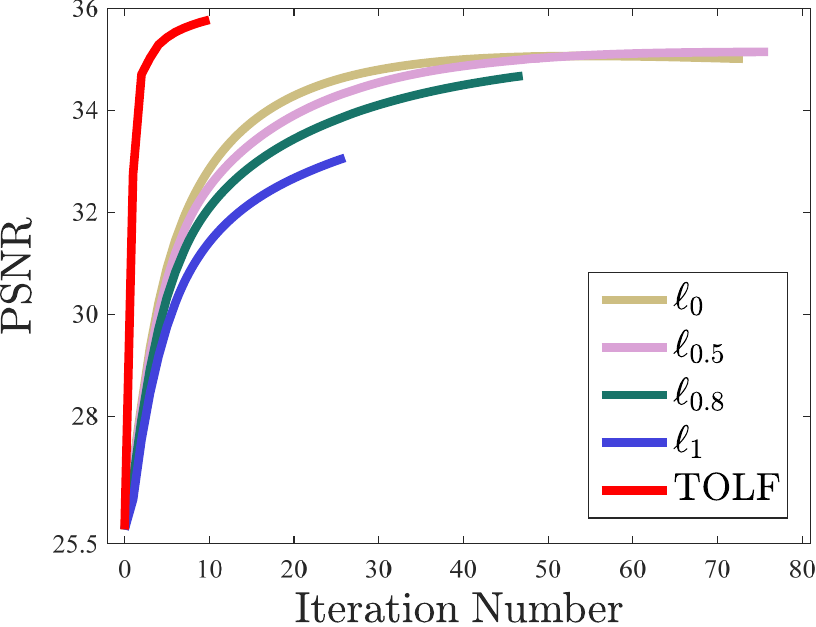}\\
			\footnotesize (a) Relative Error &\footnotesize (b) PSNR\\
		\end{tabular}
		\caption{Iteration behaviors of convex and non-convex optimization formulations. These models are based on Eq.~\eqref{eq:sc} but with different regularization, i.e., convex $\ell_1$-norm, solved by FISTA (denoted as $\ell_1$) and TOLF, and non-convex $\ell_p$-norm ($p=0,0.5,0.8$), solved by \cite{li2015accelerated}. }
		\label{fig:NonConvexFormulations}
	\end{figure}

\section{Experimental Results}\label{sec:exp}
This section first explored the iteration behaviors of TOLF to verify our theoretical results of TOLF and then researched parameter and network architecture analysis. Finally, we compared our proposed algorithm with state-of-the-art approaches on three real-world image processing applications. All the experiments were conducted on a PC with an Intel Core i7 CPU at 3.7GHz, 32GB RAM, and an NVIDIA GeForce GTX 1080Ti 11GB GPU.

\begin{table}[t]
	\centering
	\caption{Quantitative comparisons between compound regularization and our TOLF on different settings. ``*'' represents that DC is performed before denoising.}
	\renewcommand\arraystretch{1.25}
	\begin{tabular}{c@{\extracolsep{0.1cm}}c@{\extracolsep{0.22cm}}c@{\extracolsep{0.22cm}}c@{\extracolsep{0.22cm}}c@{\extracolsep{0.22cm}}c@{\extracolsep{0.22cm}}c}
		\toprule
		\multirow{2}{*}{\footnotesize Method}&\multicolumn{3}{c}{\footnotesize Compound Regularization }&\multicolumn{3}{c}{{\footnotesize TOLF}}\\
		\cline{2-7}
		~&\footnotesize (DC)&\footnotesize (RF)&\footnotesize (DnCNN)&\footnotesize (DC)&\footnotesize (RF$^*$)&\footnotesize (DnCNN$^*$)\\
		\midrule
		\footnotesize 	PSNR&\footnotesize {30.27}&\footnotesize {31.73}&\footnotesize {32.54}&\footnotesize 31.64&\footnotesize 32.22&\footnotesize 33.13\\
		\footnotesize SSIM&\footnotesize {0.8684}&\footnotesize {0.9001}&\footnotesize {0.9197}&\footnotesize 0.8997&\footnotesize {0.9103}&\footnotesize 0.9324\\
		\bottomrule
	\end{tabular}
	\label{tab:Compound}
\end{table}

\begin{figure}[t]
	\centering
	
	{
		\begin{tabular}{c@{\extracolsep{0.4em}}c}
			\includegraphics[width=0.46\linewidth]{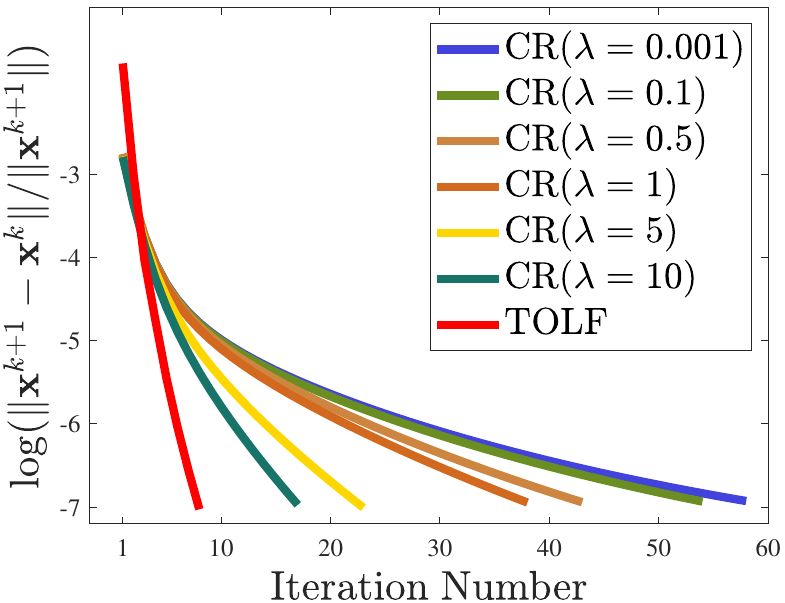}&
			\includegraphics[width=0.46\linewidth]{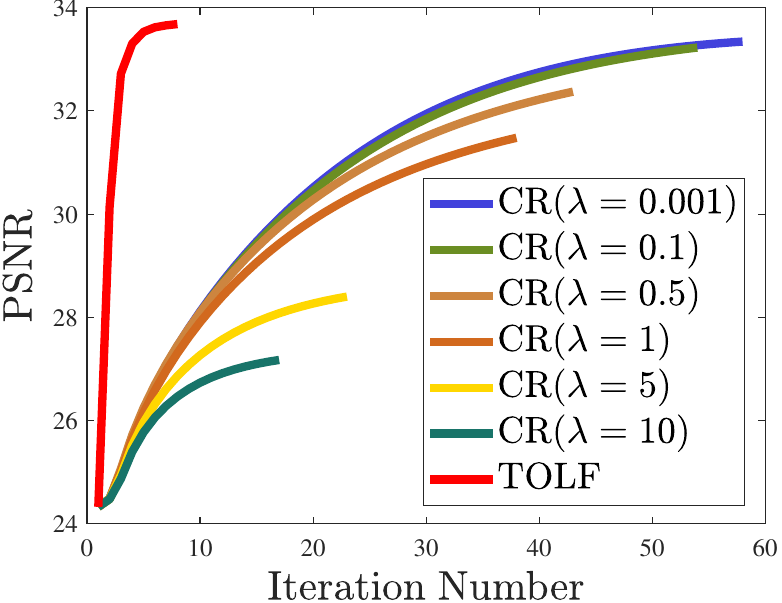}\\
			\footnotesize (a) Relative Error &\footnotesize (b) PSNR\\
		\end{tabular}	
	}
	\caption{{Iteration behavior of Compound Regularization (CR) with different regularization parameters $\lambda$ and our TOLF. We just adopt the settings with the best performance in Table~\ref{tab:Compound} (i.e., Compound Regularization (DnCNN) and TOLF (DnCNN)).}}
	\label{fig:Compound}
\end{figure}

\begin{figure}[t]
	\centering
	{
		\begin{tabular}{c@{\extracolsep{0.2em}}c@{\extracolsep{0.2em}}c}
			\includegraphics[width=0.31\linewidth]{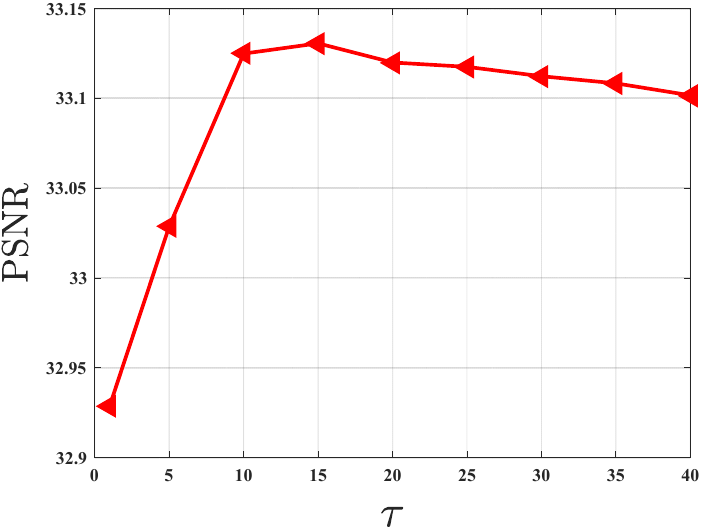}&
			\includegraphics[width=0.31\linewidth]{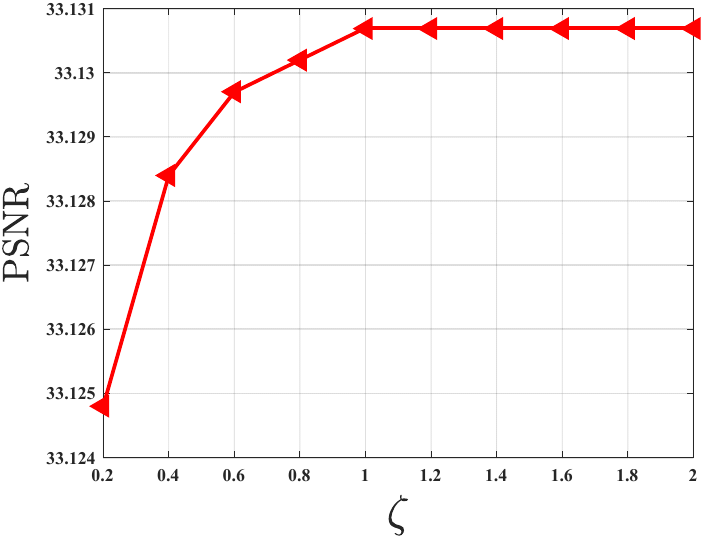}&
			\includegraphics[width=0.31\linewidth]{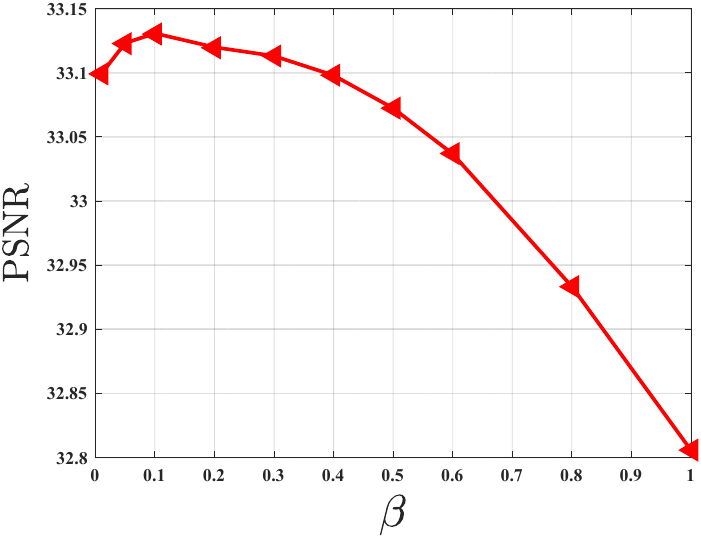}\\
			\footnotesize (a) Effects of $\tau$  &\footnotesize (b) Effects of $\zeta$&\footnotesize (b) Effects of $\beta$ \\
		\end{tabular}
	}
	\caption{{Sensitivity analysis with respect to parameters $\tau, \zeta$, and $\beta$. }}
	\label{fig:Parameters}
\end{figure}

\begin{table}[t]
	\centering
	\caption{{Quantitative results of different networks (i.e., TNRD, DnCNN, and IRCNN). ``*'' represents that DC is performed before denoising. }}
	{
		\renewcommand\arraystretch{1.25}
		\begin{tabular}{c@{\extracolsep{0.1cm}}c@{\extracolsep{0.15cm}}c@{\extracolsep{0.15cm}}c@{\extracolsep{0.15cm}}c@{\extracolsep{0.15cm}}c@{\extracolsep{0.15cm}}c}
			\toprule
			\multirow{2}{*}{\footnotesize Method}&\multicolumn{3}{c}{\footnotesize PP-ADMM}&\multicolumn{3}{c}{\footnotesize TOLF}\\
			\cline{2-7}
			~&\footnotesize (TNRD)&\footnotesize (IRCNN)&\footnotesize (DnCNN)&\footnotesize (TNRD$^*$)&\footnotesize (IRCNN$^*$)&\footnotesize (DnCNN$^*$)\\
			\midrule
			\footnotesize 	PSNR&\footnotesize 31.63&\footnotesize 32.58&\footnotesize 32.60&\footnotesize 32.33&\footnotesize 32.97&\footnotesize 33.13\\
			\footnotesize SSIM&\footnotesize 0.8902&\footnotesize 0.9240&\footnotesize 0.9296&\footnotesize 0.9112&\footnotesize {0.9295}&\footnotesize 0.9324\\
			\bottomrule
		\end{tabular}
	}
	\label{tab:DifferentNetworks}
\end{table}

	\subsection{Iteration Behaviors Analysis}\label{sec:IterationBehaviors}
	
	In this part, we compared TOLF to classical convex, non-convex and plug-and-play optimization techniques on the image restoration task. Specifically, we first considered to adopt different optimization mechanisms, including standard convex scheme (e.g., FISTA~\cite{beck2009fast}), plug-and-play strategy (i.e., Plug-and-Play ADMM~\cite{chan2016plug}, PP-ADMM for short) and our TOLF, to solve the classical convex model in Eq.~\eqref{eq:sc}. As for PP-ADMM and TOLF, we introduced two different task-specific operations including the classical image filer, i.e., the recursive filter~\cite{chan2016plug} (RF for short), and the task-based deconvolution process~\cite{liu2019convergence}, i.e., $\arg\min_{\mathbf{x}}\|\mathbf{KF}\mathbf{x}-\mathbf{y}\|^2 + \alpha\|\mathbf{Fx}-\mathbf{y}\|^2$ with a trade-off $\alpha>0$ (fix it as $10^{-4}$ for all the experiments) (DC for short). 
	{The lower-level subproblem was solved by APG~\cite{bioucas2007new} with the relative errors $\|\mathbf{x}^{k+1}-\mathbf{x}^{k}\|/\|\mathbf{x}^{k+1}\|\leq10^{-3}$, to find an approximate solution to $\bar{\mathbf{x}}$, which works as an initialization to the problem in Eq.~\eqref{rechare}. }
	In Table~\ref{tab:OptimizationMechanism}, we reported quantitative performances of these compared methods on an example image from Levin et al.'s benchmark~\cite{Levin2009Understanding}. Since we cannot obtain convergence sequences for PP-ADMM even after 80 iterations, we had to report the best results during their iterations, i.e., the 37th and 29th steps for PP-ADMM (RF) and PP-ADMM (DC), respectively.
	It can be seen that in most cases introducing task-specific operations improved the performance of Eq.~\eqref{eq:sc} for image restoration. That is, the results of PP-ADMM (RF), TOLF (DC) and TOLF (RF) were all better than the classical FISTA method. Meanwhile, we also observed that PP-ADMM (DC) was even worse than FISTA. This is mainly because the DC operation is repeatedly performed within the plug-and-play mechanism, which may overly smooth the image details.
	
\begin{table*}[t]
	\renewcommand\arraystretch{1.2} 
	\centering
	{
		\caption{Averaged image restoration performance on Levin et al.' benchmark~\cite{Levin2009Understanding}.} \label{tab:ImageRestorationQuan}
		\begin{tabular}{c@{\extracolsep{0.65em}}c@{\extracolsep{0.65em}}c@{\extracolsep{0.65em}}c@{\extracolsep{0.65em}}c@{\extracolsep{0.65em}}c@{\extracolsep{0.65em}}c@{\extracolsep{0.65em}}c@{\extracolsep{0.65em}}c@{\extracolsep{0.65em}}c@{\extracolsep{0.65em}}c@{\extracolsep{0.65em}}c@{\extracolsep{0.65em}}c@{\extracolsep{0.65em}}c@{\extracolsep{0.65em}}c@{\extracolsep{0.65em}}c}
			\toprule
			&\footnotesize FTVd&\footnotesize FISTA&\footnotesize HL&\footnotesize IDDBM3D&\footnotesize EPLL&\footnotesize MLP&\footnotesize IRCNN&\footnotesize MSWNNM&\footnotesize FDN&\footnotesize GLRA&\footnotesize PP-ADMM&\footnotesize MEDAEP&\footnotesize eFIMA&\footnotesize iFIMA& \footnotesize TOLF \\
			\midrule
			\footnotesize 	PSNR&\footnotesize 29.38&\footnotesize 31.81&\footnotesize 30.12& \footnotesize 31.53&\footnotesize 31.65& \footnotesize 31.32&\footnotesize 32.28& \footnotesize 32.50&\footnotesize 32.04&\footnotesize 19.75&\footnotesize 32.60&\footnotesize 19.77&\footnotesize 32.68&\footnotesize 32.72& \footnotesize \textbf{33.13}\\
			\footnotesize SSIM&\footnotesize 0.8819&\footnotesize 0.9010&\footnotesize 0.8961& \footnotesize 0.9043&\footnotesize 0.9258&\footnotesize 0.8991& \footnotesize 0.9200&\footnotesize 0.9247&\footnotesize 0.9277&\footnotesize 0.5134&\footnotesize 0.9296&\footnotesize 0.4639&\footnotesize 0.9291&\footnotesize 0.9296& \footnotesize \textbf{0.9324}\\
			\bottomrule
		\end{tabular}
	}
\end{table*}

\begin{figure*}[t]
	\centering
	{
		\begin{tabular}{c@{\extracolsep{0.3em}}c@{\extracolsep{0.3em}}c@{\extracolsep{0.3em}}c@{\extracolsep{0.3em}}c@{\extracolsep{0.3em}}c@{\extracolsep{0.3em}}c@{\extracolsep{0.3em}}c@{\extracolsep{0.3em}}c}
			\includegraphics[width=0.101\linewidth]{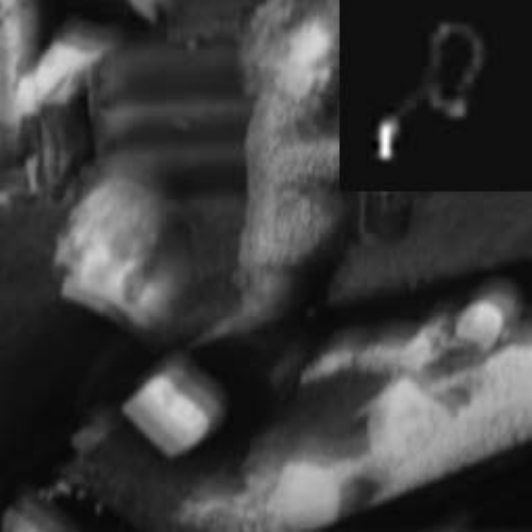}&
			\includegraphics[width=0.101\linewidth]{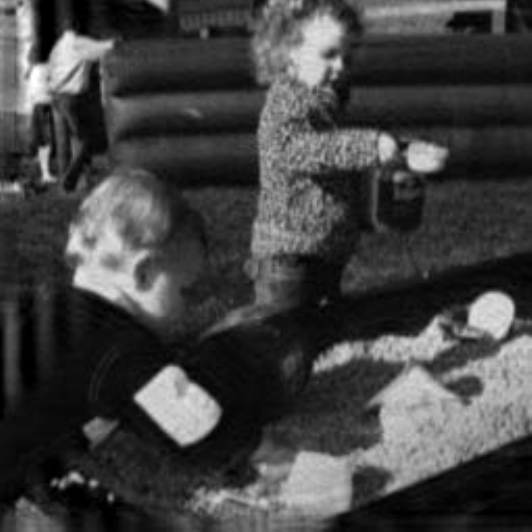}&
			\includegraphics[width=0.101\linewidth]{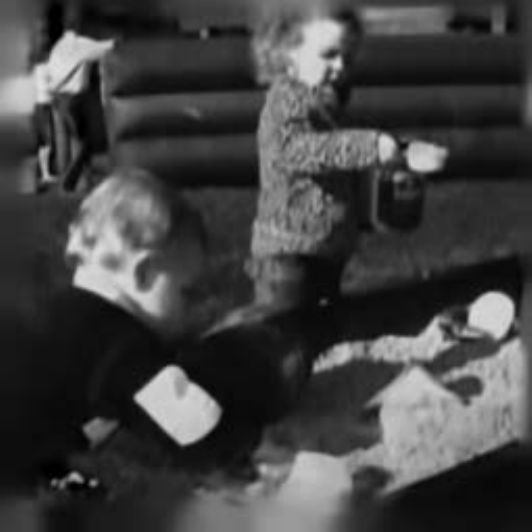}&
			\includegraphics[width=0.101\linewidth]{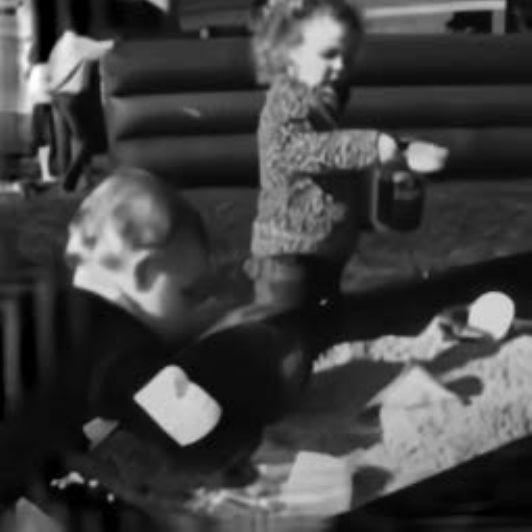}&
			\includegraphics[width=0.101\linewidth]{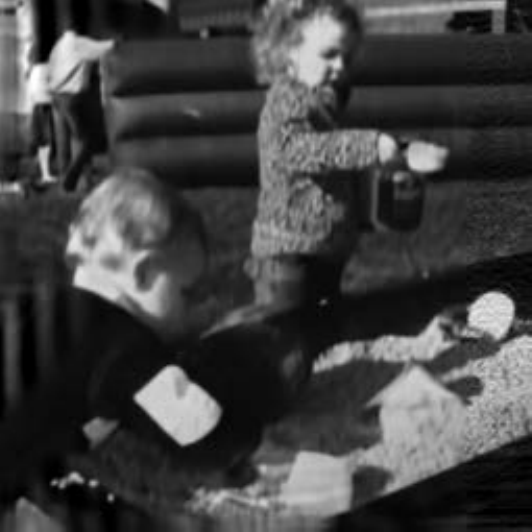}&
			\includegraphics[width=0.101\linewidth]{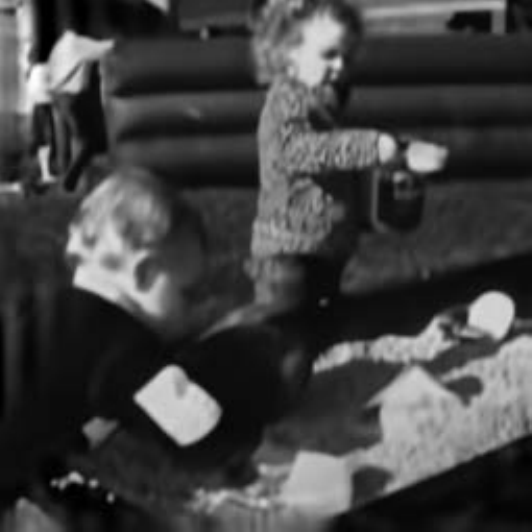}&
			\includegraphics[width=0.101\linewidth]{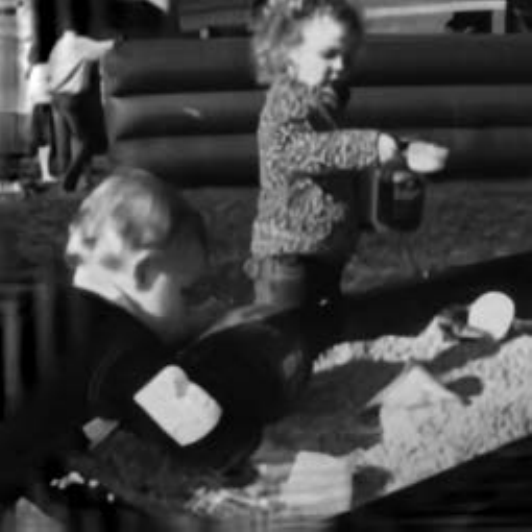}&
			\includegraphics[width=0.101\linewidth]{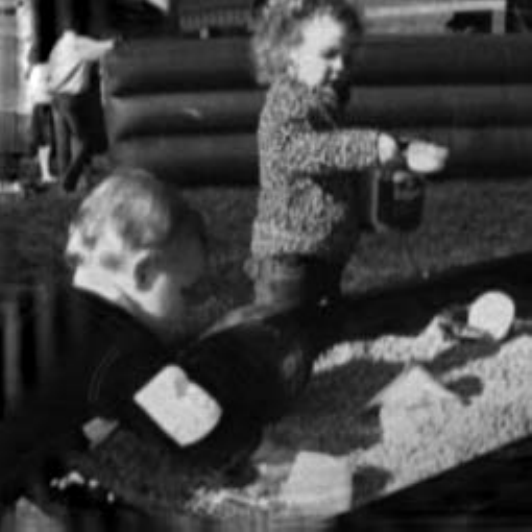}&
			\includegraphics[width=0.101\linewidth]{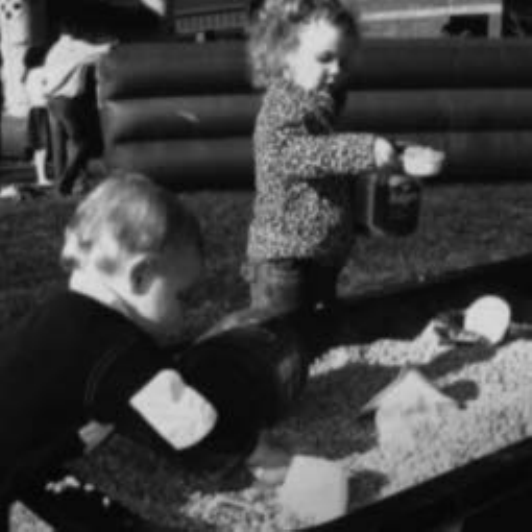}\\
			\footnotesize PSNR/SSIM&\footnotesize 31.21/0.9104& \footnotesize 30.34/0.9253&\footnotesize 31.45/0.9282&\footnotesize 31.42/0.9267&\footnotesize 31.83/0.9261&\footnotesize 31.50/0.9342&\footnotesize \textbf{32.01/0.9346}&\footnotesize \textemdash\\
			\includegraphics[width=0.101\linewidth]{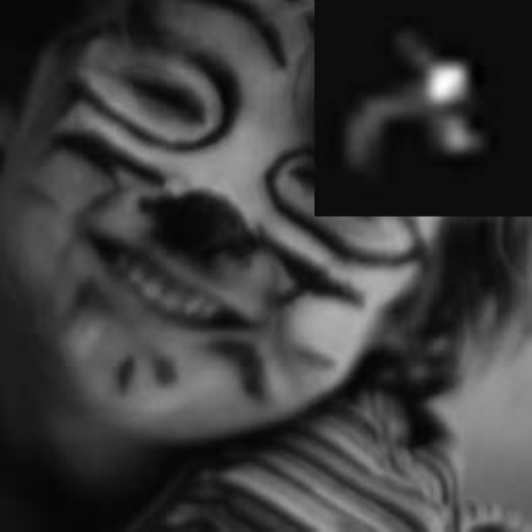}&
			\includegraphics[width=0.101\linewidth]{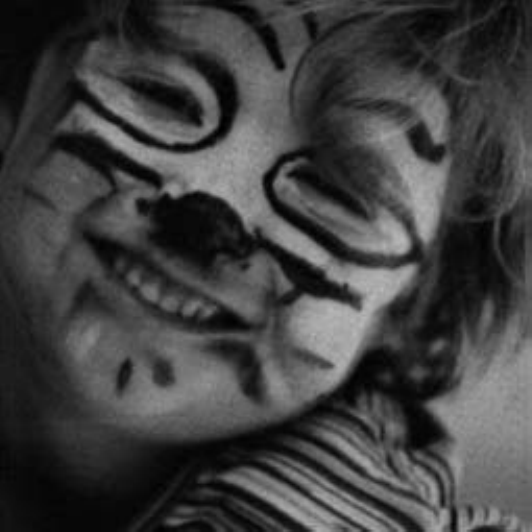}&
			\includegraphics[width=0.101\linewidth]{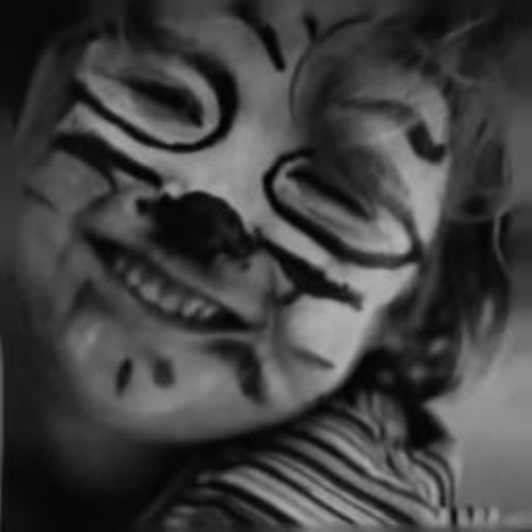}&
			\includegraphics[width=0.101\linewidth]{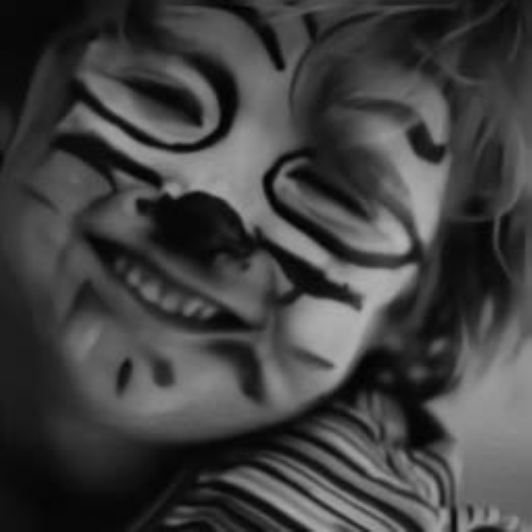}&
			\includegraphics[width=0.101\linewidth]{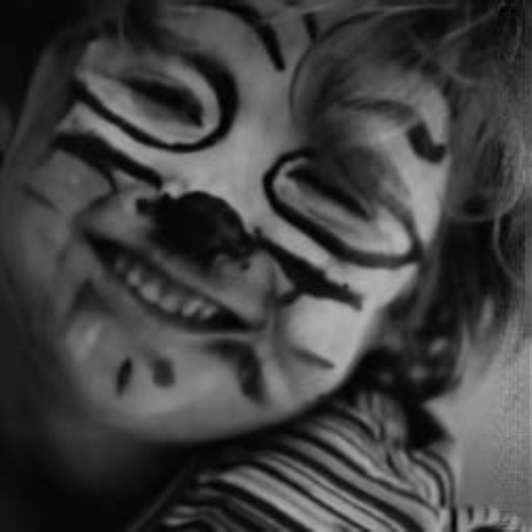}&
			\includegraphics[width=0.101\linewidth]{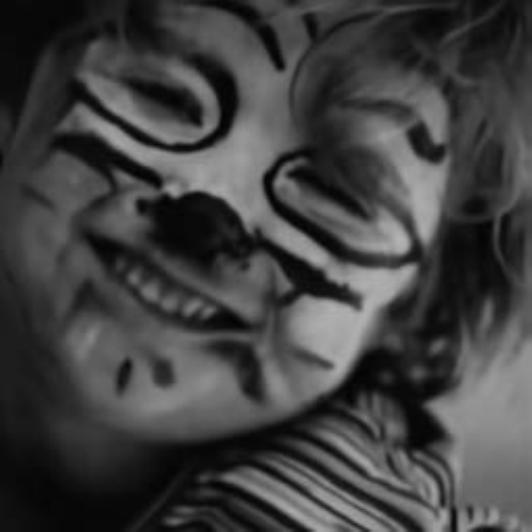}&
			\includegraphics[width=0.101\linewidth]{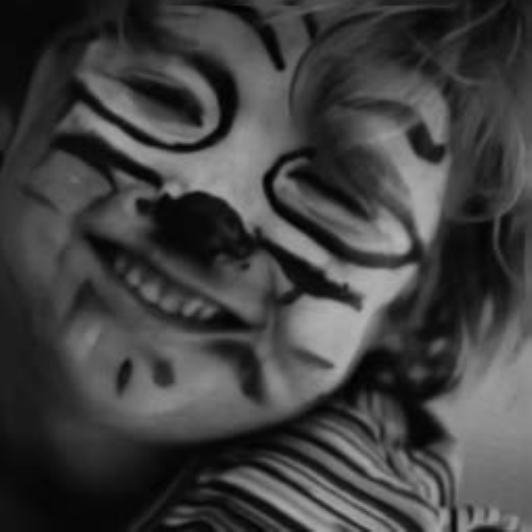}&
			\includegraphics[width=0.101\linewidth]{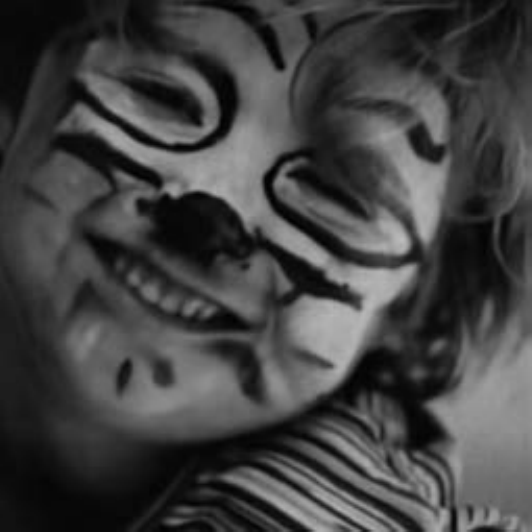}&
			\includegraphics[width=0.101\linewidth]{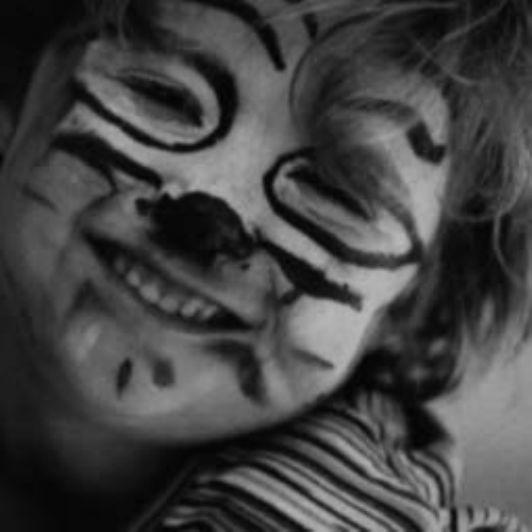}\\
			\footnotesize PSNR/SSIM&\footnotesize 35.20/0.9495& \footnotesize 35.75/0.9672&\footnotesize 35.55/0.9629&\footnotesize 36.03/0.9679&\footnotesize 35.51/0.9633&\footnotesize 36.33/0.9747&\footnotesize \textbf{37.97/0.9749}&\footnotesize \textemdash\\
			\footnotesize Input&\footnotesize FISTA&\footnotesize EPLL&\footnotesize IRCNN&\footnotesize MSWNNM&\footnotesize PP-ADMM&\footnotesize iFIMA& \footnotesize TOLF&\footnotesize Ground Truth\\
		\end{tabular}
		\caption{Image restoration results on two examples in Levin et al.' dataset. We chose top six performance methods from Table.~\ref{tab:ImageRestorationQuan}.} 			\label{fig:ImageRestoration1}
	}
\end{figure*}

\begin{figure*}[t]
	\centering
	{
		\begin{tabular}{c@{\extracolsep{0.3em}}c@{\extracolsep{0.3em}}c@{\extracolsep{0.3em}}c@{\extracolsep{0.3em}}c@{\extracolsep{0.3em}}c}
			\includegraphics[width=0.155\linewidth]{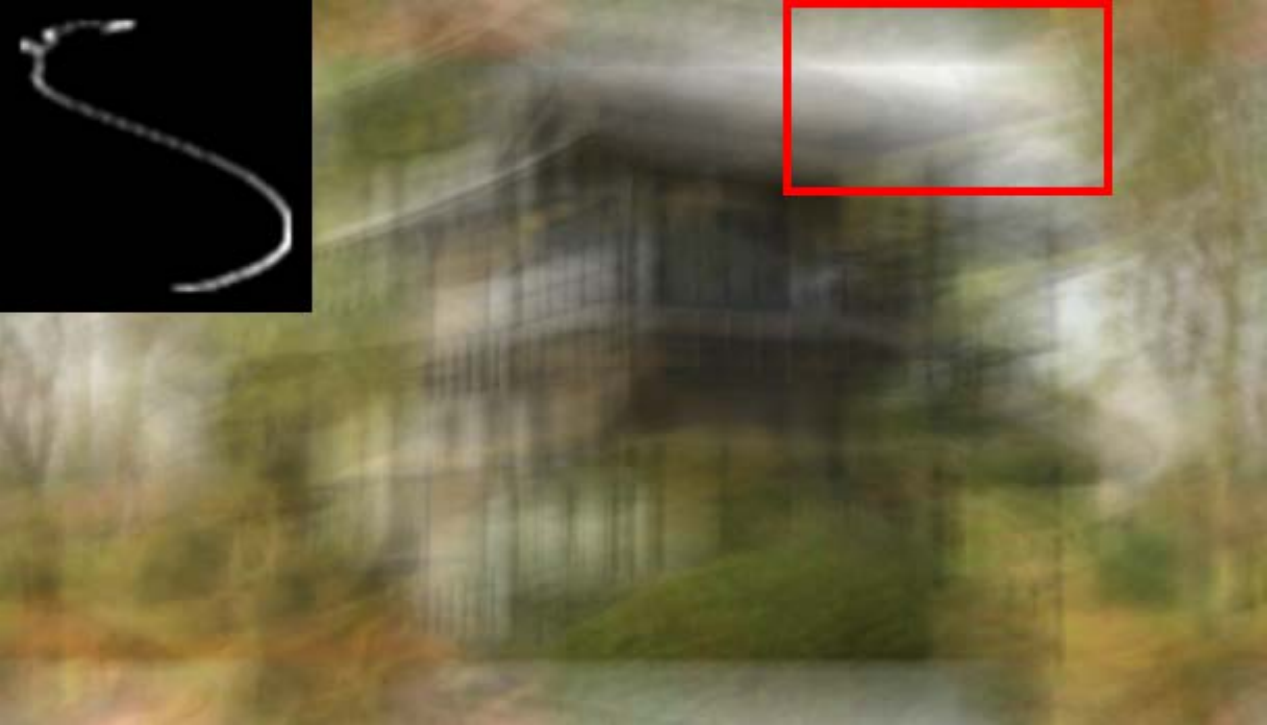}&
			\includegraphics[width=0.155\linewidth]{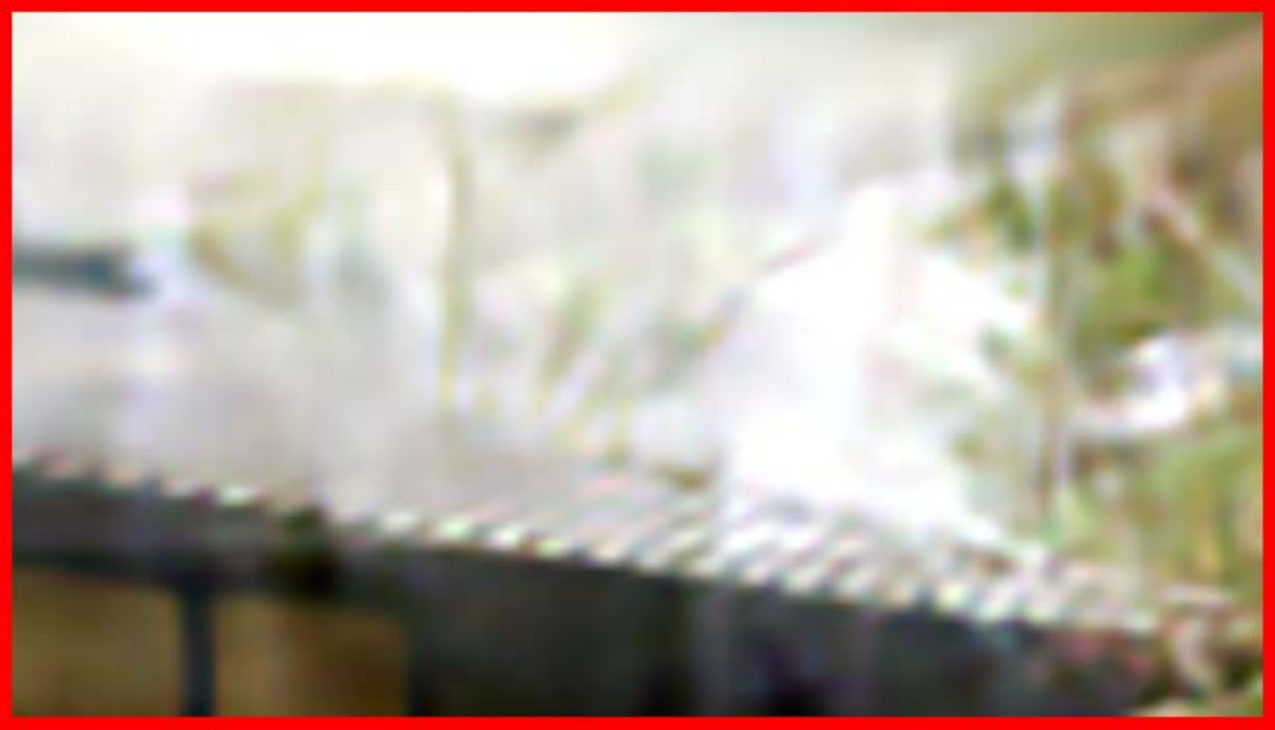}&
			\includegraphics[width=0.155\linewidth]{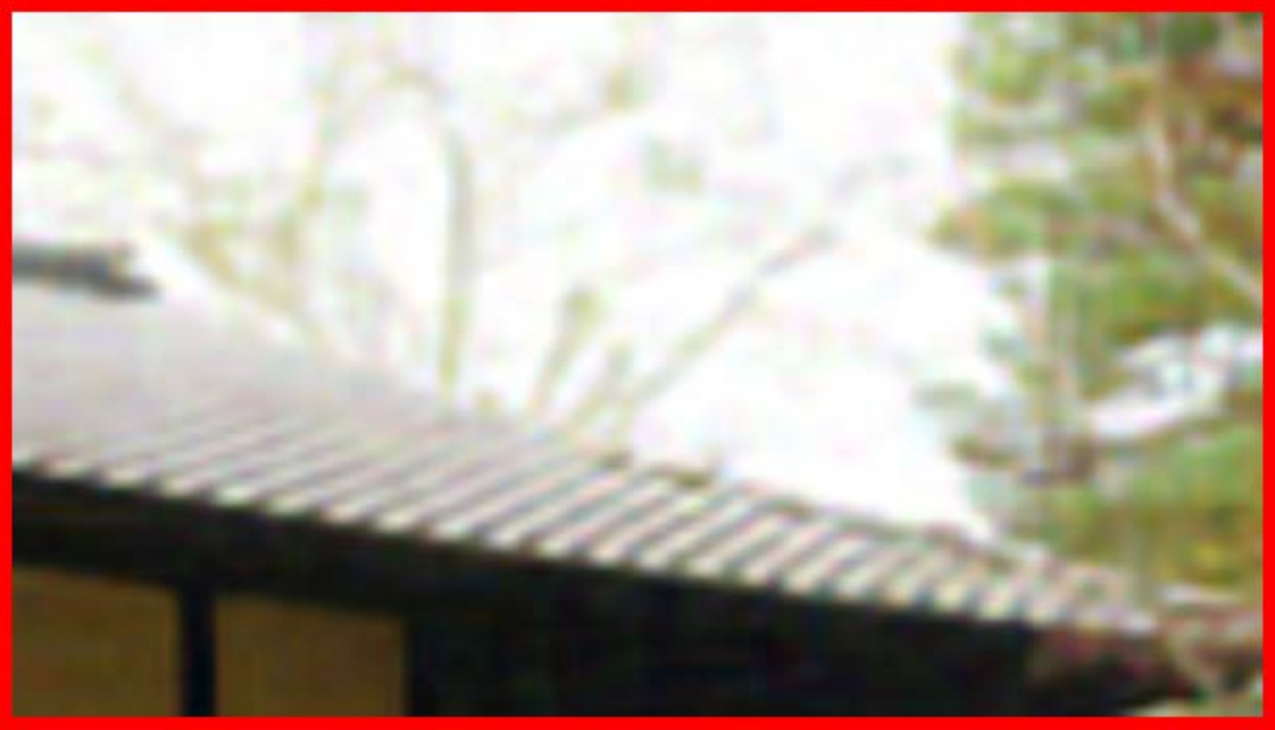}&
			\includegraphics[width=0.155\linewidth]{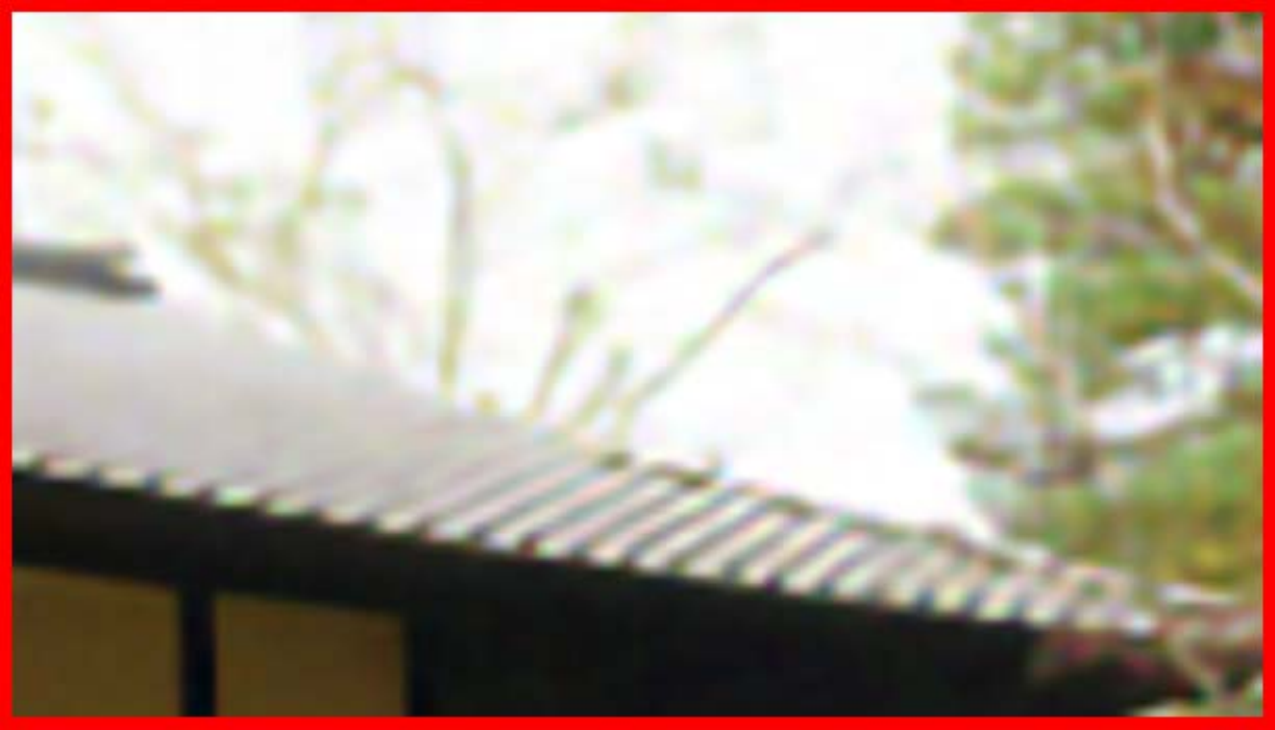}&
			\includegraphics[width=0.155\linewidth]{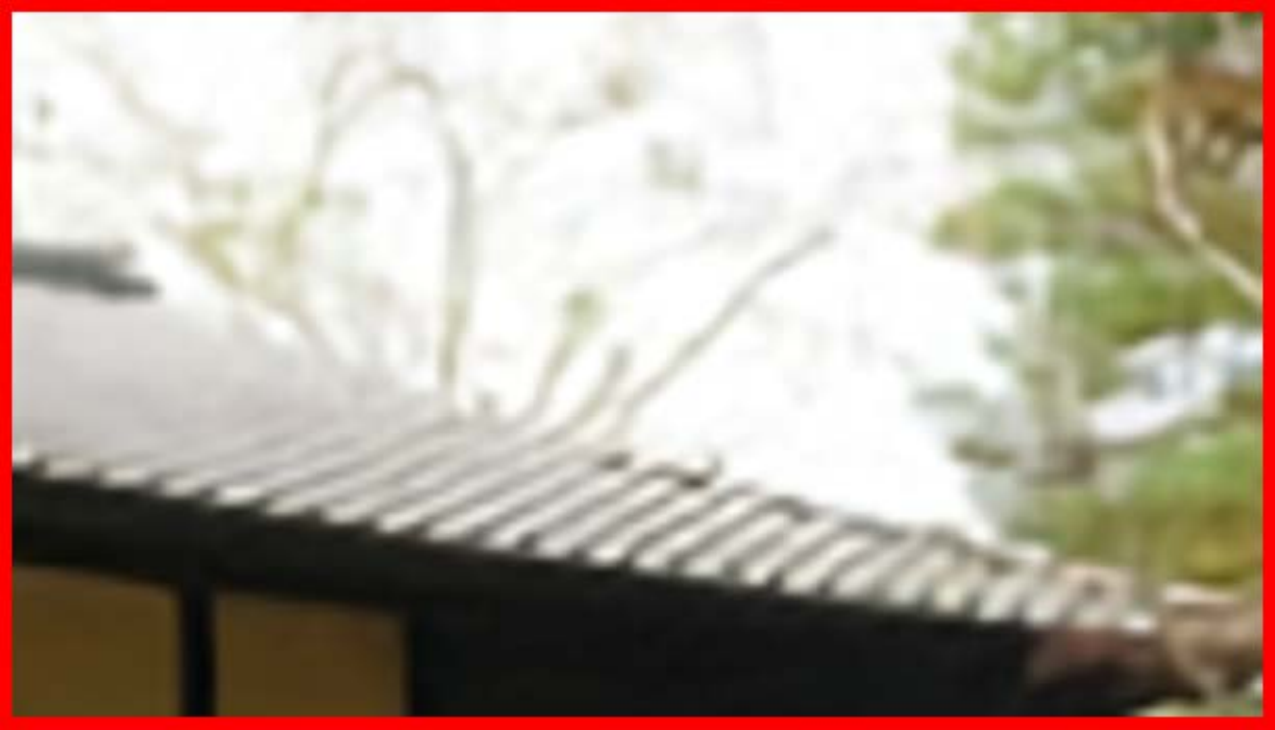}&
			\includegraphics[width=0.155\linewidth]{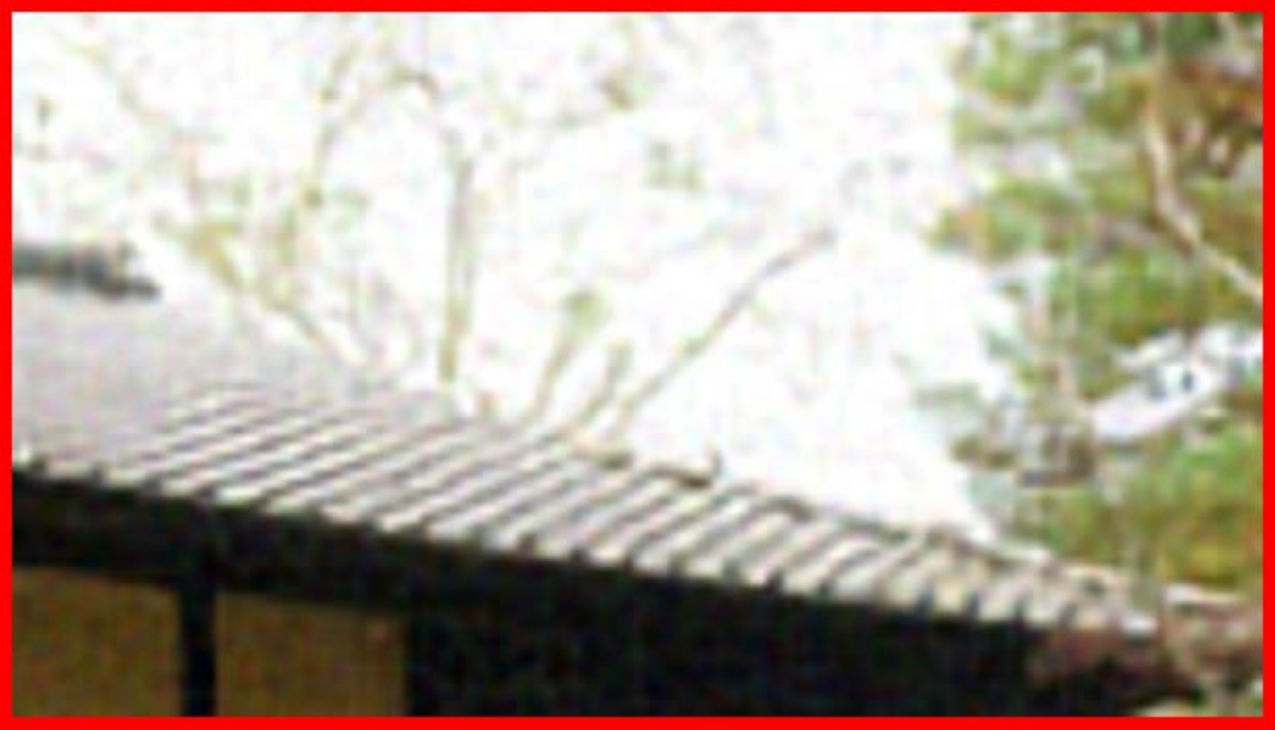}\\
			\footnotesize Input&\footnotesize EPLL: 19.08&\footnotesize IDDBM3D: 25.27&\footnotesize MSWNNM: 25.81&\footnotesize iFIMA: 25.91&\footnotesize TOLF: \textbf{26.48}\\
		\end{tabular}
		\caption{Image restoration results on a challenging color image with a large-sized kernel ($75\times 75$). The PSNR score is reported below each subfigure.}
	}
	\label{fig:ImageRestoration2}
\end{figure*}

Notably, our TOLF (DC) is superior to TOLF (RF), but PP-ADMM (RF) is superior to PP-ADMM (DC). It is because DC was only performed at the first stage of TOLF, which actually provided a task-specific warm start to optimize the blurry input for our iterations, but RF is task-independent which just presents a denoising operation for the plugged step. In other words, different from PP-ADMM, the task-specific operation is a crucial component for the warm start procedure in our TOLF. 

We also compared the iteration behaviors of FISTA, PP-ADMM and TOLF in Fig.~\ref{fig:OptimizationMechanism}. For the last two methods, we only chose the inner operator $\mathcal{T}$ with better performances in Table~\ref{tab:OptimizationMechanism} (i.e., RF for PP-ADMM and DC for TOLF) to provide clearer illustrations. We observed that the speed of FISTA was slow and the iterations converged after 70 steps. It also can be seen that PP-ADMM (RF) did not converge even after 80 steps, but it obtained lower reconstruction errors than FISTA near the 30th step (see the right subfigure). In contrast, TOLF (DC) converged only after 10 iterations and achieved the lowest reconstruction errors.

Next, we adopted some recently proposed non-convex regularization techniques~\cite{krishnan2009fast,pan2016l0} (e.g., replace $\ell_1$-norm by $\ell_p$-norm, $p=0,0.5,0.8$) to reformulate Eq.~\eqref{eq:sc} for image restoration. The same non-convex accelerated proximal gradient scheme \cite{li2015accelerated} was performed to solve these $\ell_p$-regularized models. Their iteration behaviors are visualized in Fig.~\ref{fig:NonConvexFormulations}. We also plotted the iteration curves of FISTA and our TOLF on the original $\ell_1$-norm regularized model in this figure. It can be seen that these non-convex regularization models improved the practical performance (i.e., PSNR in the right subfigure) of the original convex model in Eq.~\eqref{eq:sc} (solved by FISTA, denoted as $\ell_1$), but their trajectories fluctuated after several iterations (see the left subfigure). In contrast, TOLF obtained the fastest convergence speed and the best final performance among all the compared methods, even by only solving a convex model.

Finally, we presented numerical results and iterative behaviors in terms of the compound regularization (i.e., {$\min_{\mathbf{x}}F(\mathbf{x})+\lambda\Psi(\mathbf{x})$}, where $\lambda$ is the positive balancing parameter) and our model (i.e., $\min_{\mathbf{x}}F(\mathbf{x}) \;\; s.t. \ \mathbf{x}\in \arg\min_{\mathbf{x}}\Psi(\mathbf{x})$). 
To ensure fairness, we adopted the proximal ADMM (used for our TOLF) to solve the model of compound regularization. In addition, we considered the same $\mathcal{T}$ to define $\mathbf{u}$ in $\Psi(\mathbf{x})$. As presented in Table~\ref{tab:OptimizationMechanism}, we recognized that DC (deconvolution process) is a crucial step for obtaining $\mathbf{u}$, so we  would like to emphasize that we performed DC before denoising operation (e.g., RF, and DnCNN~\cite{zhang2017beyond}) in the following experiments. 	
Table~\ref{tab:Compound} reported quantitative results of these two models by adopting different settings. {Note that $\lambda=0.1$ in the compound regularization}. Thanks to our newly-introduced latent feasibility, it can be easily seen that our results were consistently superior to compound regularization in all cases. 
The fact that the quantitative scores of TOLF (RF$^*$) (32.22/0.9103) obtained a significant improvement than TOLF (RF) (30.95/0.8871) ($^*$ represents that DC is performed before denoising), also justified the necessity of the task-specific operation for T. 
{Fig.~\ref{fig:Compound} displayed iteration behaviors of compound regularization and our TOLF. To make a rigorous evaluation, we change settings of compound regularization by adopting different $\lambda$. We can easily observe that with the value of $\lambda$ decreasing, the compound  regularization converged more slowly but obtained better performance. By contrast, our TOLF realized a higher value at a faster speed than all naive compound regularizations with different regularization parameters. This experiment indicates that compound regularization cannot promote numerical improvement and accelerate the convergence speed simultaneously, while TOLF we proposed can obtain a satisfying outcome.
}

\begin{figure*}[t]
	\centering
	\begin{tabular}{c@{\extracolsep{0.5em}}c@{\extracolsep{0.5em}}c}
		\includegraphics[width=0.31\linewidth]{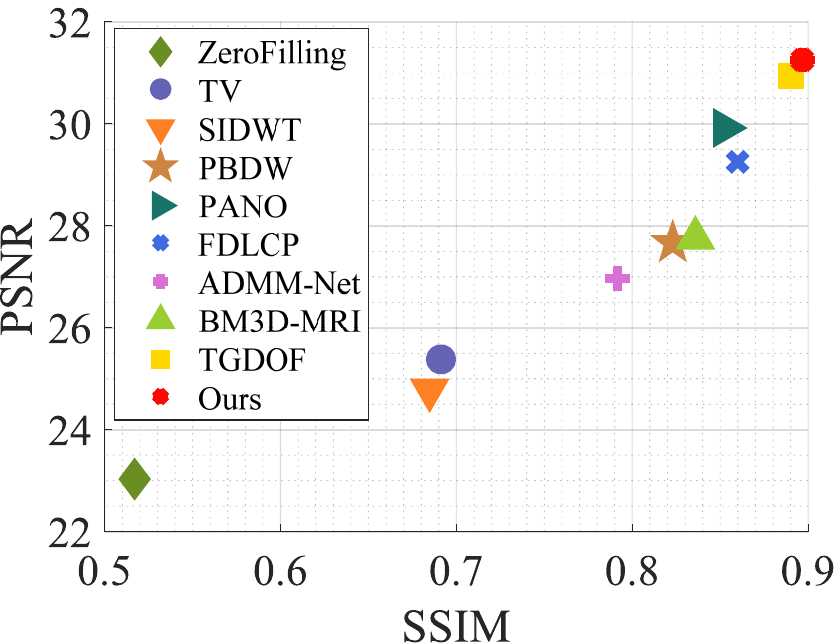}&
		\includegraphics[width=0.31\linewidth]{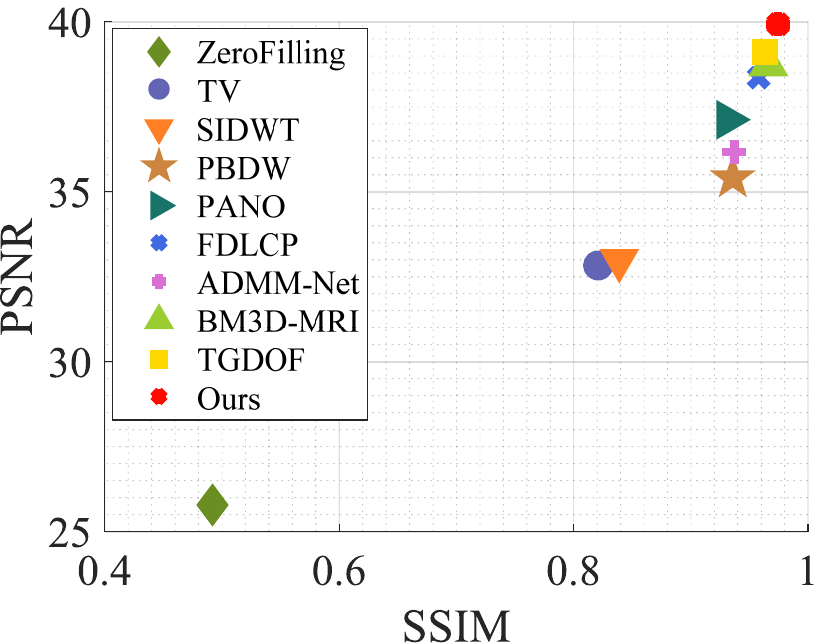}&
		\includegraphics[width=0.31\linewidth]{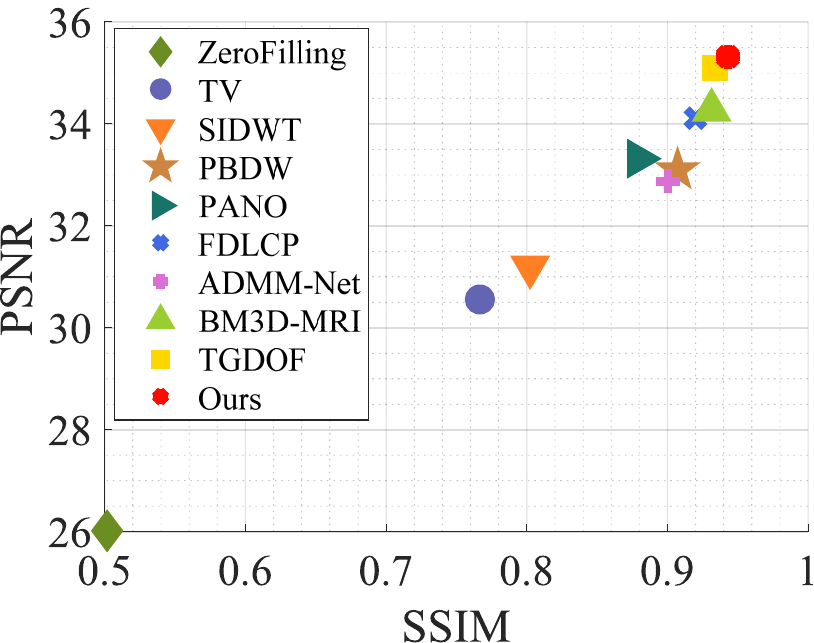}\\
		\includegraphics[width=0.31\linewidth]{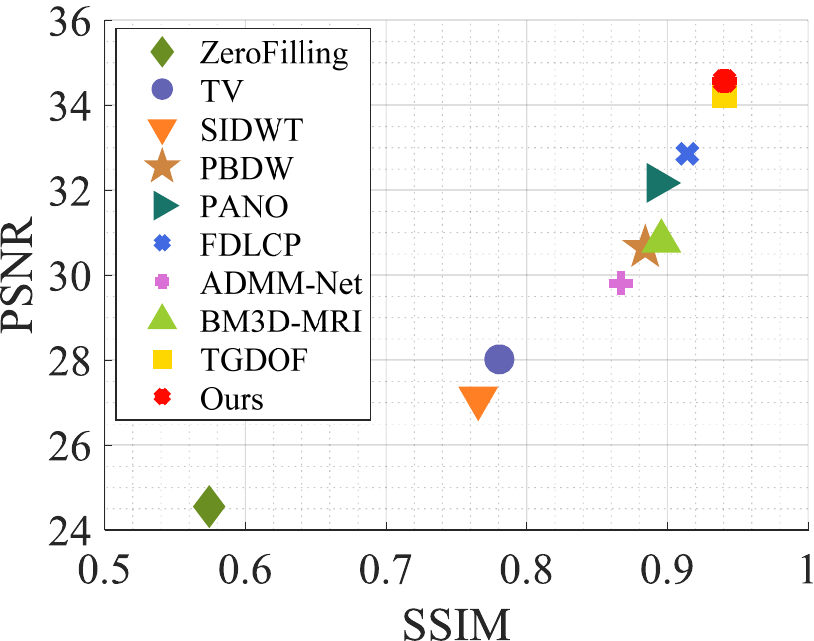}&
		\includegraphics[width=0.31\linewidth]{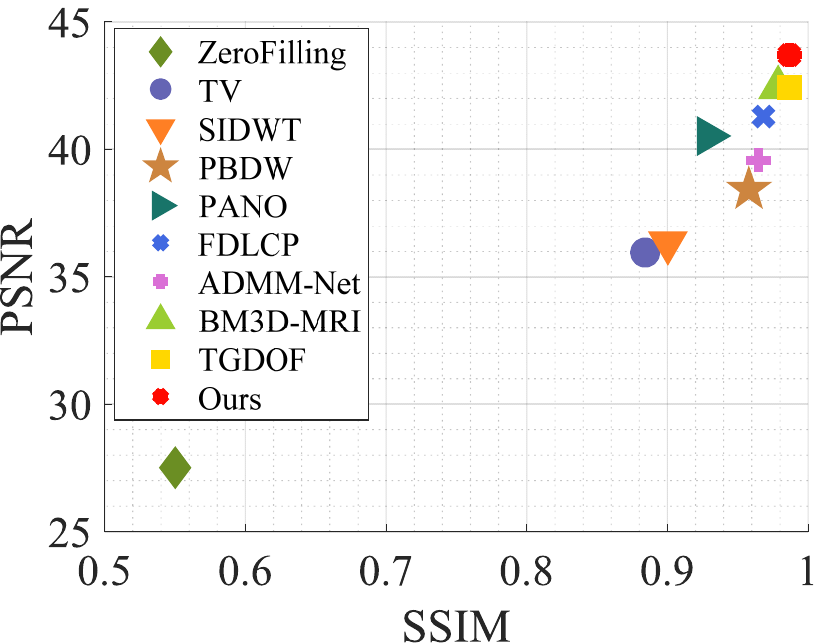}&
		\includegraphics[width=0.31\linewidth]{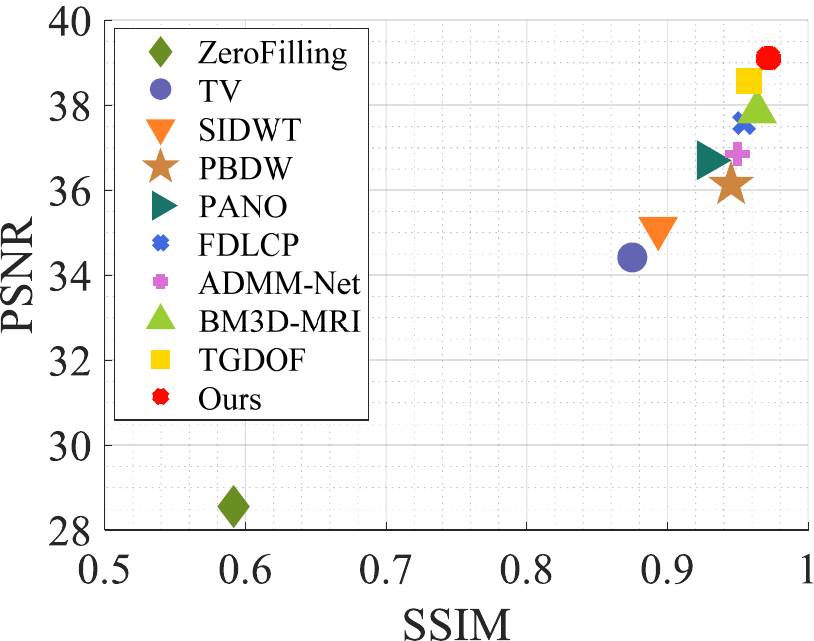}\\
		\footnotesize Cartesian mask&\footnotesize Gaussian mask&\footnotesize Radial mask\\
	\end{tabular}
	\caption{Averaged CS-MRI results on IXI dataset among state-of-the-art methods. Top row: $20\%$ sampling rate. Bottom row: $30\%$ sampling rate. In each subfigure, the upper right is the best.}
	\label{fig:CSMRIQuan}
\end{figure*}

\begin{figure*}[!htb]
	\centering
	\begin{tabular}{c@{\extracolsep{0.3em}}c@{\extracolsep{0.3em}}c@{\extracolsep{0.3em}}c@{\extracolsep{0.3em}}c@{\extracolsep{0.3em}}c@{\extracolsep{0.3em}}c}
		\includegraphics[width=0.133\linewidth]{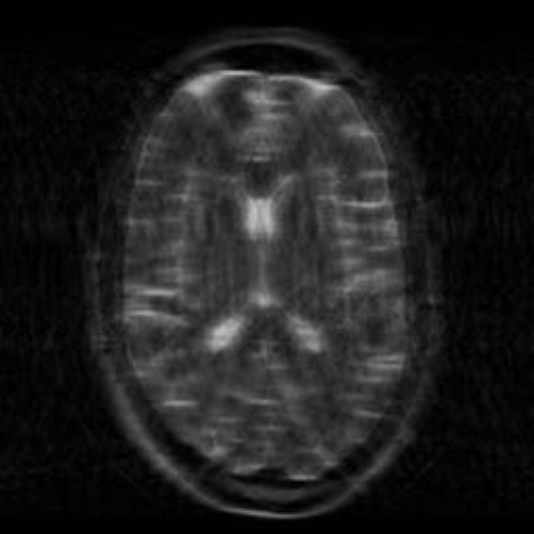}&
		\includegraphics[width=0.133\linewidth]{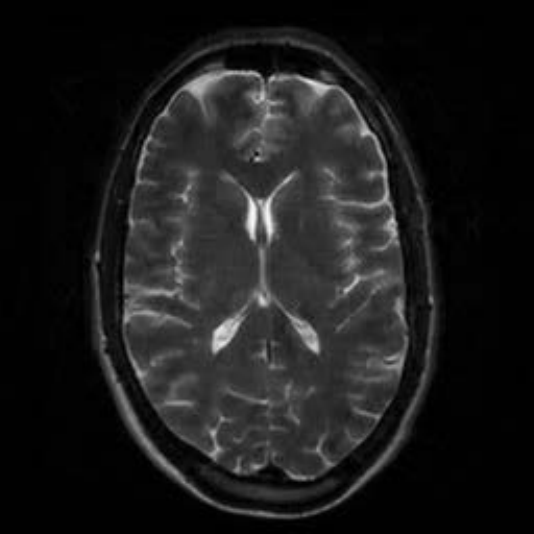}&
		\includegraphics[width=0.133\linewidth]{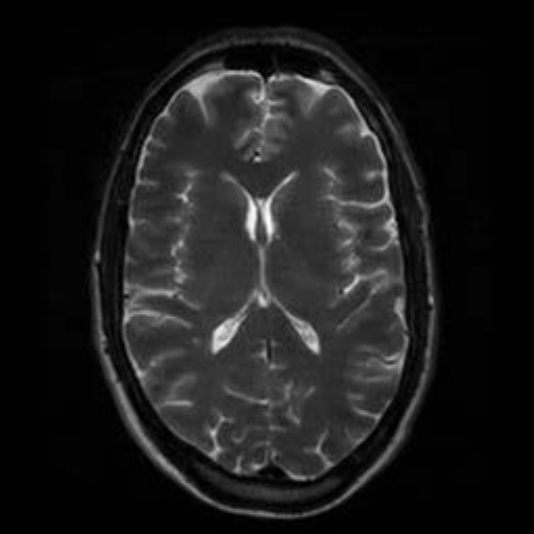}&
		\includegraphics[width=0.133\linewidth]{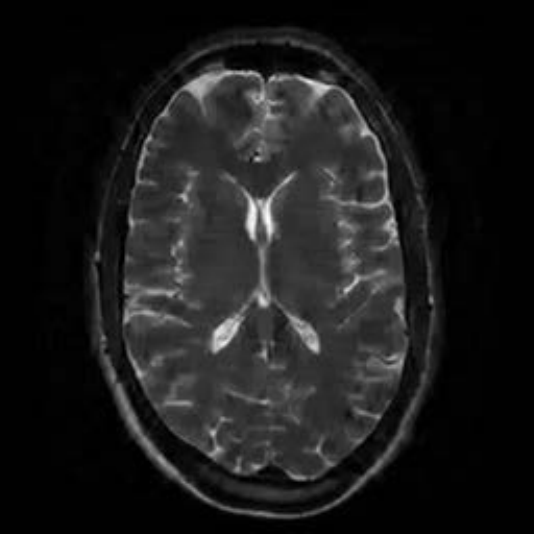}&
		\includegraphics[width=0.133\linewidth]{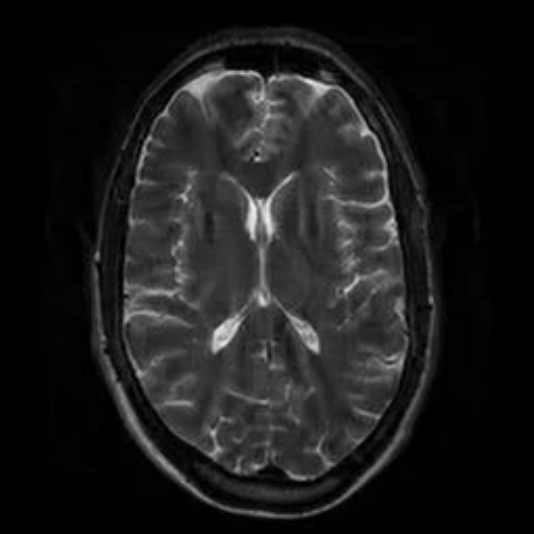}&
		\includegraphics[width=0.133\linewidth]{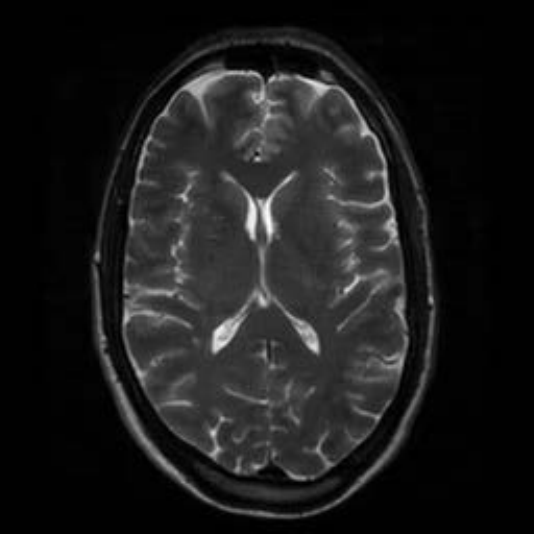}&
		\includegraphics[width=0.133\linewidth]{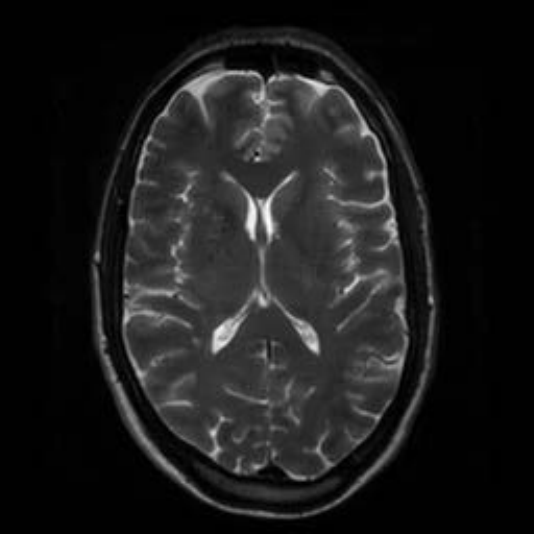}\\
		\textemdash& \footnotesize 34.73/0.9267& \footnotesize 36.82/0.9511& \footnotesize 31.97/0.9009& \footnotesize 32.45/0.9168&  \footnotesize 37.80/0.9583&\footnotesize \textbf{38.35}/\textbf{0.9635}\\
		\includegraphics[width=0.133\linewidth]{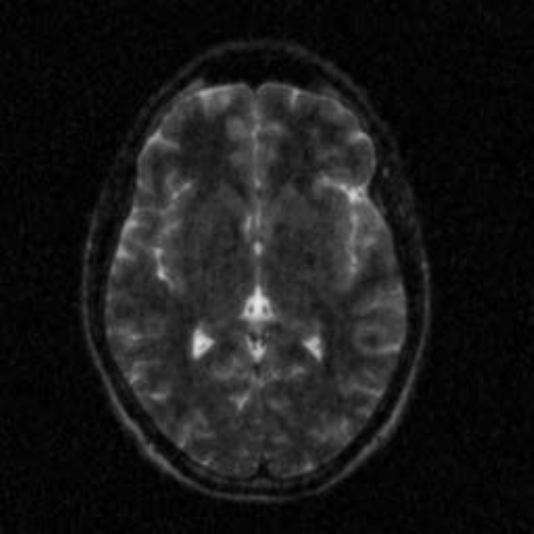}&
		\includegraphics[width=0.133\linewidth]{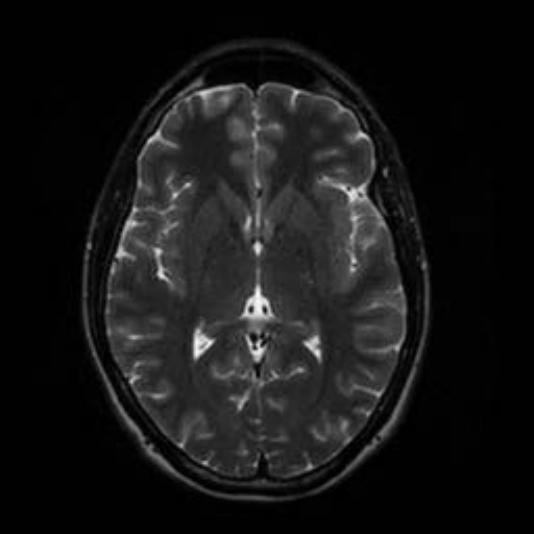}&
		\includegraphics[width=0.133\linewidth]{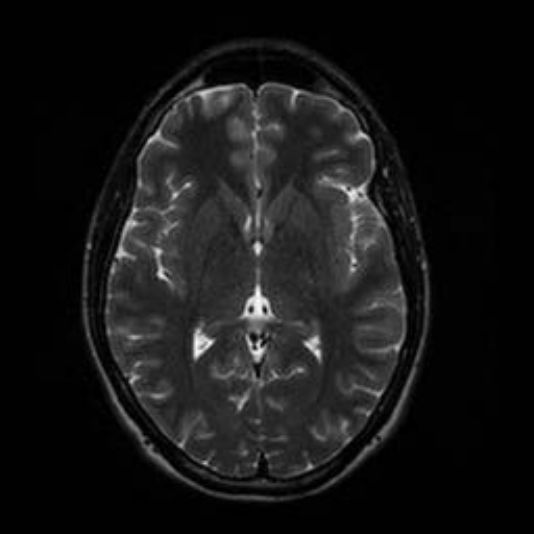}&
		\includegraphics[width=0.133\linewidth]{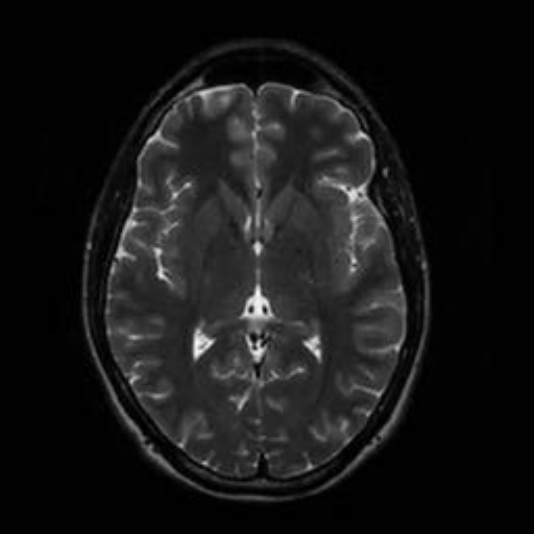}&
		\includegraphics[width=0.133\linewidth]{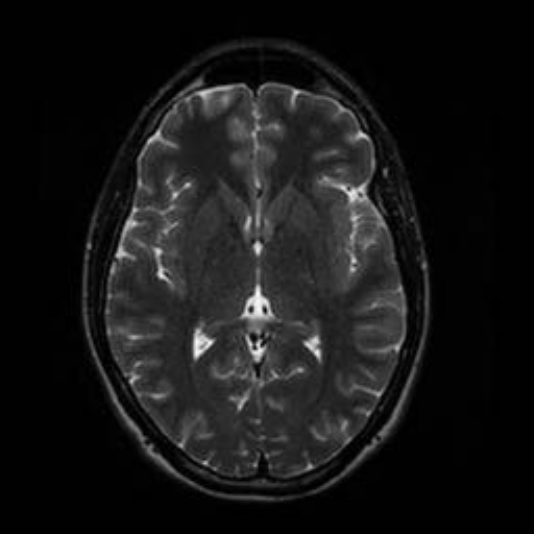}&
		\includegraphics[width=0.133\linewidth]{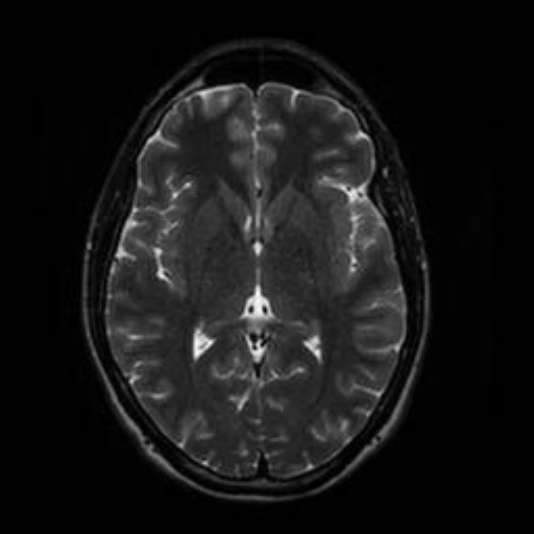}&
		\includegraphics[width=0.133\linewidth]{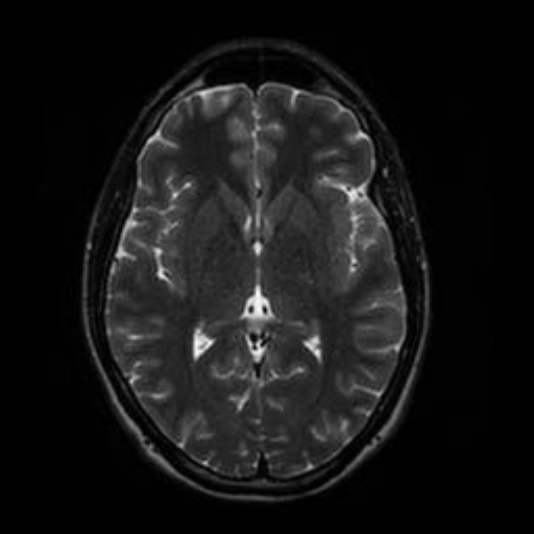}\\
		\textemdash&\footnotesize 43.66/0.9792&\footnotesize 44.12/0.9761&\footnotesize 42.39/0.9776&\footnotesize 46.31/0.9911&\footnotesize 45.83/0.9832&\footnotesize \textbf{46.88}/\textbf{0.9917}\\
		\includegraphics[width=0.133\linewidth]{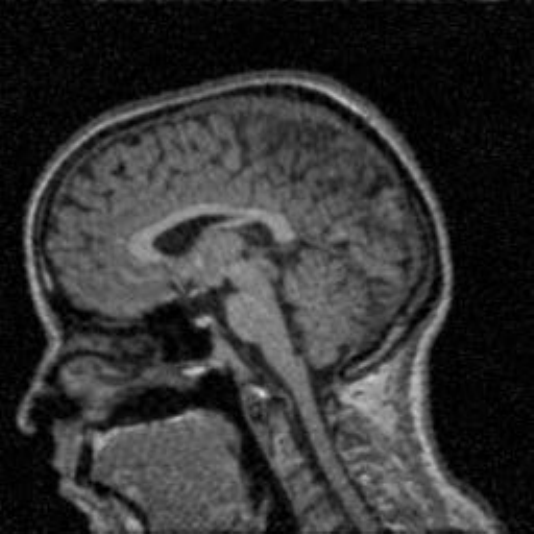}&
		\includegraphics[width=0.133\linewidth]{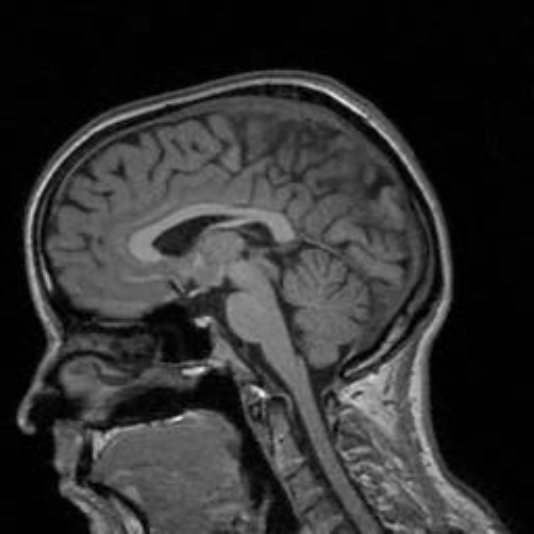}&
		\includegraphics[width=0.133\linewidth]{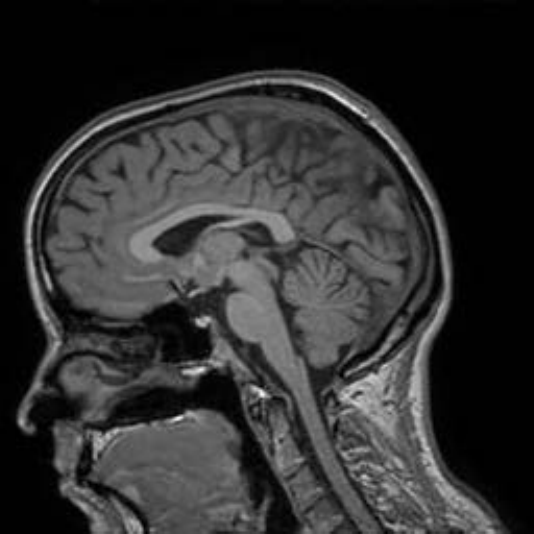}&
		\includegraphics[width=0.133\linewidth]{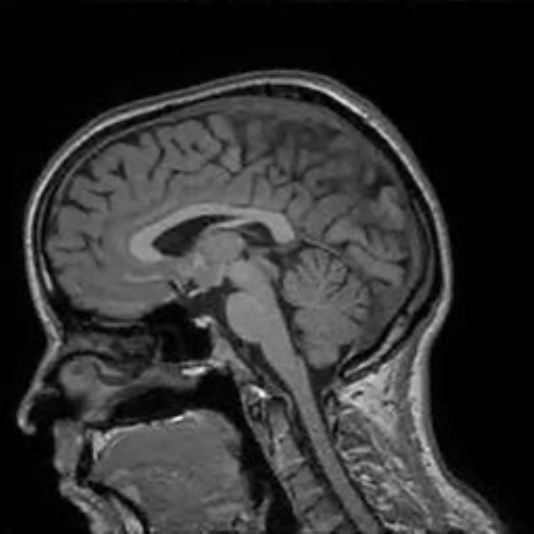}&
		\includegraphics[width=0.133\linewidth]{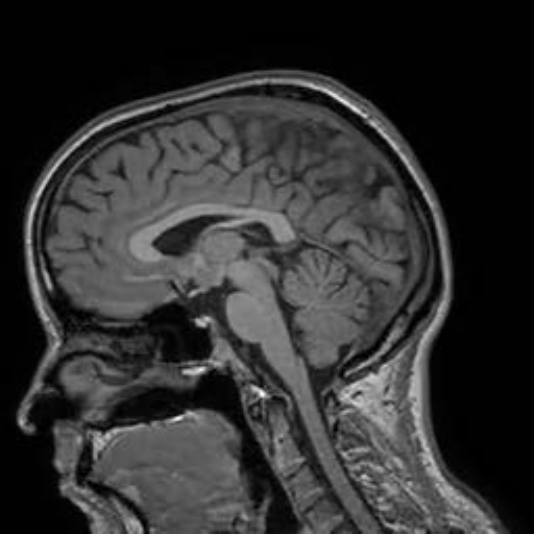}&
		\includegraphics[width=0.133\linewidth]{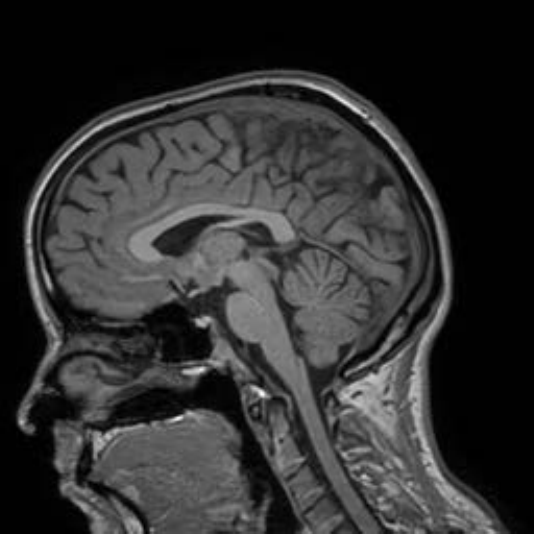}&
		\includegraphics[width=0.133\linewidth]{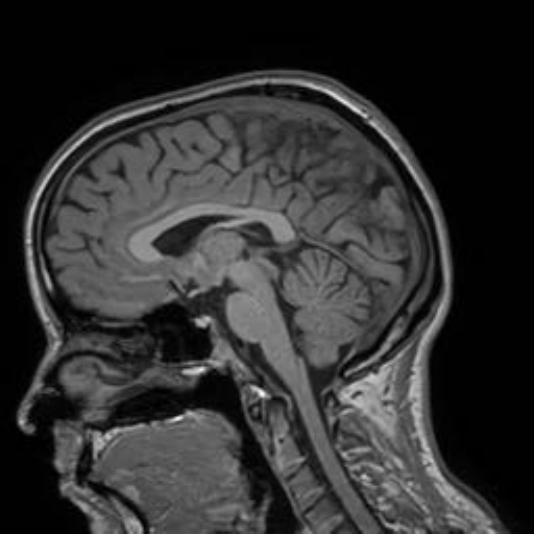}\\
		\textemdash&\footnotesize 33.80/0.8891&\footnotesize 34.85/0.9408&\footnotesize 33.79/0.9320&\footnotesize 35.02/0.9556&\footnotesize 35.92/0.9504&\footnotesize \textbf{36.04}/\textbf{0.9641}\\
		\footnotesize Zerofilling&\footnotesize PANO&\footnotesize FDLCP&\footnotesize ADMM-Net&\footnotesize BM3D-MRI&\footnotesize TGDOF& \footnotesize TOLF (Ours)\\
	\end{tabular}	
	\caption{Visual comparison among state-of-the-art methods of Compressive Sensing MRI at the sparse $k$-space data with different undersampling patterns and at a 30\% sampling rates (Top row: Cartesian mask. Middle row: Gaussian mask. Bottom row: Radial mask). }
	\label{fig:CSMRI1}
\end{figure*}	

\begin{figure*}[t]
	\centering
	\begin{tabular}{c@{\extracolsep{0.3em}}c@{\extracolsep{0.3em}}c@{\extracolsep{0.3em}}c@{\extracolsep{0.3em}}c@{\extracolsep{0.3em}}c@{\extracolsep{0.3em}}c}
		\includegraphics[width=0.133\linewidth]{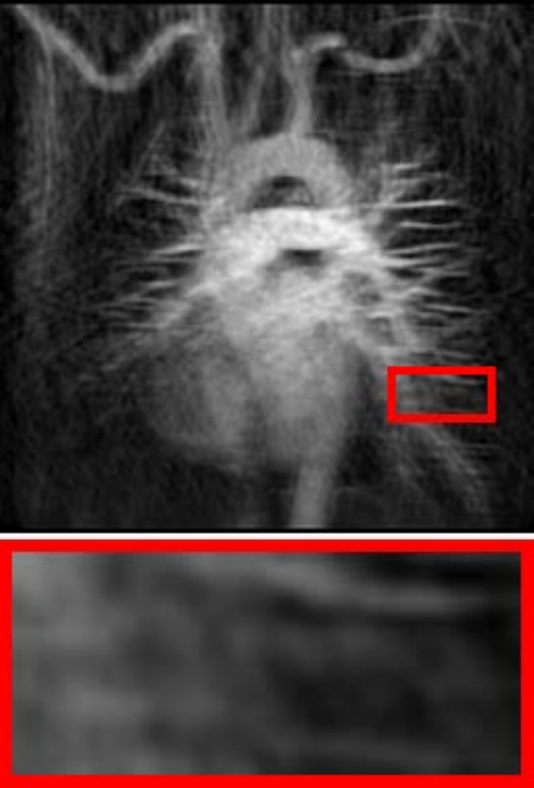}&
		\includegraphics[width=0.133\linewidth]{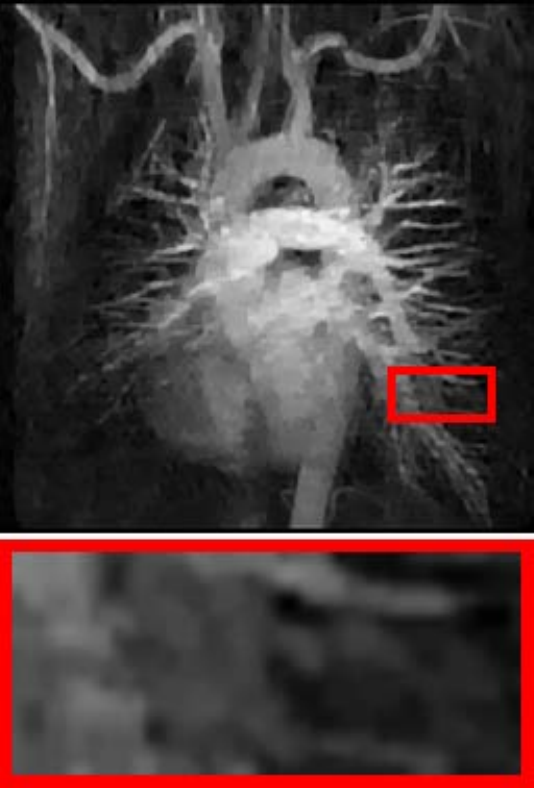}&
		\includegraphics[width=0.133\linewidth]{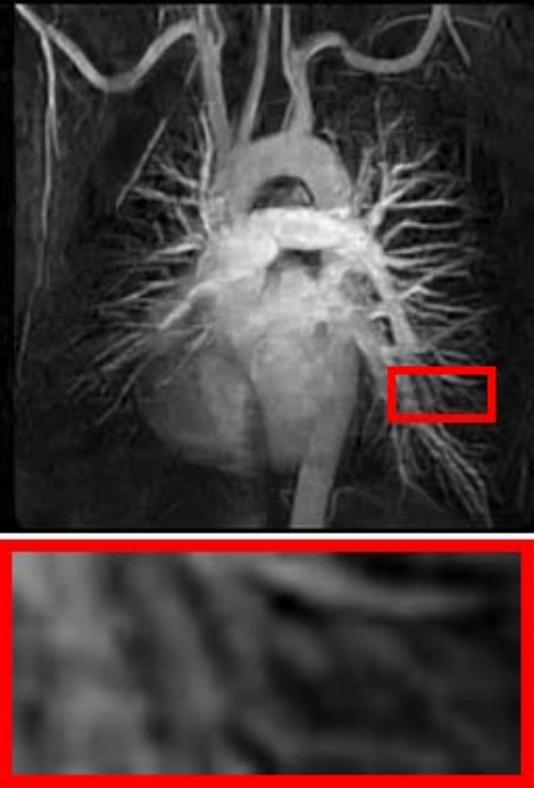}&
		\includegraphics[width=0.133\linewidth]{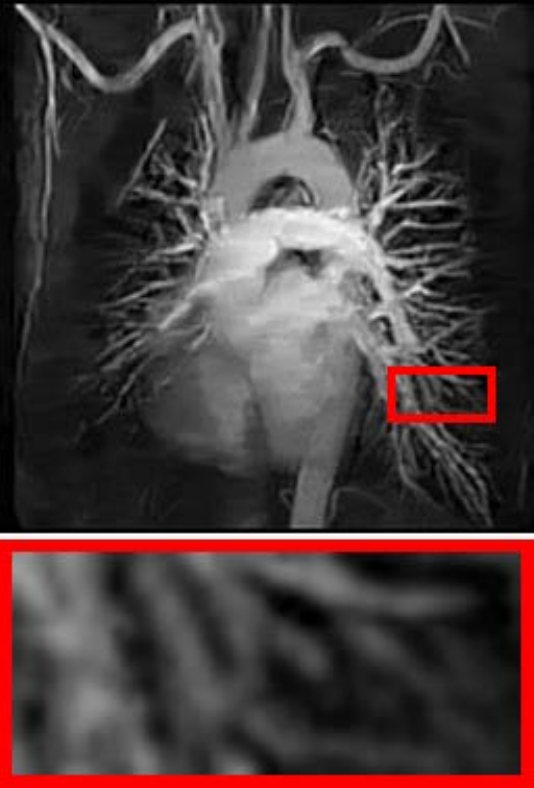}&
		\includegraphics[width=0.133\linewidth]{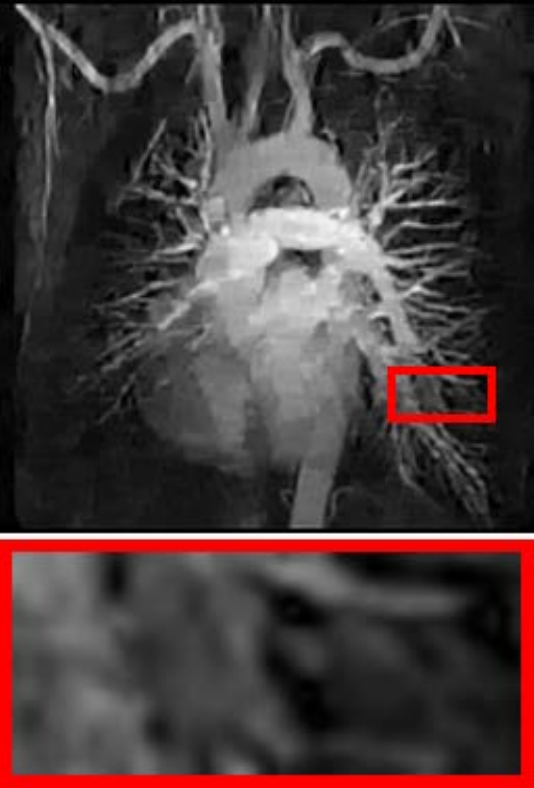}&
		\includegraphics[width=0.133\linewidth]{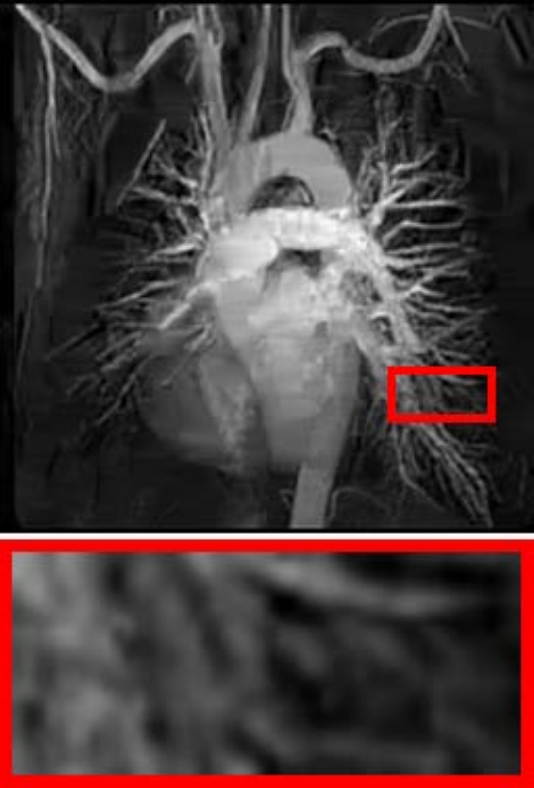}&
		\includegraphics[width=0.133\linewidth]{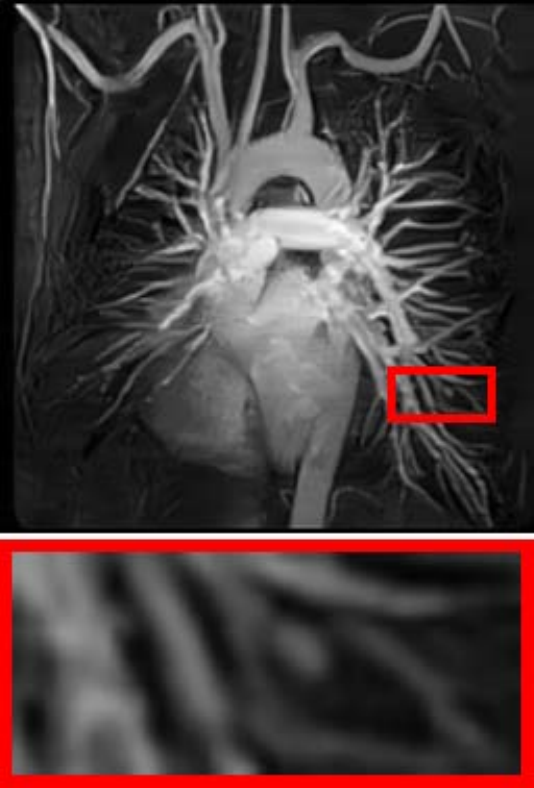}\\	
		\footnotesize PSNR/SSIM& \footnotesize 24.42/0.7024&\footnotesize 27.72/0.8155& \footnotesize 26.84/0.7929& \footnotesize 25.38/0.7488& \footnotesize 26.26/0.7740& \footnotesize \textbf{28.78/0.8440}\\
		\footnotesize Zerofilling& \footnotesize TV&\footnotesize PANO&\footnotesize FDLCP& \footnotesize ADMM-Net& \footnotesize BM3D-MRI& \footnotesize TOLF\\
	\end{tabular}
	\caption{CS-MRI results on chest data with Cartesian mask (30\% sampling rate).}
	\label{fig:CSMRI}
\end{figure*}

\begin{figure*}[t]
	\centering
	\begin{tabular}{c@{\extracolsep{0.3em}}c@{\extracolsep{0.3em}}c@{\extracolsep{0.3em}}c@{\extracolsep{0.3em}}c@{\extracolsep{0.3em}}c}
		\includegraphics[width=0.157\linewidth]{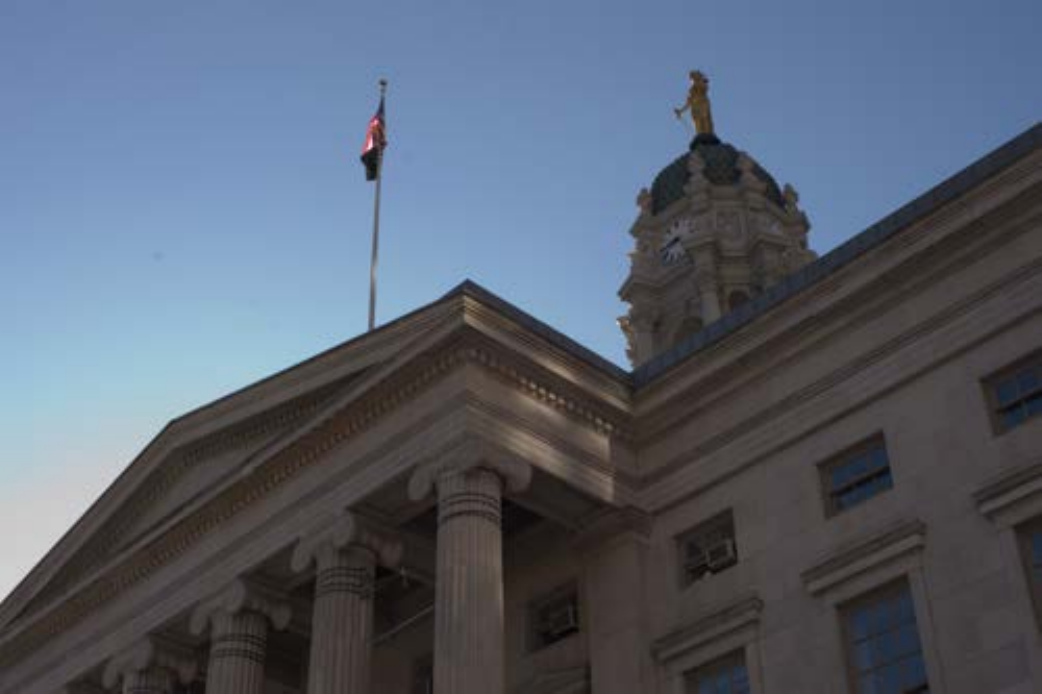}&
		\includegraphics[width=0.157\linewidth]{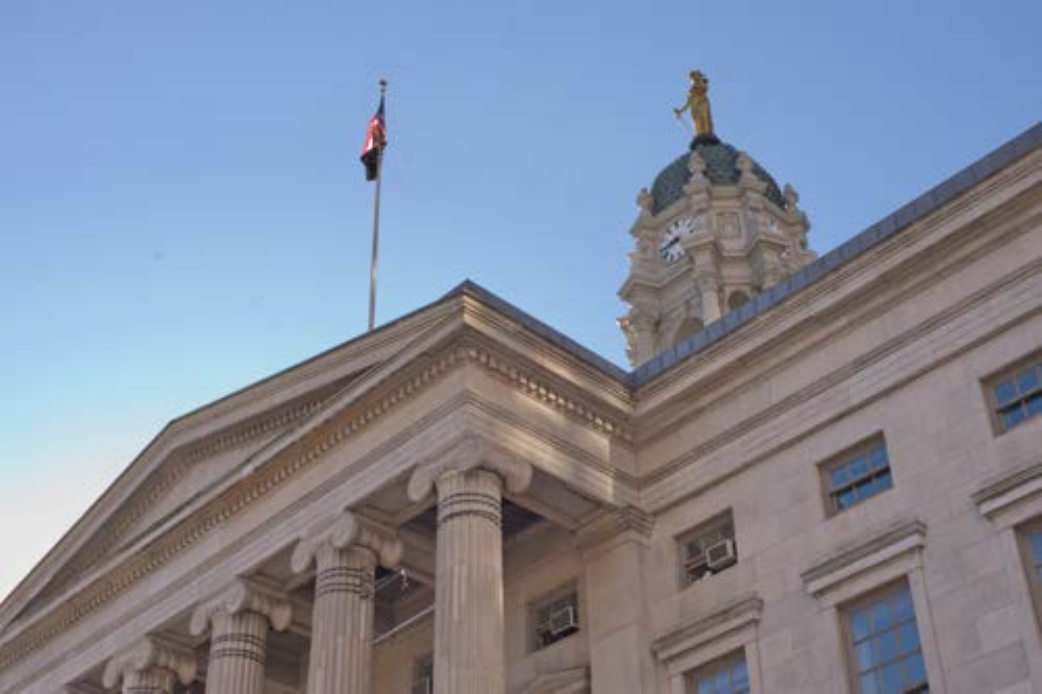}&	
		\includegraphics[width=0.157\linewidth]{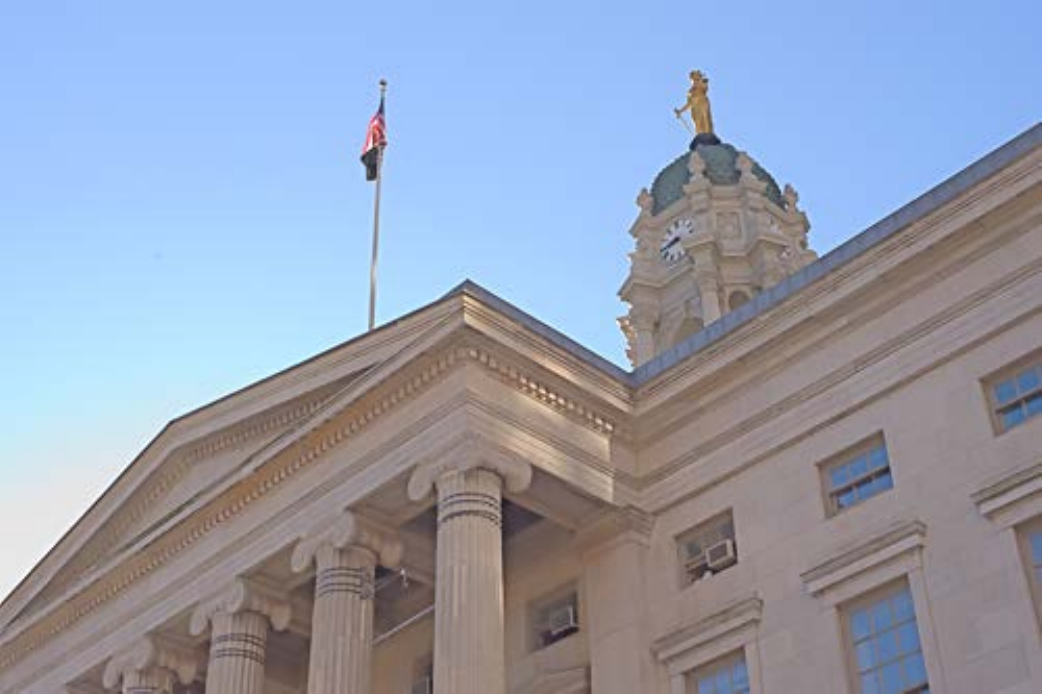}&
		\includegraphics[width=0.157\linewidth]{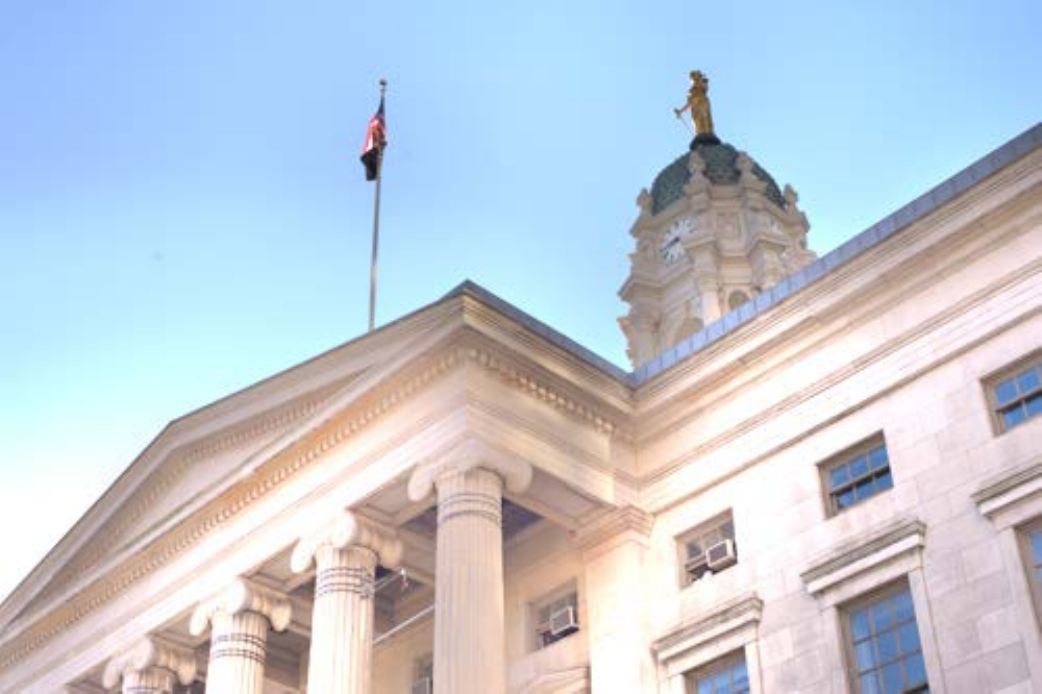}&
		\includegraphics[width=0.157\linewidth]{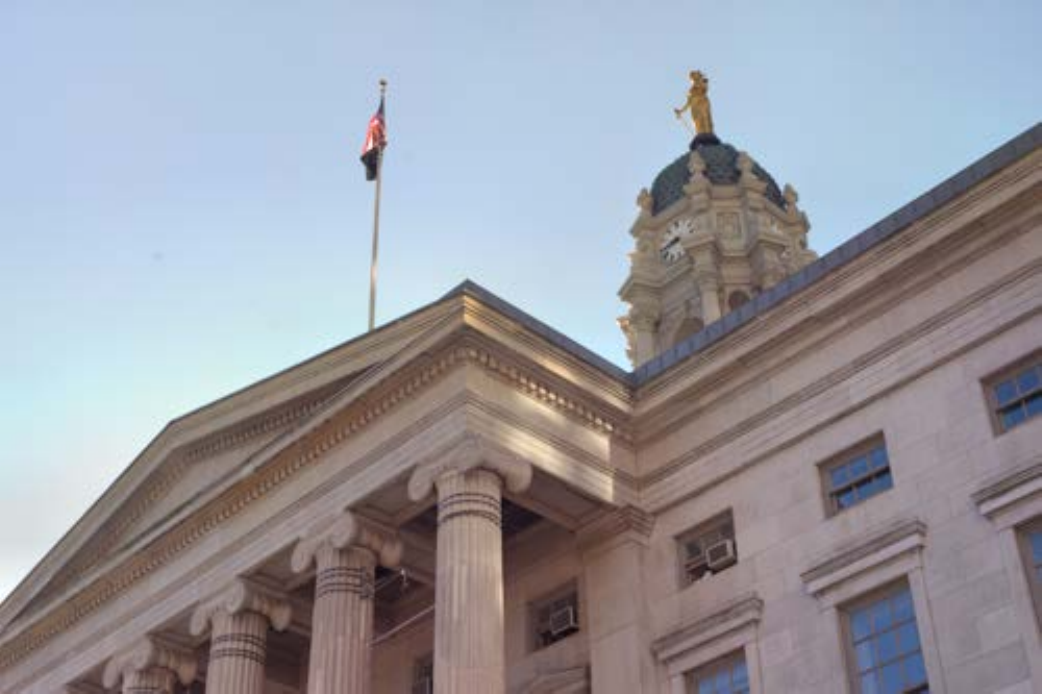}&
		\includegraphics[width=0.157\linewidth]{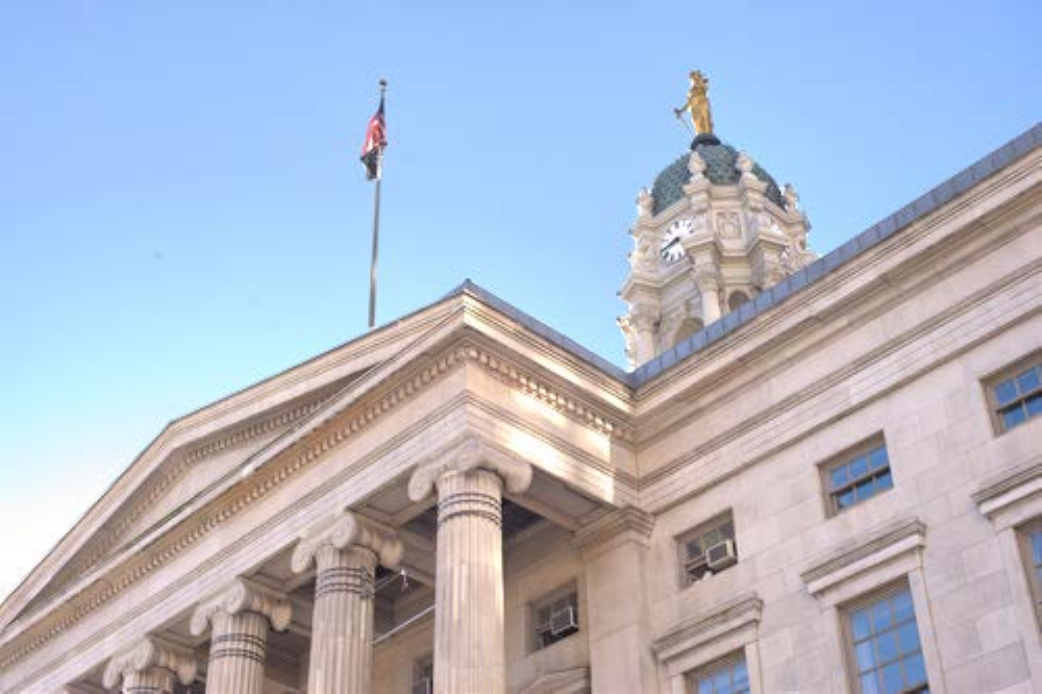}\\
		\includegraphics[width=0.157\linewidth]{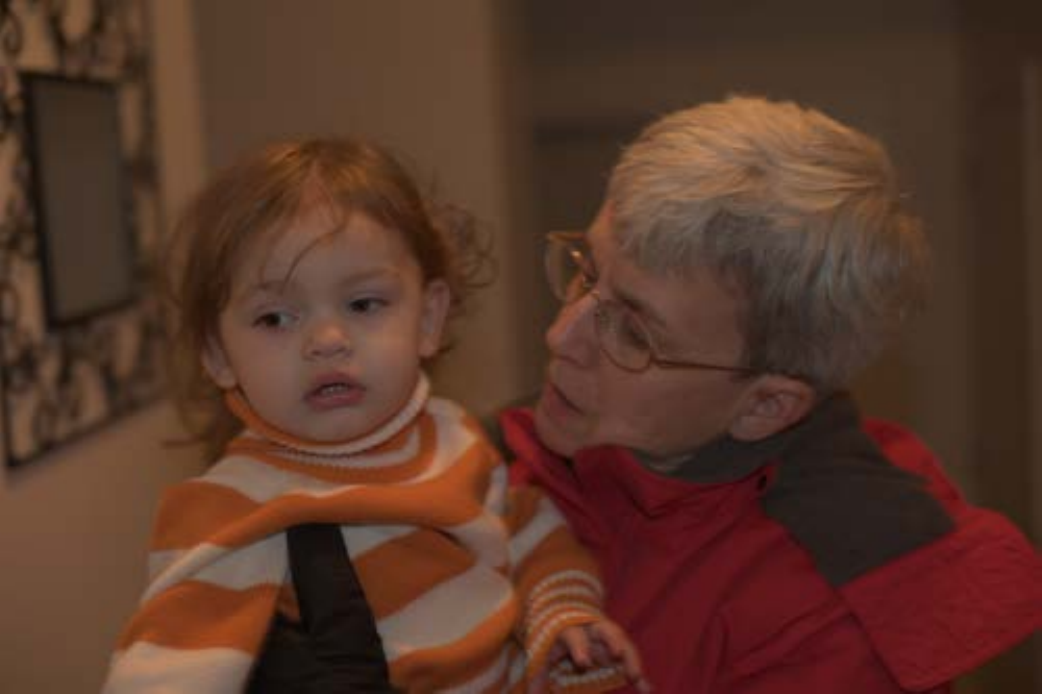}&
		\includegraphics[width=0.157\linewidth]{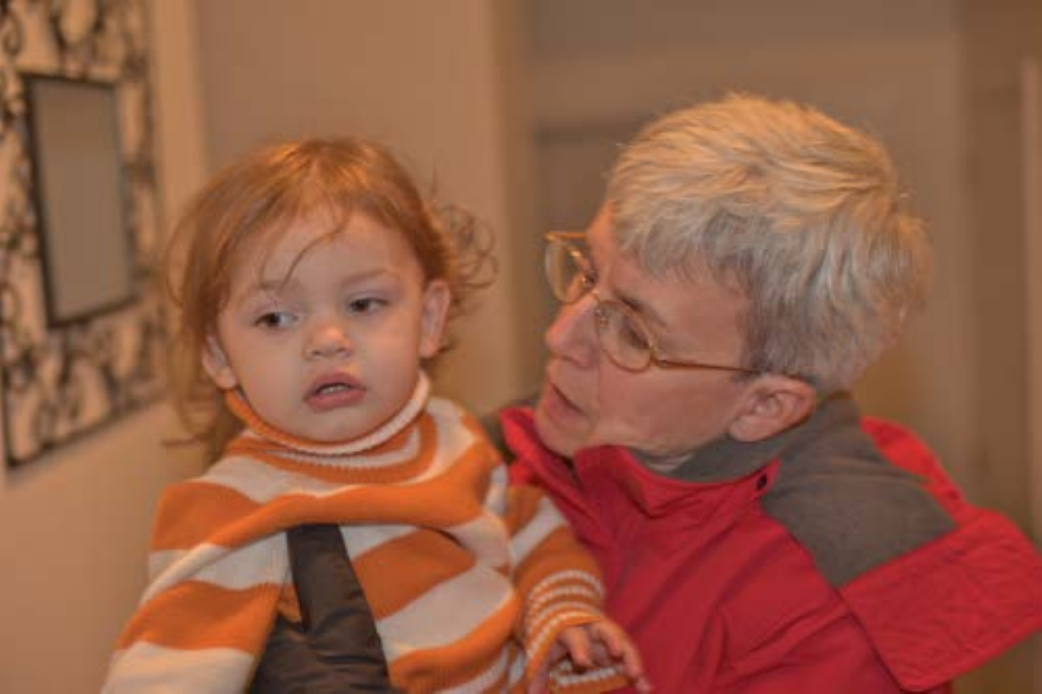}&	
		\includegraphics[width=0.157\linewidth]{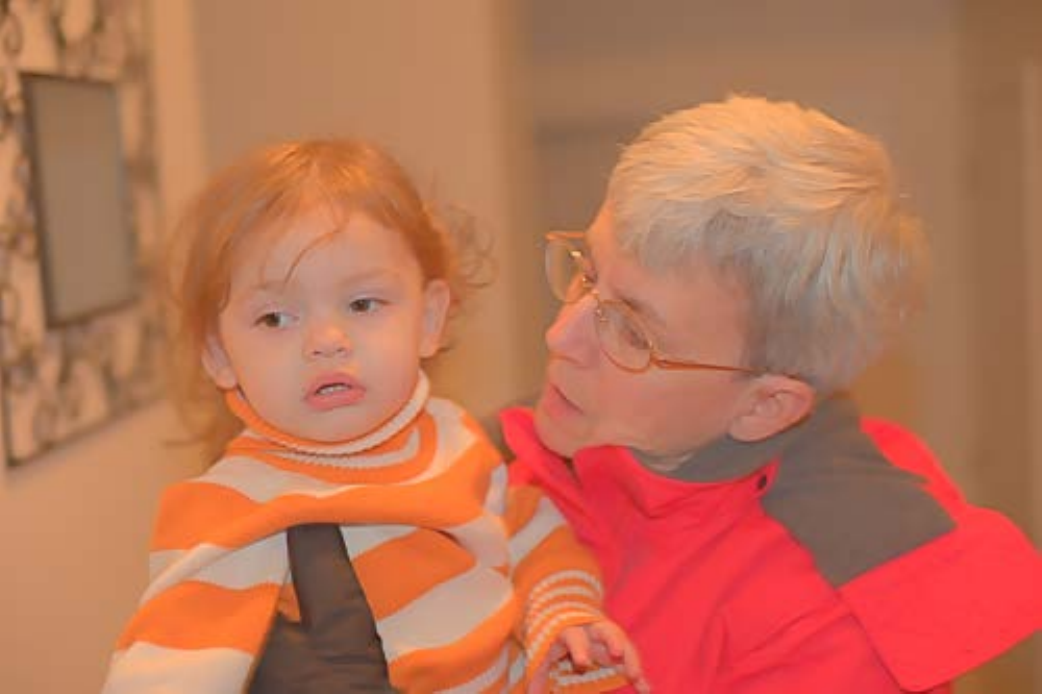}&
		\includegraphics[width=0.157\linewidth]{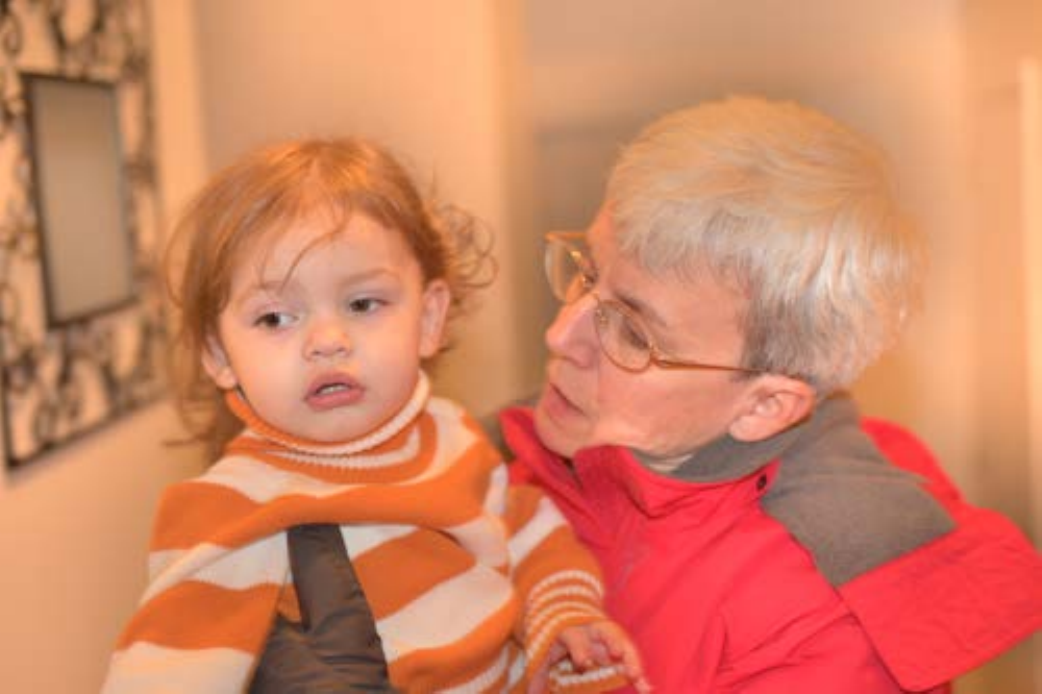}&
		\includegraphics[width=0.157\linewidth]{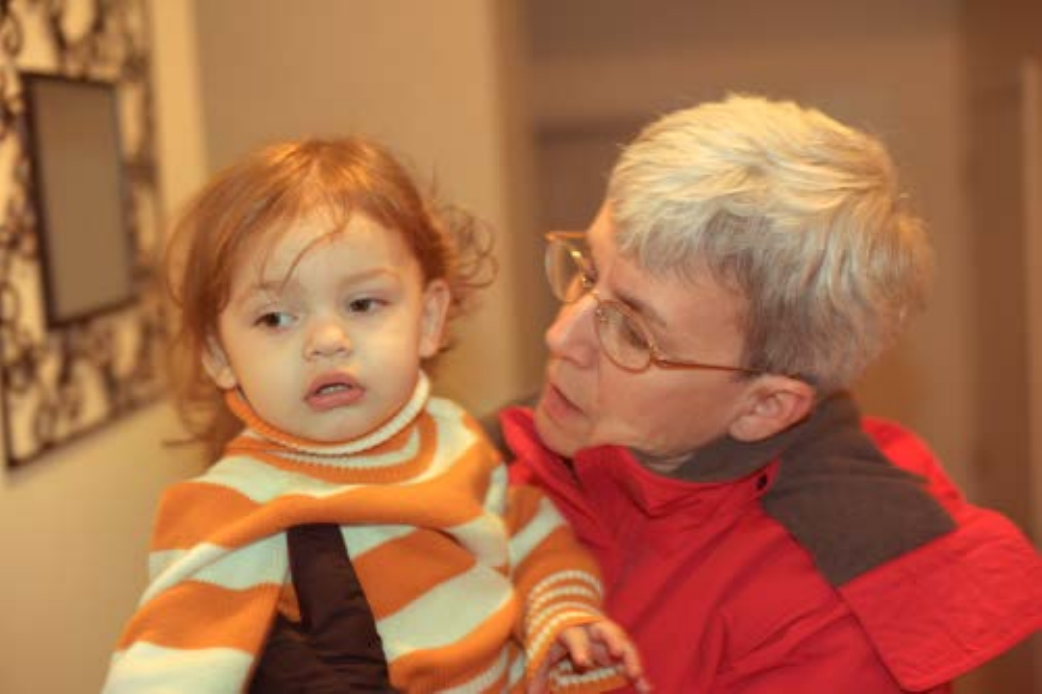}&
		\includegraphics[width=0.157\linewidth]{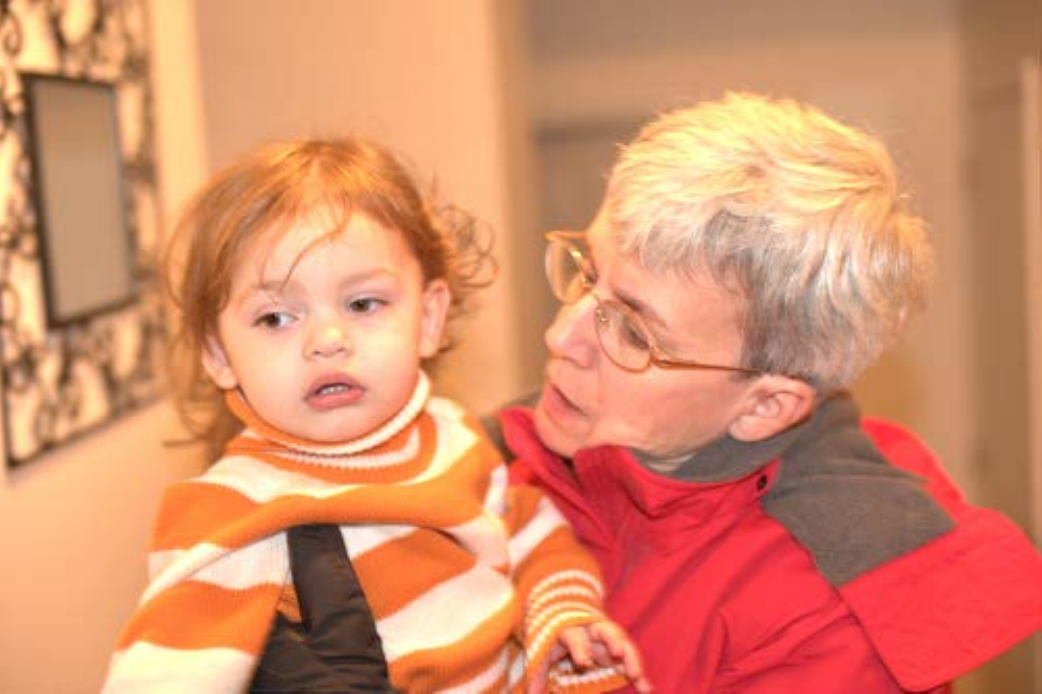}\\
		\footnotesize Input&\footnotesize JIEP&\footnotesize RRM&\footnotesize LightenNet&\footnotesize DeepUPE&\footnotesize TOLF\\
	\end{tabular}
	\caption{Illustrating visual results on testing images collected by~\cite{wang2019underexposed}.}
	\label{fig:LLIE_MIT}
\end{figure*}

\begin{figure*}[t]
	\centering
	\begin{tabular}{c@{\extracolsep{0.3em}}c@{\extracolsep{0.3em}}c@{\extracolsep{0.3em}}c@{\extracolsep{0.3em}}c@{\extracolsep{0.3em}}c}
		\includegraphics[width=0.157\linewidth]{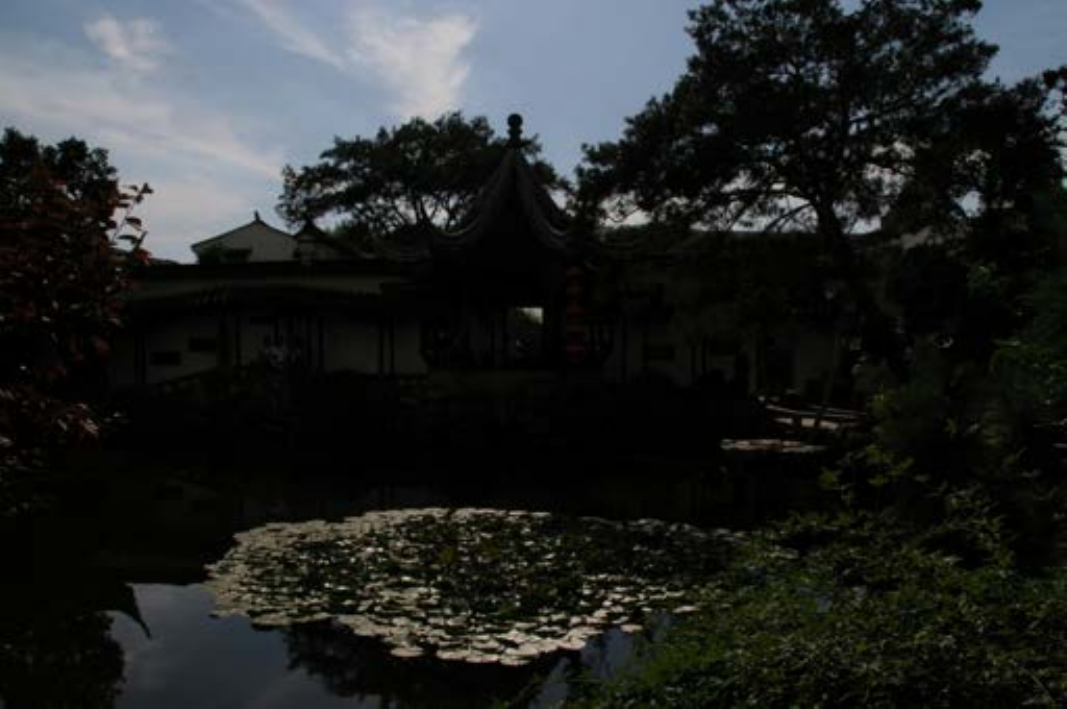}&
		\includegraphics[width=0.157\linewidth]{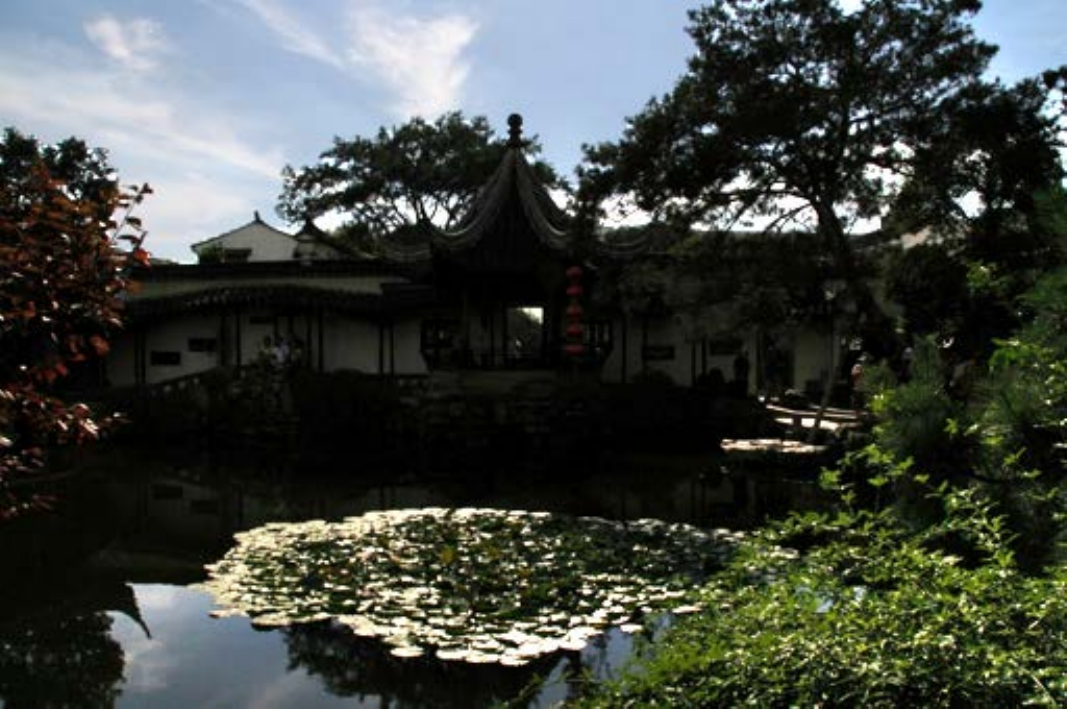}&	
		\includegraphics[width=0.157\linewidth]{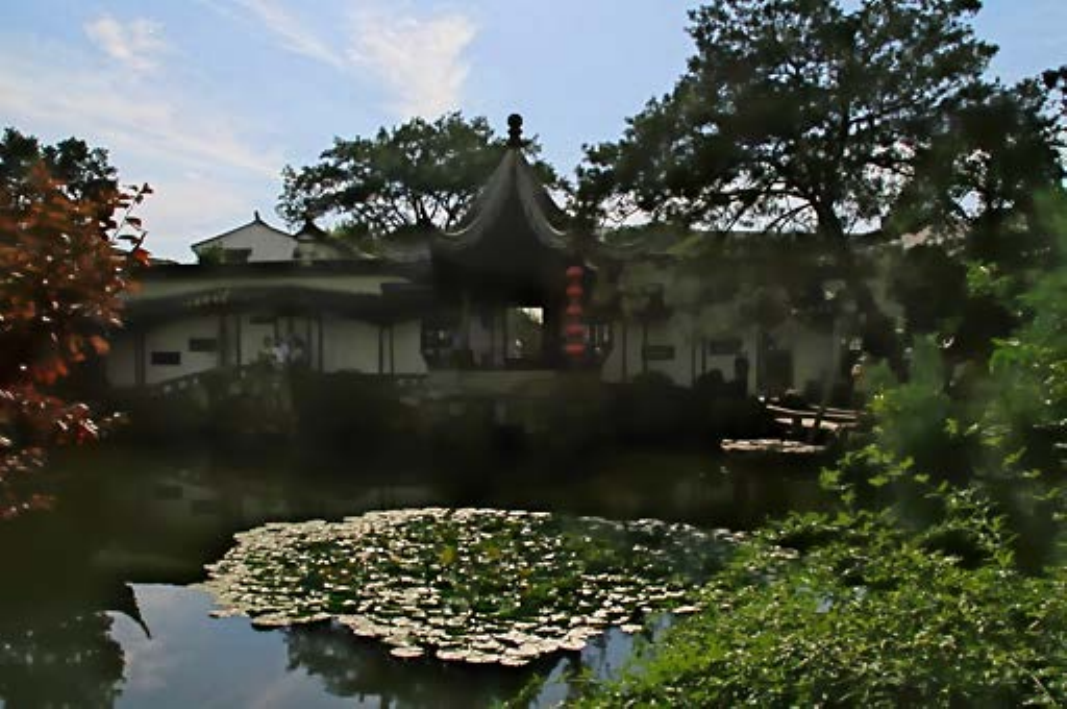}&
		\includegraphics[width=0.157\linewidth]{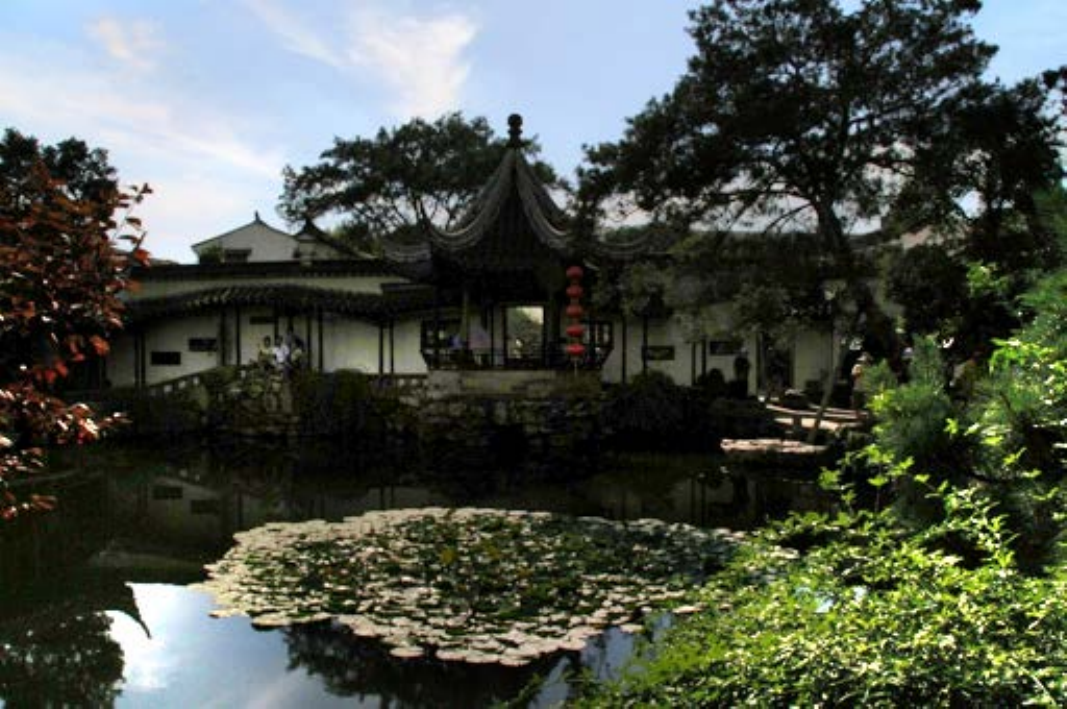}&
		\includegraphics[width=0.157\linewidth]{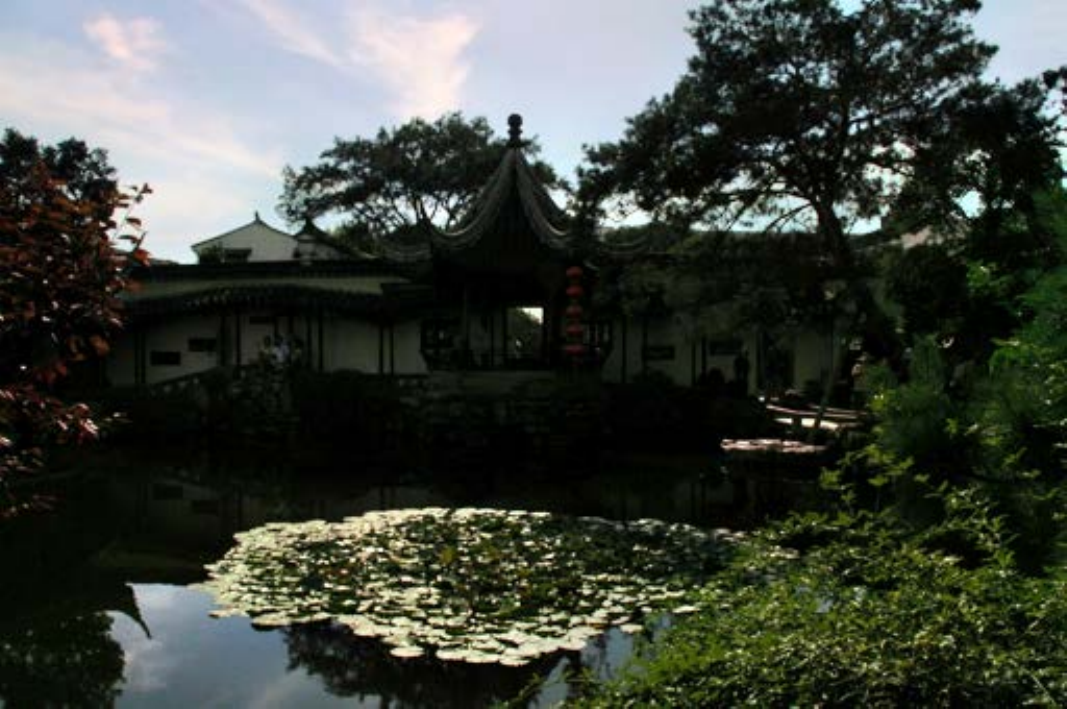}&
		\includegraphics[width=0.157\linewidth]{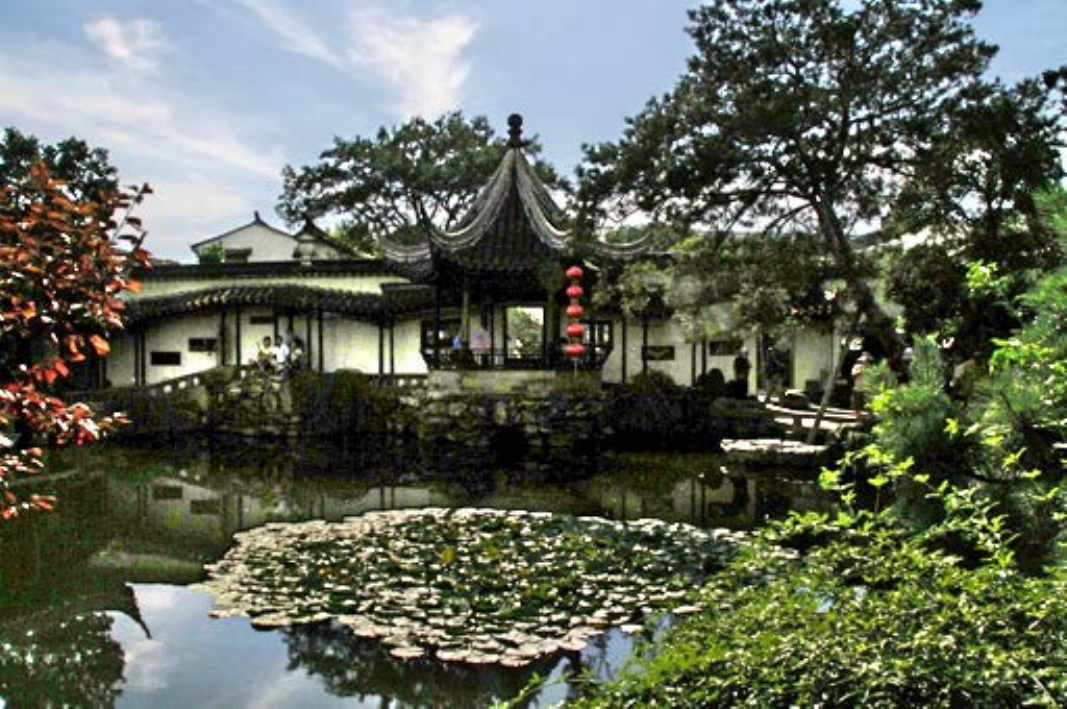}\\
		\includegraphics[width=0.157\linewidth]{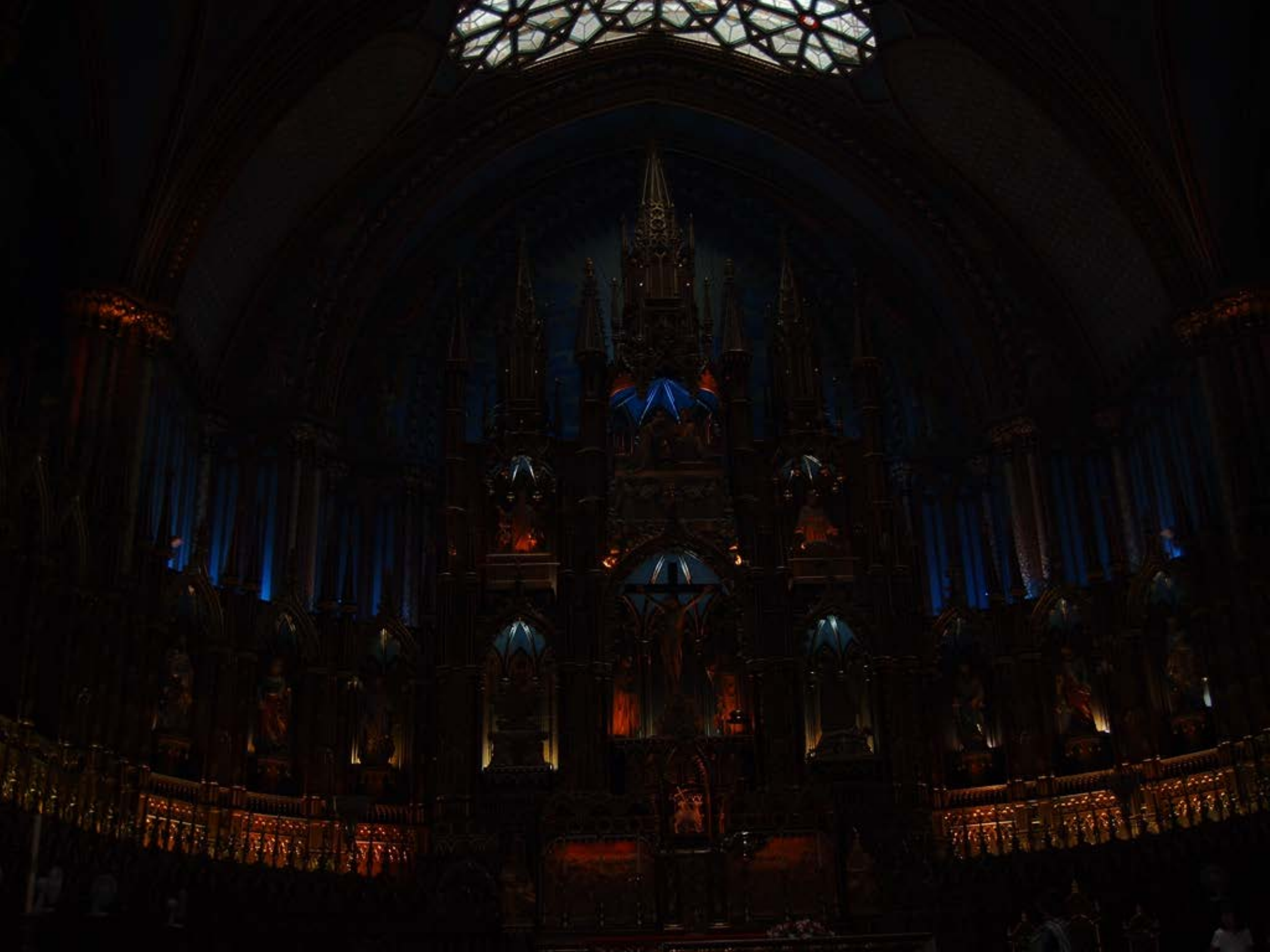}&
		\includegraphics[width=0.157\linewidth]{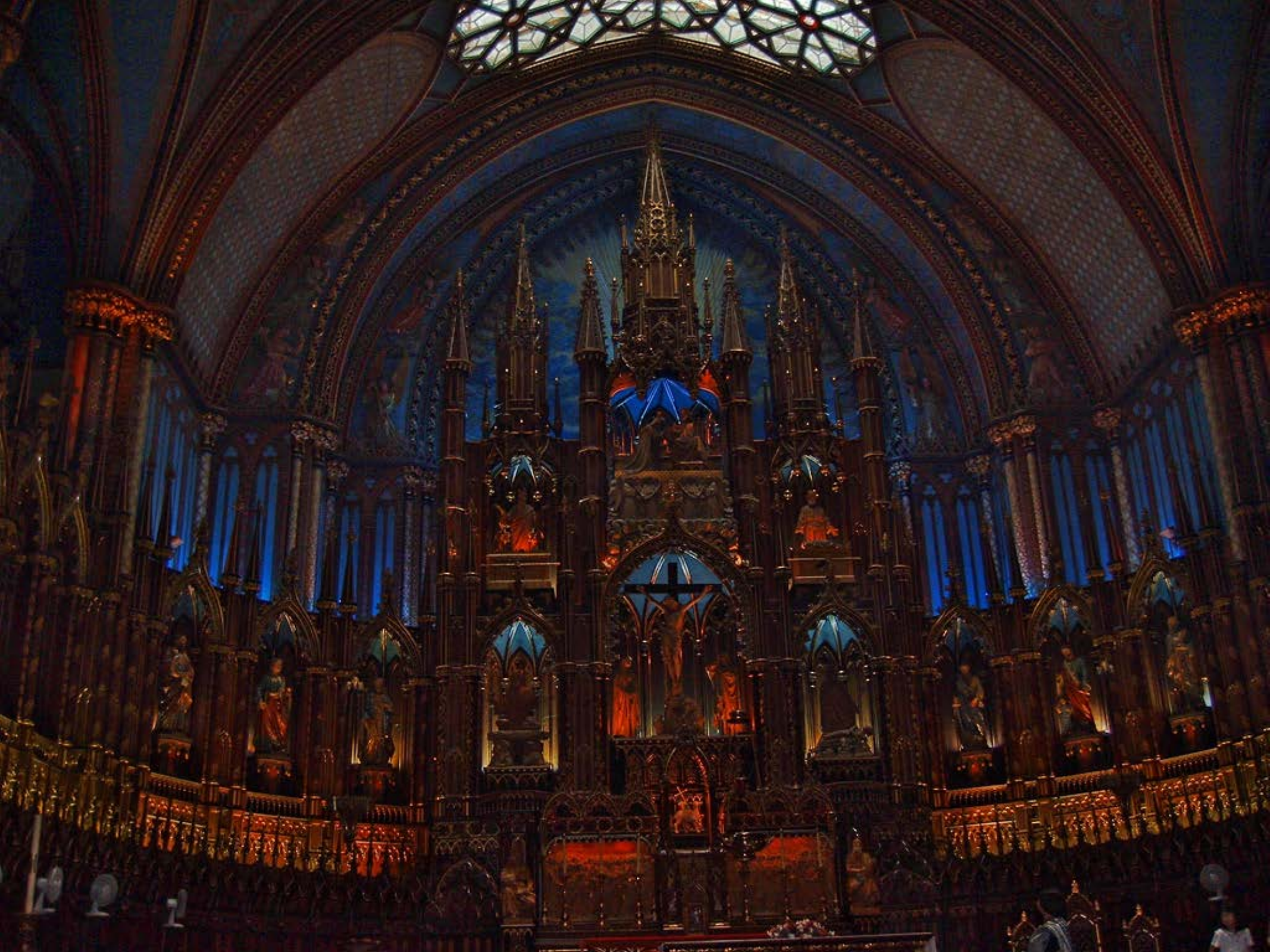}&	
		\includegraphics[width=0.157\linewidth]{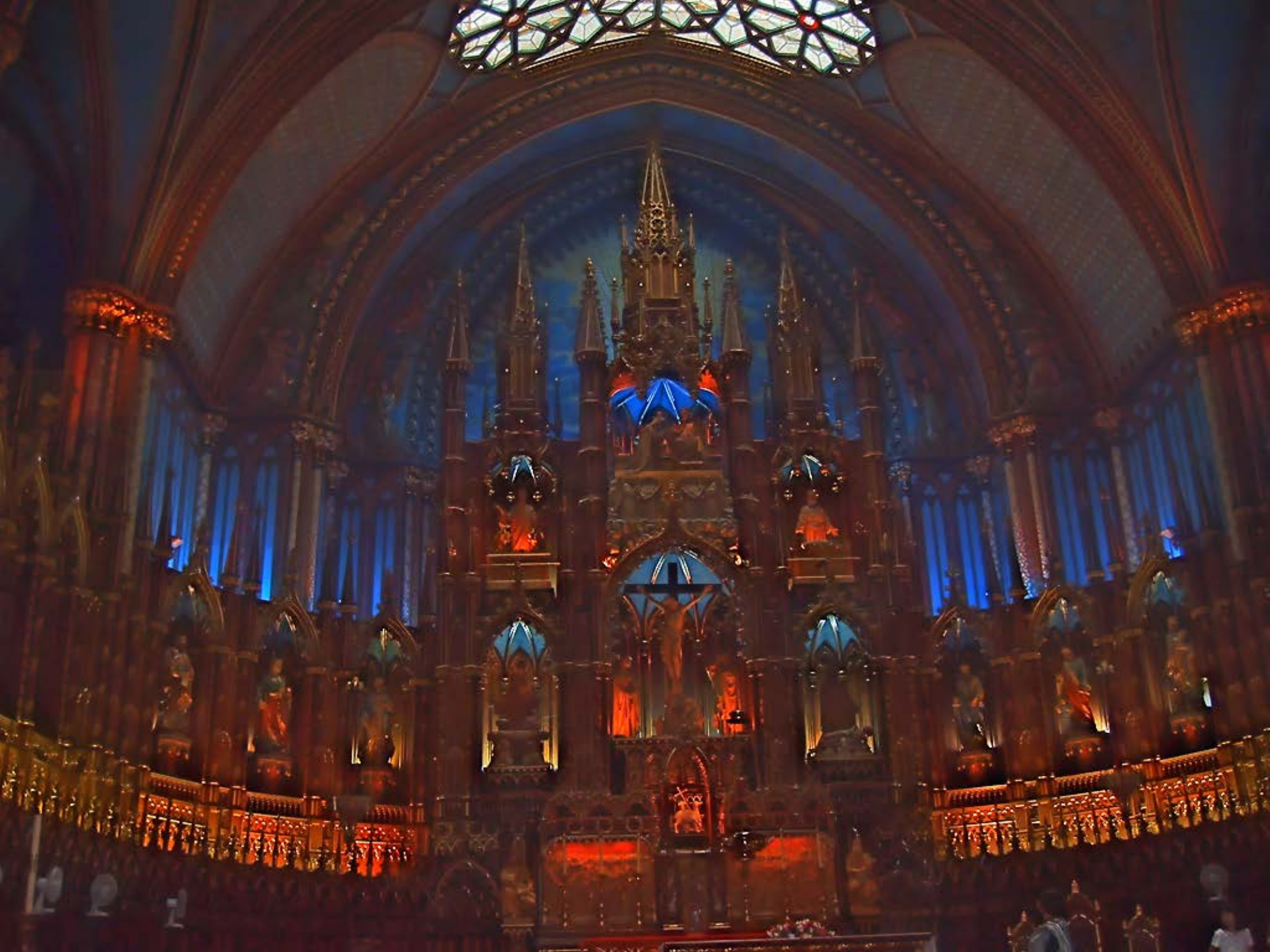}&
		\includegraphics[width=0.157\linewidth]{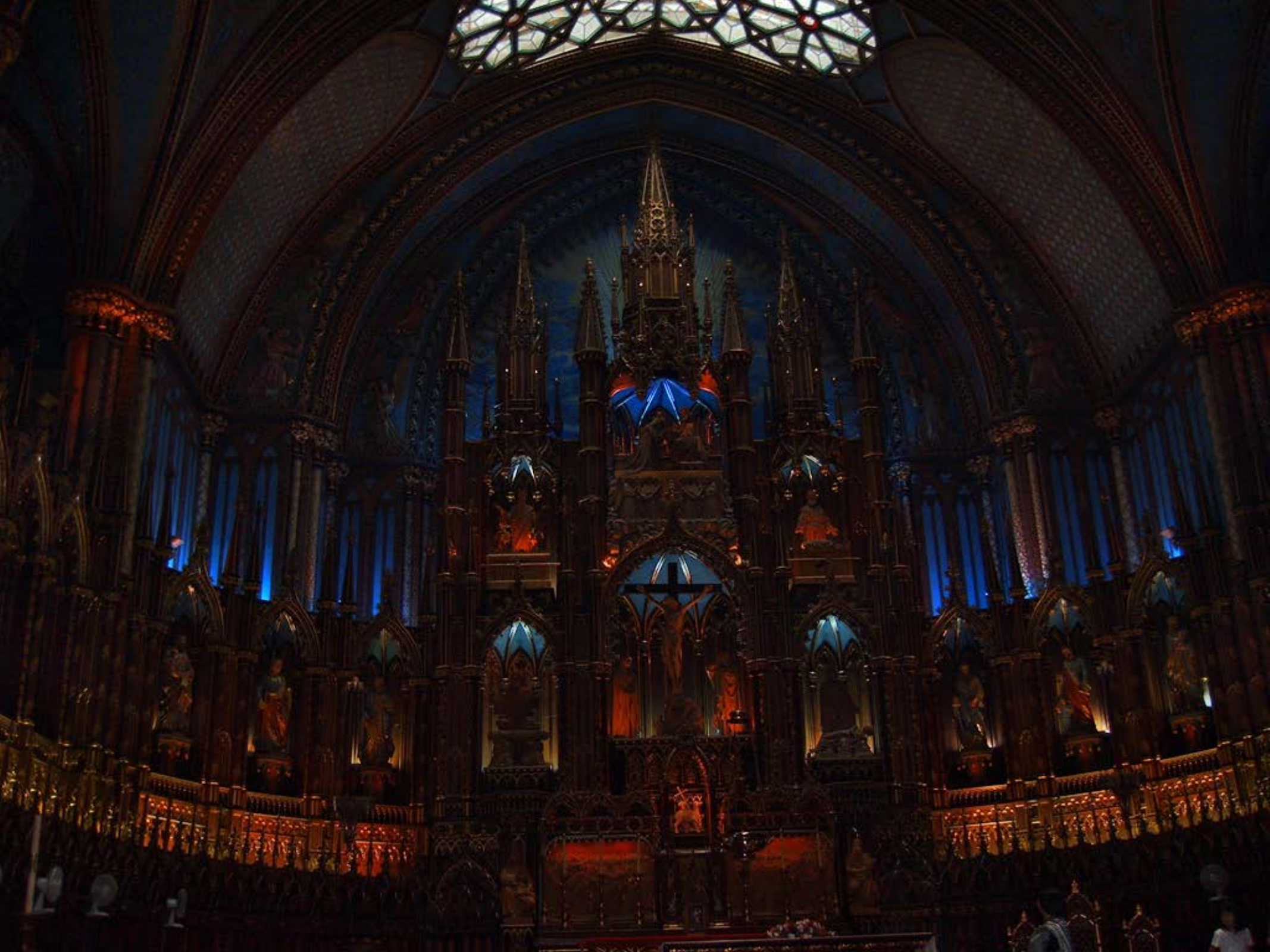}&
		\includegraphics[width=0.157\linewidth]{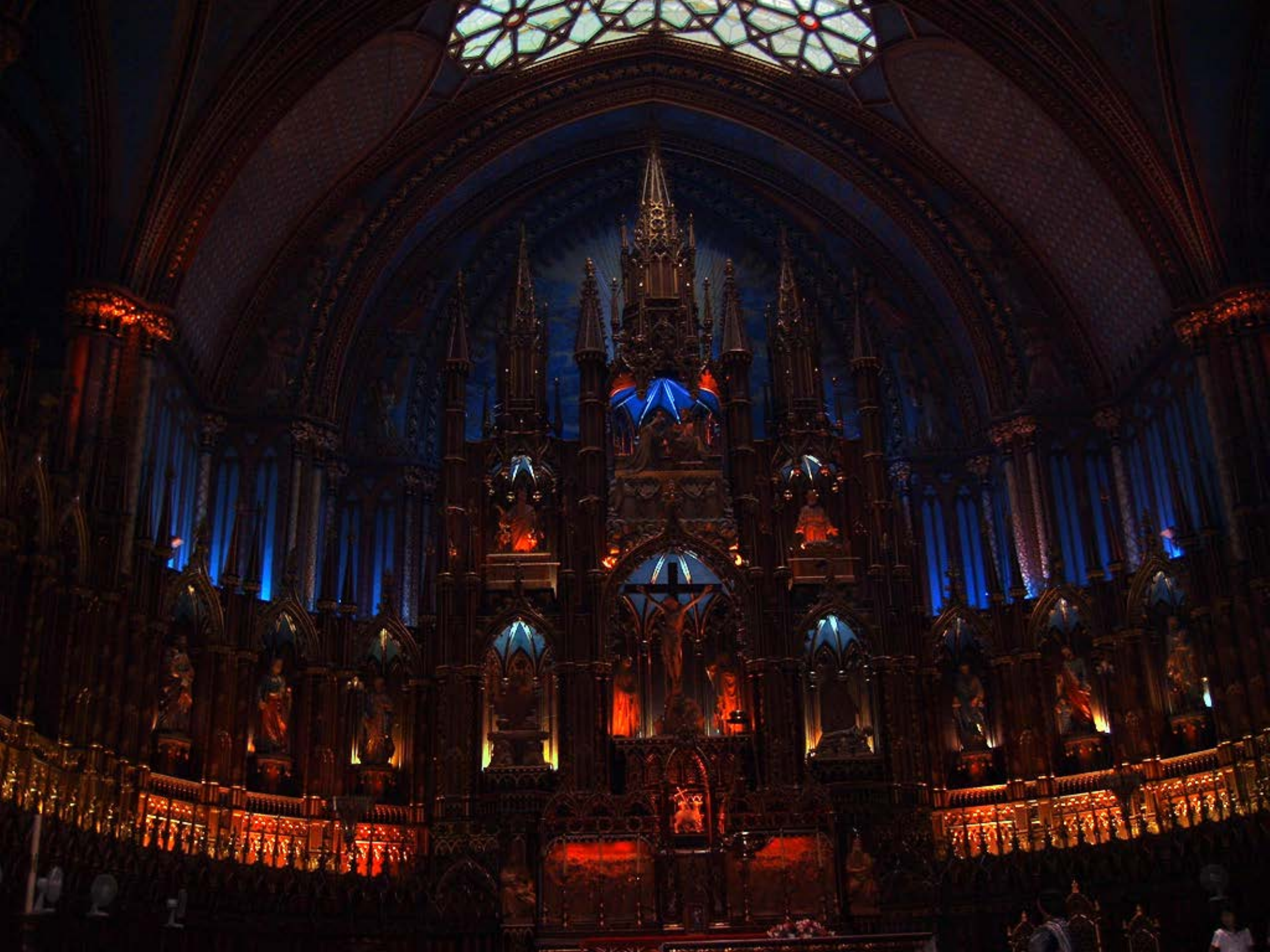}&
		\includegraphics[width=0.157\linewidth]{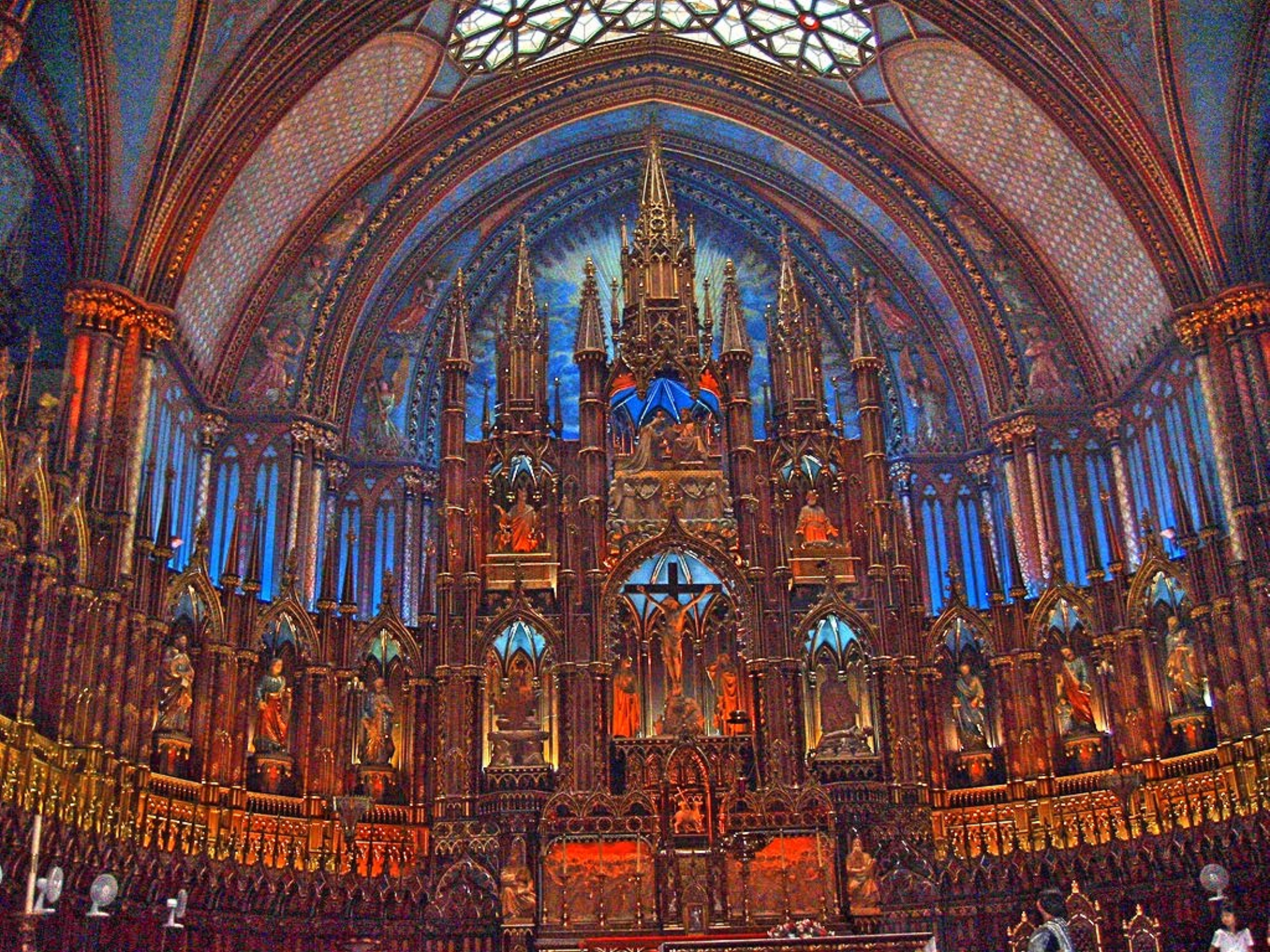}\\
		\footnotesize Input&\footnotesize JIEP&\footnotesize RRM&\footnotesize LightenNet&\footnotesize DeepUPE&\footnotesize TOLF\\
	\end{tabular}
	\caption{Illustrating visual results on challenging low-light images.}
	\label{fig:LLIE}
\end{figure*}

	\subsection{Parameter and Network Architecture Analysis}\label{sec:ParameterAnalysis}
	The proposed algorithm involves many parameters as described in Sec.~\ref{sec:IterationScheme}, some of which are iteration variables introduced based on the proximal ADMM scheme, i.e., $\lambda_{\mathbf{x}},\lambda_{\mathbf{z}},\lambda_{{\mathbf{s}_{1}}},\lambda_{\mathbf{s}_2},\lambda_{r}$. As for them, we followed the commonly-used setting (set as zero) to initialize them. Here we mainly explored the effects of some algorithmic parameters including $\tau,\zeta$, and $\beta$. As shown in Fig.~\ref{fig:Parameters}, $\tau$ and $\zeta$ were insensitive to different settings to some extent, while the parameter $\beta$ was sensitive when it increased. 
	Large $\beta$ caused poor performance. Additionally, based on these results, we defined $\tau=15$, $\zeta=1$, $\beta=0.1$ as our default settings for solving image restoration. 
	
	Table~\ref{tab:DifferentNetworks} reported quantitative scores of using different networks for PP-ADMM and our method. Among them, TNRD~\cite{chen2016trainable} is a famous image denoising framework based on a nonlinear reaction-diffusion model; DnCNN~\cite{zhang2017beyond} is a well-known CNN-based network for image denoising; IRCNN is a recently-proposed plug-and-play framework for image restoration, and here we just utilized its denoising architecture. It can be easily seen that DnCNN reached the highest scores both in PP-ADMM and our method because it owned stronger denoising ability than IRCNN and TNRD. Thus we chose DnCNN as $\mathcal{T}$ for image restoration in the experiments mentioned above and in the following experiments. Moreover, the results of our method were consistently better than PP-ADMM under different networks. It showed the superiority of our designed computational framework against the classical PP-ADMM after plugging the learnable architecture.

	\subsection{Image Processing Applications}
	\textbf{Image Restoration.}
	We compared TOLF with many approaches, including traditional optimization methods (i.e., FTVd~\cite{Li2013An}, FISTA~\cite{beck2009fast}, HL~\cite{krishnan2009fast}, IDDBM3D~\cite{danielyan2012bm3d}, EPLL~\cite{zoran2011learning}), and learning-based methods (i.e., MLP~\cite{Schuler2013A}, IRCNN~\cite{Zhang2017Learning}, MSWNNM~\cite{MSWNNM}, FDN~\cite{kruse2017learning}, GLRA~\cite{ren2018deep}, PP-ADMM~\cite{chan2016plug}, MEDAEP~\cite{li2019multi}, FIMA~\cite{liu2019convergence} (contains eFIMA and iFIMA)). As for TOLF, we introduced a denoising CNN~\cite{zhang2017beyond} architecture as $\mathcal{T}$ (according to Table~\ref{tab:DifferentNetworks}). We conducted experiments on the Levin et al.' benchmark~\cite{Levin2009Understanding}, which includes 32 images of the size 255$\times$255 and blurred by 8 different kernels of the size ranging from 13$\times$13 to 27$\times$27. We reported the quantitative scores in Table~\ref{tab:ImageRestorationQuan}. The visual comparisons on an example image from this benchmark were plotted in Fig.~\ref{fig:ImageRestoration1}. Obviously, deep-learning-based IRCNN achieved much better performance than other traditional optimization methods. 
	The recently-proposed FIMA (includes eFIMA and iFIMA) considered integrating the data and knowledge by an optimization unrolling strategy, thus its results were even better. Nevertheless, thanks to the novel modeling mechanism, TOLF obtained the best quantitative and qualitative results. 
	In addition, a color image (612$\times$342) corrupted by a very large kernel (75$\times$75) was used to further evaluate the performance, shown in Fig.~\ref{fig:ImageRestoration2}. Again, TOLF recovered richer textures and details, and thus performed the best.

	\textbf{Compressed Sensing MRI (CS-MRI).}
	We then evaluated TOLF on the CS-MRI task. {Here we defined the task-specific operation $\arg\min_{\mathbf{x}}\|\mathbf{PHF}\mathbf{x}-\mathbf{y}\|^2 + \alpha\|\mathbf{Fx}-\mathbf{y}\|^2$ with a trade-off $\alpha>0$ (fix it as $10^{-3}$) and the pre-trained denoising model described in~\cite{Zhang2017Learning} together as $\mathcal{T}$.}
	Specifically, we conducted experiments on 55 images from the widely-used IXI MRI benchmark\footnote{http://brain-development.org/ixi-dataset/.}. Our experiments contained three types of undersampling patterns (i.e., Cartesian, Gaussian, and Radial mask) and two sampling rates (i.e., 20$\%$, 30$\%$) to generate the sparse $k$-space data.
	We compared TOLF with many state-of-the-art methods including TV~\cite{lustig2008compressed}, SIDWT~\cite{baraniuk2007compressive}, PBDW~\cite{qu2012undersampled}, PANO~\cite{qu2014magnetic}, FDLCP~\cite{zhan2016fast}, ADMM-Net~\cite{yang2018admm},
	TGDOF~\cite{Liu2019TGDOF}, and BM3D-MRI~\cite{eksioglu2016decoupled}.
	Fig.~\ref{fig:CSMRIQuan} illustrated the quantitative results of these approaches on the test dataset. Our TOLF achieved the best results according to both PSNR and SSIM scores (i.e., the upper right red points). At a sampling rate of 30\%, Fig.~\ref{fig:CSMRI} demonstrated CS-MRI results on a challenging chest data~\cite{Liu2019TGDOF} with Cartesian mask. Obviously, TOLF obtained the best quantitative and qualitative performance. 
	
	\textbf{Low-Light Image Enhancement (LLIE).} 
	Lastly, we applied TOLF to solve LLIE. {Here we defined the task-specific operation as $\mathbf{x}=\mathbf{y}\oslash\mathcal{T}(\mathbf{y})$, where $\oslash$ represents the element-wise division.}
	We compared TOLF with two optimization methods (i.e., JIEP~\cite{cai2017joint} and RRM~\cite{li2018structure}) and two recent deep learning techniques (i.e., LightenNet~\cite{li2018lightennet} and DeepUPE~\cite{wang2019underexposed}). Fig.~\ref{fig:LLIE} illustrated the visual results of these methods on two challenging images. Quantitative results were reported in Fig.~\ref{fig:LLIEQuan}. By comparison, TOLF obtained the best visual quality and the highest scores.
	In Fig.~\ref{fig:AnalyzingT}, we compared the performance of TOLF with different warm-start operations $\mathcal{T}$, including the naive low-light input (denoted as $\mathbf{y}$), the result of the gamma correction (denoted as $\mathbf{y}^{{1}/{a}}$ with $a=2.2$), a simplified relative total variation filter~\cite{xu2012structure} (denoted as $\mathtt{RTV}(\mathbf{y}))$, and the simple denoising CNN architecture used in the above image restoration experiment (denoted as $\mathtt{CNN}(\mathbf{y})$). We first observed that TOLF with $\mathtt{RTV}$ and $\mathtt{CNN}$ performed better than the other two choices. On the other hand, we emphasize that even with different warm starts, the TOLF process consistently improved the overall performance.

\begin{figure}[t]
	\centering
	\begin{tabular}{c@{\extracolsep{0.9em}}c}
		\includegraphics[width=0.44\linewidth]{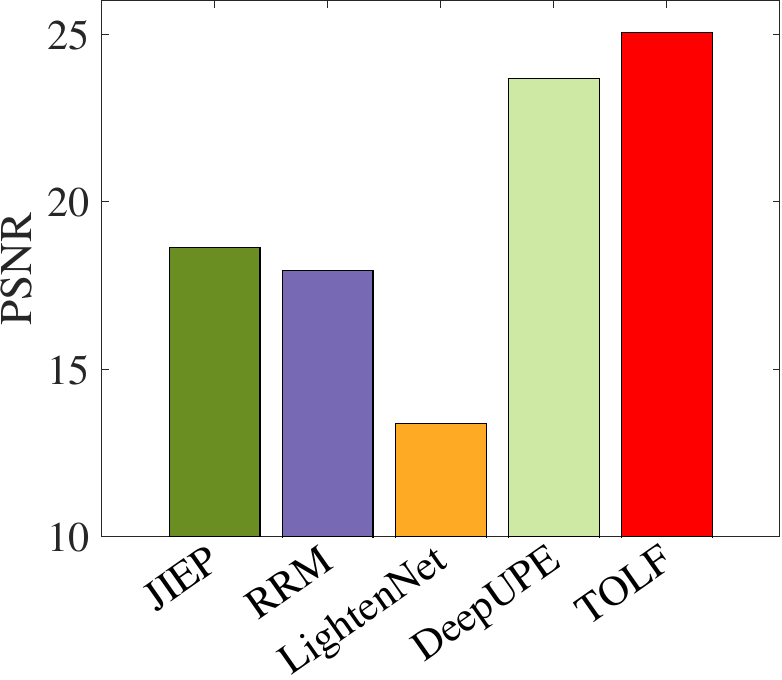}&
		\includegraphics[width=0.44\linewidth]{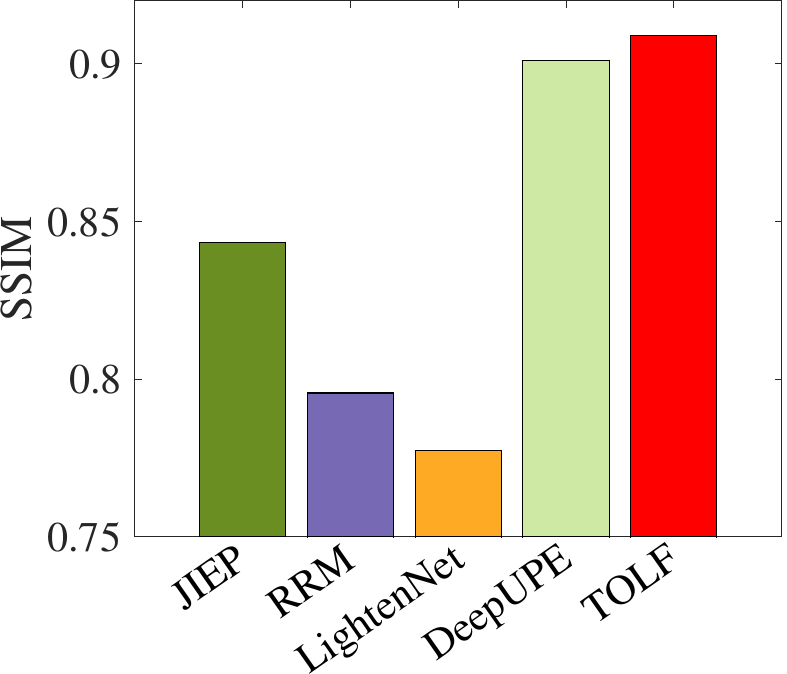}\\
	\end{tabular}
	\caption{Averaged LLIE results on 500 test images collected by~\cite{wang2019underexposed}.}
	\label{fig:LLIEQuan}
\end{figure}

	\section{Conclusion and Future Work}
	This paper developed a task-oriented convex bilevel optimization with latent feasibility for handling complex problems. The convergence and stability were strictly proved to realize our solid theoretical guarantee. Experiments on iteration behaviors verified the properties of TOLF. Extensive comparisons on three real-world applications demonstrated our outstanding performance against existing advanced methods.
	
{
Actually, our TOLF is designed towards general learning and vision models. In this work, we mainly focus on low-level vision tasks to evaluate the performance. 
In the future, we will consider extending our designed method for more challenging vision tasks, e.g., weakly supervised learning. Here we provide two possible research directions for related readers. The one is to follow the existing deep unrolling schemes to unroll the TOLF and introduce the task-specific architecture into the iteration step to further establish an end-to-end network. The other is to extend the TOLF to generate a gradient-based propagation algorithm for improving the training efficiency towards general learning issues. 
}

\appendix  

%

%
%

\begin{figure}[t]
	\centering
	\begin{tabular}{c@{\extracolsep{0.3em}}c@{\extracolsep{0.3em}}c@{\extracolsep{0.3em}}c}
		\includegraphics[width=0.23\linewidth]{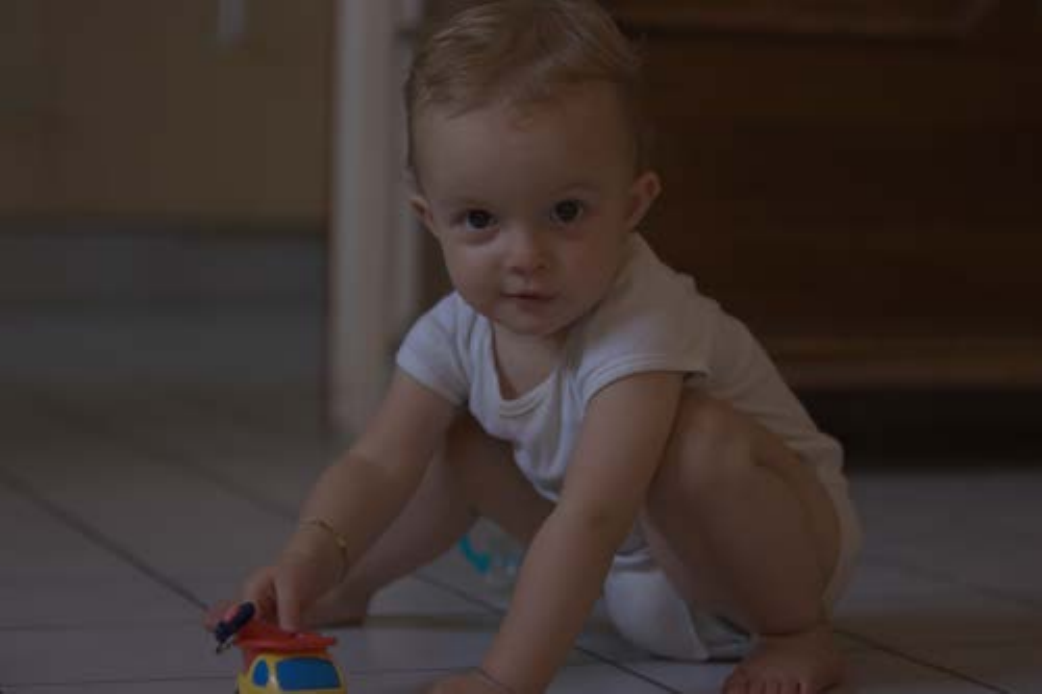}&
		\includegraphics[width=0.23\linewidth]{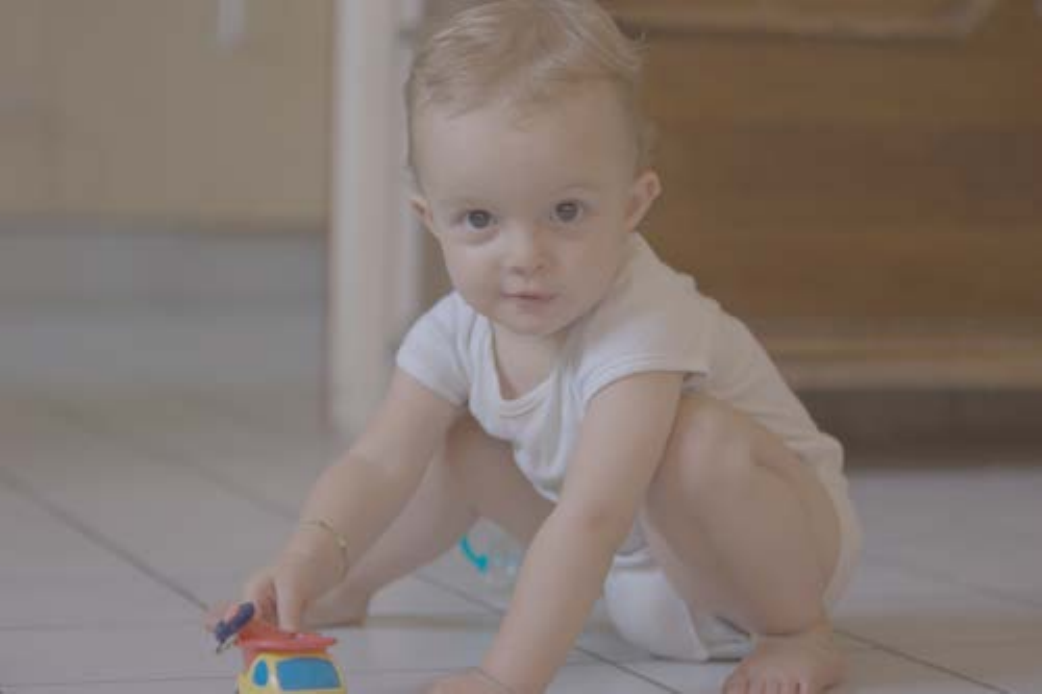}&
		\includegraphics[width=0.23\linewidth]{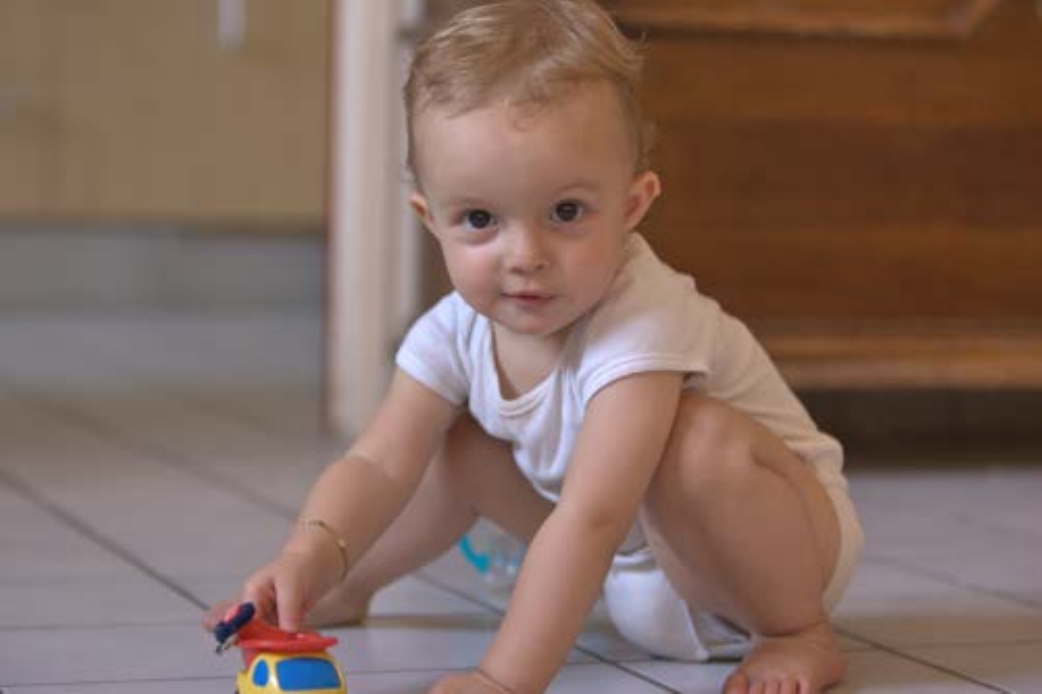}&
		\includegraphics[width=0.23\linewidth]{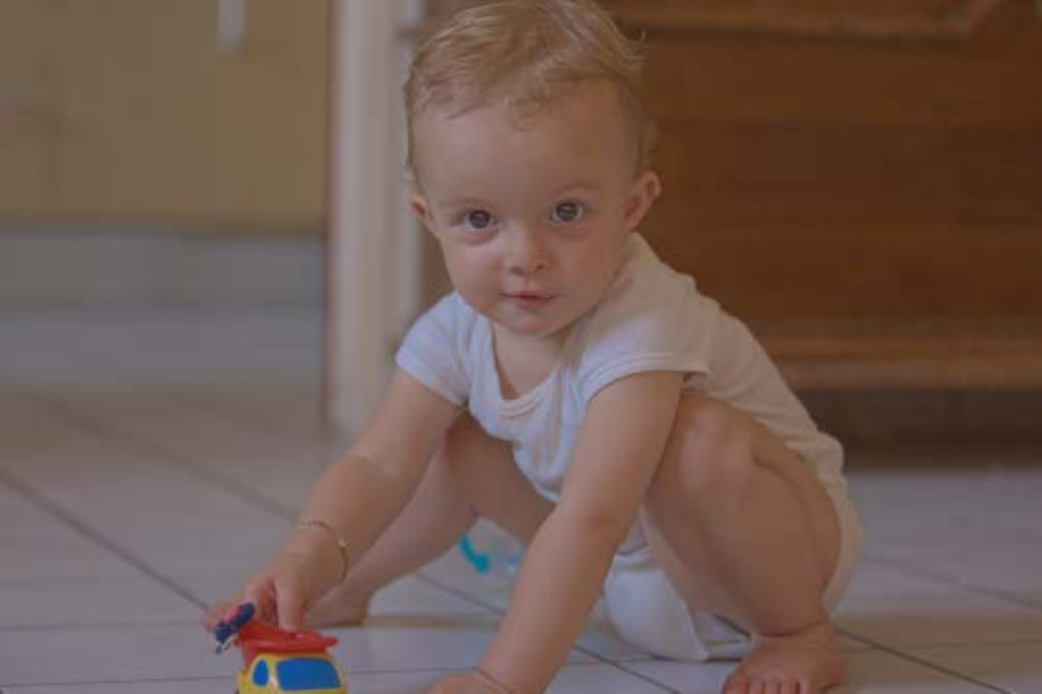}\\
		\includegraphics[width=0.23\linewidth]{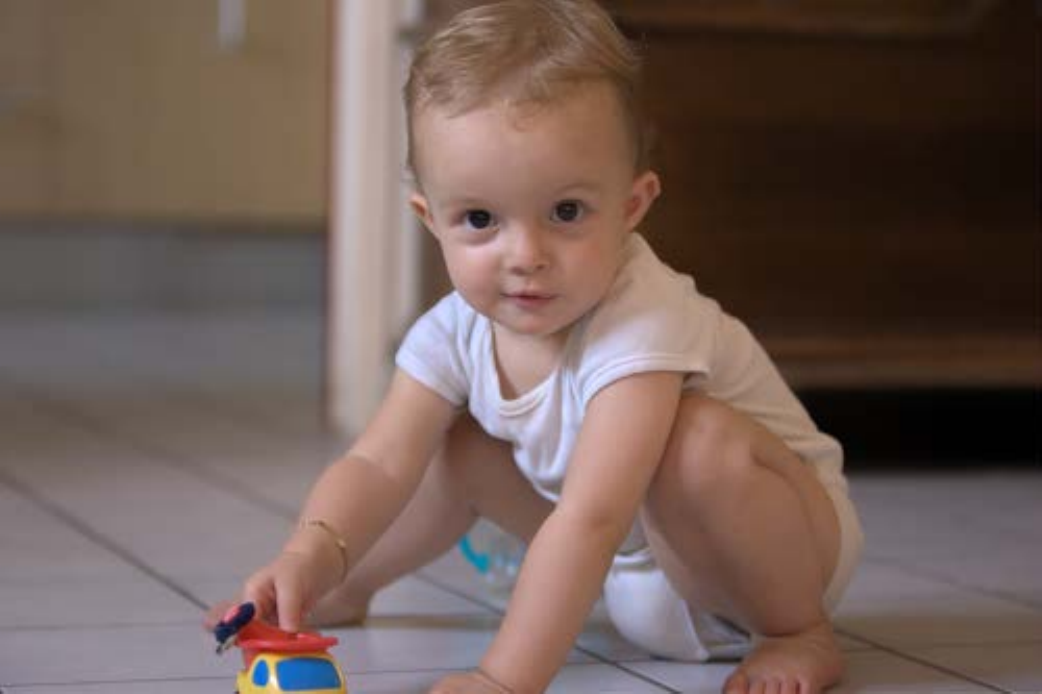}&
		\includegraphics[width=0.23\linewidth]{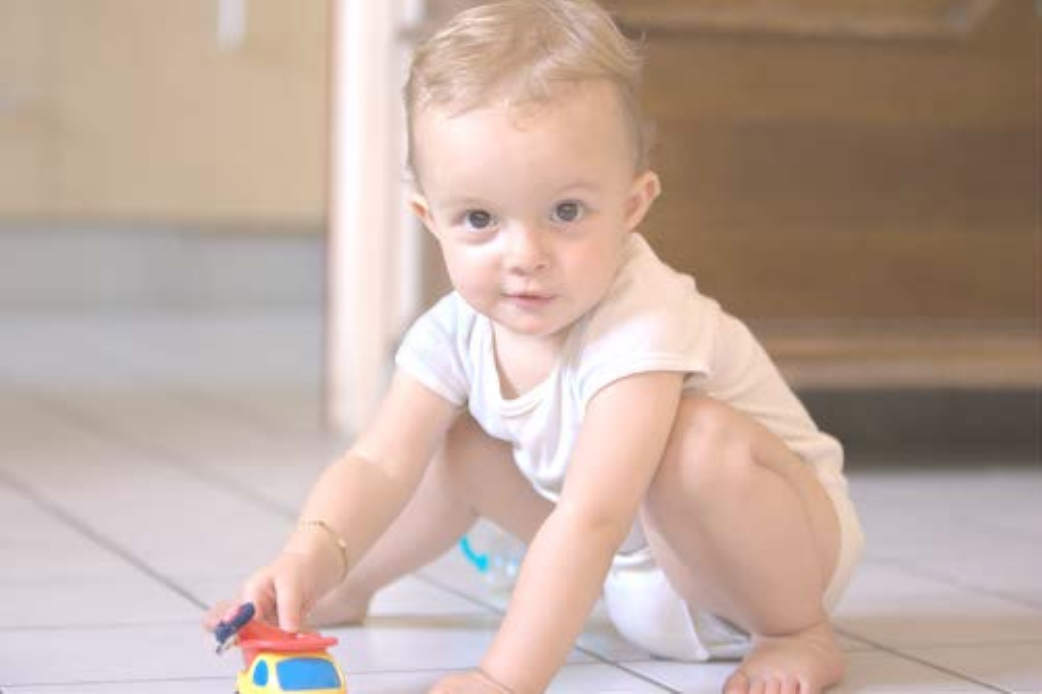}&
		\includegraphics[width=0.23\linewidth]{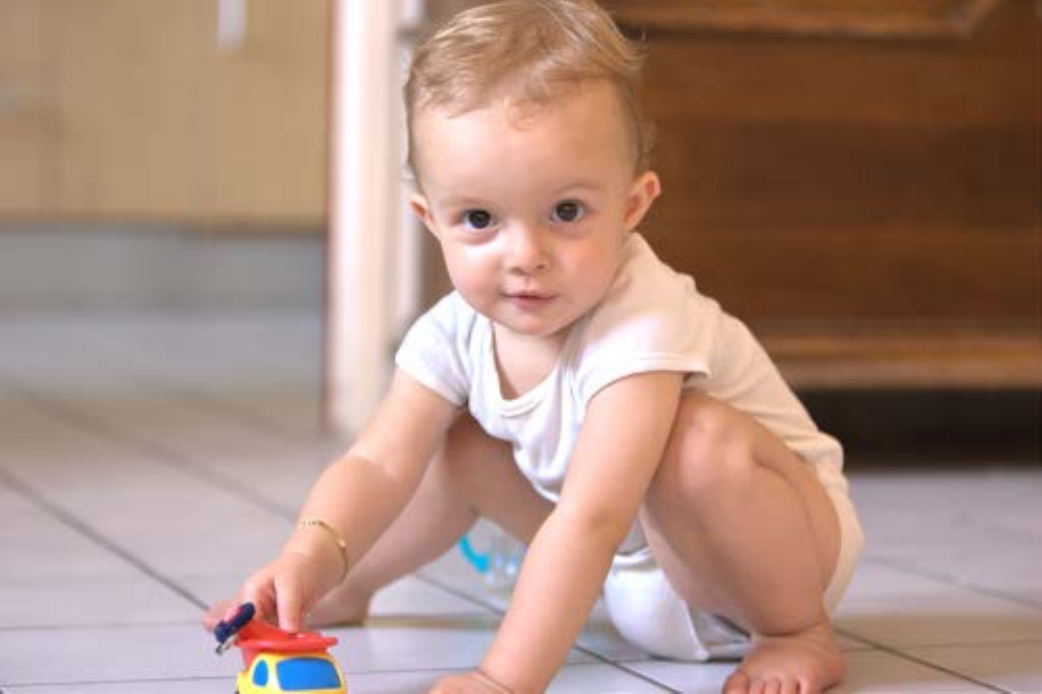}&
		\includegraphics[width=0.23\linewidth]{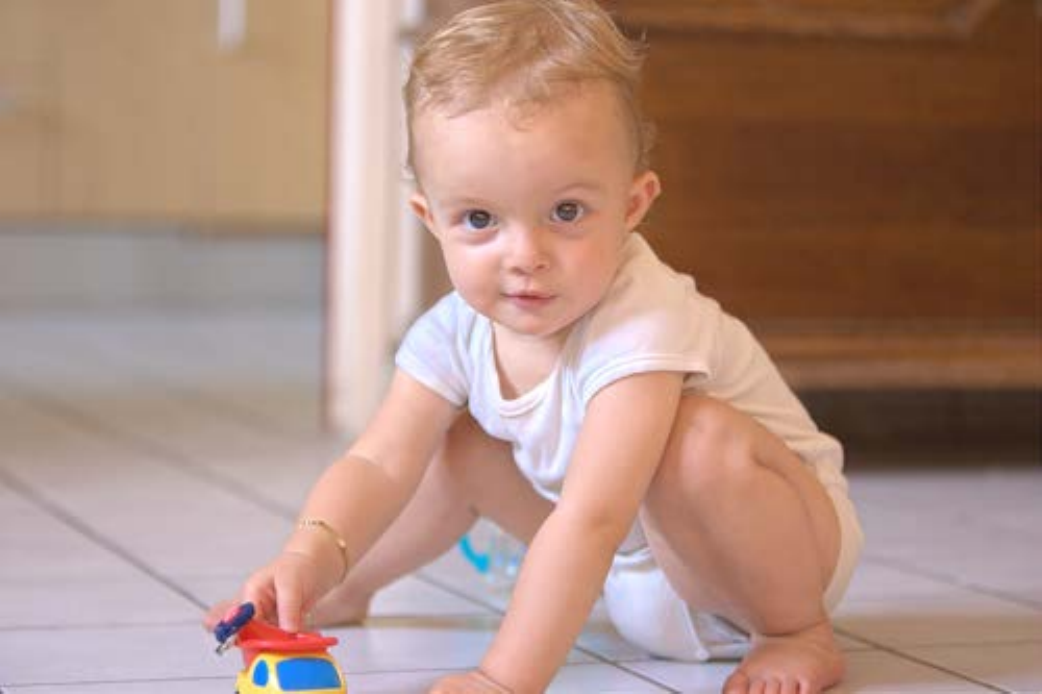}\\
		\footnotesize $\mathbf{y}$ &\footnotesize $\mathbf{y}^\frac{1}{a}$ &\footnotesize $\mathtt{RTV}(\mathbf{y})$&\footnotesize $\mathtt{CNN}(\mathbf{y})$\\
	\end{tabular}
	\caption{Illustrating the LLIE results of TOLF with different $\mathcal{T}$. Top row: results of warm start from $\mathcal{T}$. Bottom row: final enhanced results of TOLF.}
	\label{fig:AnalyzingT}
\end{figure}

In this part, we present the detailed proofs for all the theoretical results in our algorithm.

\subsection{Proof of Theorem 1}
\begin{proof}
	Given a solution $\bar{\mathbf{x}}\in \mathcal{X}$.
	First, for any $\mathbf{x} \in \left\{\mathbf{x}| \mathcal{A}(\mathbf{x})=\mathcal{A}(\bar{\mathbf{x}}), \varphi(\mathbf{x}) \leq \varphi(\bar{\mathbf{x}}) \right\}$, we have $\Psi(\mathbf{x}) = h(\mathcal{A}(\bar{\mathbf{x}})) + \varphi(\mathbf{x}) \le h(\mathcal{A}(\bar{\mathbf{x}})) + \varphi(\bar{\mathbf{x}}) = \min_{\mathbf{x}} \Psi(\mathbf{x})$, and thus
	\begin{equation}\label{inclusion1}
	\left\{\mathbf{x}| \mathcal{A}(\mathbf{x})=\mathcal{A}(\bar{\mathbf{x}}), \varphi(\mathbf{x}) \leq \varphi(\bar{\mathbf{x}}) \right\} \subseteq \mathcal{X}.
	\end{equation}
	For any $\mathbf{x} \in \mathcal{X}$, if $\mathcal{A}(\mathbf{x}) \neq \mathcal{A}(\bar{\mathbf{x}})$, let $\mathbf{x}_\alpha = (1-\alpha)\bar{\mathbf{x}} + \alpha\mathbf{x}$, and then $\mathbf{x}_\alpha \in \mathcal{X}$ because $\mathcal{X}$ is convex. 
	As $h$ is locally strongly convex around $\bar{x}$, there exists neighborhood $\mathcal{N}$ of $\mathcal{A}(\bar{\mathbf{x}})$ such that $h$ is strongly convex on $\mathcal{N}$.
	There exists sufficiently small $\alpha > 0$ such that $\mathcal{A}(\mathbf{x}_\alpha) \in \mathcal{N}$ and $\mathcal{A}(\mathbf{x}_\alpha) \neq \mathcal{A}(\bar{\mathbf{x}})$ . Then, there exists $\sigma > 0$ such that
 	\begin{equation*}
			\begin{aligned}
			h(\mathcal{A}(\mathbf{x}_\alpha)) \ge \, &h(\mathcal{A}(\bar{\mathbf{x}})) + \alpha\langle \mathbf{D} h(\mathcal{A}(\bar{\mathbf{x}})), \mathcal{A}(\mathbf{x})-\mathcal{A}(\bar{\mathbf{x}}) \rangle \\ & + \frac{\sigma}{2}\alpha^2\|\mathcal{A}(\mathbf{x})-\mathcal{A}(\bar{\mathbf{x}})\|^2.
			\end{aligned}
	\end{equation*} 
	And since $0 \in \mathcal{A}^T \mathbf{D} h(\mathcal{A}(\bar{\mathbf{x}})) + \partial \varphi(\bar{\mathbf{x}})$, by the convexity of $\varphi$, we have
	\[
	\varphi(\mathbf{x}_\alpha) \ge \varphi(\bar{\mathbf{x}}) + \alpha\langle - \mathcal{A}^T \mathbf{D} h(\mathcal{A}(\bar{\mathbf{x}})), \mathbf{x}-\bar{\mathbf{x}}\rangle.
	\]
	Combining the two inequalities given above, we obtain
	\[
	\Psi(\mathbf{x}_\alpha) \ge \Psi(\bar{\mathbf{x}}) + \frac{\sigma}{2}\alpha^2\|\mathcal{A}(\mathbf{x})-\mathcal{A}(\bar{\mathbf{x}})\|^2 > \Psi(\bar{\mathbf{x}}),
	\]
	which contradicts to the fact that $\mathbf{x}_\alpha \in \mathcal{X}$. Next, since $\mathcal{A}(\mathbf{x}) = \mathcal{A}(\bar{\mathbf{x}})$, and $\Psi(\mathbf{x}) = \Psi(\bar{\mathbf{x}})$, we have $\varphi(\mathbf{x}) = \varphi(\bar{\mathbf{x}})$, and thus
	\begin{equation}\label{inclusion2}
	\mathcal{X} \subseteq \left\{\mathbf{x}| \mathcal{A}(\mathbf{x})=\mathcal{A}(\bar{\mathbf{x}}), \varphi(\mathbf{x}) \leq \varphi(\bar{\mathbf{x}}) \right\}.
	\end{equation}
	Upon combining Eq.~\eqref{inclusion1} and Eq.~\eqref{inclusion2}, we reach the re-characterization of $\mathcal{X}$ as Eq.~\eqref{rechare}.
\end{proof}

\subsection{Proof of Corollary 1}
\begin{proof}
	By Theorem~\ref{pro:solution}, we directly have the equivalence between Eq.~\eqref{eq:bp} and Eq.~\eqref{eq:bp-linear}. And since Eq.~\eqref{eq:admm} is the iterations of two block proximal ADMM applied on Eq.~\eqref{eq:bp-linear}. Thus, the convergence of the iterations in solving problem Eq.~\eqref{eq:bp-linear} can be directly guaranteed via standard convergence results of proximal ADMM \cite{he2002new}. Moreover, by applying the investigations in \cite{ADMMlinear}, we can further obtain a linear convergence rate estimation for our iterations in solving problem Eq.~\eqref{eq:bp-linear}.	
\end{proof}

\subsection{Proof of Lemma 1}
\begin{proof}
	First, we show that $\limsup_{\mathbf{p} \rightarrow 0}\mathcal{S}_{val}(\mathbf{p}) \le \mathcal{S}_{val}(0)$. Let $\bar{\mathbf{x}} \in \mathcal{S}_{sol}(0)$. Since $\mathcal{S}_{feas}(\mathbf{p})$ is continuous at $0$, it is inner semicontinuous at $0$. Thus for any sequence $\mathbf{p}^k \rightarrow 0$, we get the existence of a sequence of points $\mathbf{x}^k \in \mathcal{S}_{feas}(\mathbf{p}^k)$ such that $\mathbf{x}^k \rightarrow \bar{\mathbf{x}}$ as $k \rightarrow \infty$. Then for any $\epsilon >0$ there exists $N > 0$ such that
	\[
	\mathcal{S}_{val}(\mathbf{p}^k) \le F(\mathbf{x}^k) \le F(\bar{\mathbf{x}}) + \epsilon = \mathcal{S}_{val}(0) + \epsilon, \qquad \forall k \ge N,
	\]
	which implies
	\[
	\limsup_{\mathbf{p} \rightarrow 0}\mathcal{S}_{val}(\mathbf{p}) \le \mathcal{S}_{val}(0).
	\]
	We next show the outer semicontinuity of $\mathcal{S}_{sol}$ at $0$. For any $\mathbf{p}^k \rightarrow 0$ with $\mathbf{x}^k \in \mathcal{S}_{opt}(\mathbf{p}^k)$ such that $\mathbf{x}^k \rightarrow \bar{\mathbf{x}}$, since $\mathcal{S}_{feas}$ is outer semicontinuous at $0$, we have $\bar{\mathbf{x}} \in \mathcal{S}_{feas}(0)$. By the continuity of $F$ and upper semicontinuity of $\mathcal{S}_{val}$ at $0$, we have
	{\small	\[
		\mathcal{S}_{val}(0) \le F(\bar{\mathbf{x}}) = \lim_{k \rightarrow \infty}F(\mathbf{x}^k) = \limsup_{k \rightarrow \infty} \mathcal{S}_{val}(\mathbf{p}^k) \le \mathcal{S}_{val}(0),
		\]}
	which implies
	\[
	\bar{\mathbf{x}} \in \mathcal{S}_{sol}(0).
	\]
	That is, $\mathcal{S}_{sol}(\mathbf{p})$ is outer semicontinuous at ${0}$ according to Definition 1 in the manuscript.
\end{proof}

\subsection{Proof of Theorem 2}
\begin{proof}
	For any $\mathbf{x}^*_{\delta}$, we have $\mathbf{x}^*_{\delta} \in \mathcal{S}_{sol}(\mathbf{p})$ with $\mathbf{p}_1 = \mathcal{A}(\bar{\mathbf{x}}_{\delta}) - \mathcal{A}(\bar{\mathbf{x}})$ and $\mathbf{p}_2 = \varphi(\bar{\mathbf{x}}_{\delta}) - \varphi(\bar{\mathbf{x}})$. Therefore, $\|\mathbf{p}\| \le \|\mathcal{A}(\bar{\mathbf{x}}_{\delta}) - \mathcal{A}(\bar{\mathbf{x}})\| + \|\varphi(\bar{\mathbf{x}}_{\delta}) - \varphi(\bar{\mathbf{x}})\| \le (\|\mathcal{A}\| + L_\varphi)d(\bar{\mathbf{x}}_{\delta},\mathcal{X}) \le (\|\mathcal{A}\| + L_\varphi)\delta$, where $L_\varphi$ is the Lipschitz continuity modulus of $\varphi$. Note that the Lipschitz continuity modulus is guaranteed to exist because $\varphi$ is a convex polyhedral function. Then the first argument follows from Lemma \ref{prop2} directly. The second argument actually follows from the fact that $\limsup_{k \rightarrow \infty} F(\mathbf{x}^*_{\delta_k}) \le \mathcal{S}_{val}(0)$ and $F$ is coercive.	
\end{proof}

\subsection{Proof of Proposition 1}
\begin{proof}
	For any $\mathbf{x}_p \in \mathcal{S}_{opt}(\mathbf{p}) \cap \mathcal{N}$, let $\mathbf{z} := \mathtt{Proj}_{\mathcal{S}_{feas}(0)}(\mathbf{x}_p)$ and $\mathbf{x}_0 := \mathtt{Proj}_{\mathcal{S}_{sol}(0)}(\mathbf{z})$, and since $\mathcal{S}_{feas}(0)$ and $\mathcal{S}_{sol}(0)$ are both closed convex sets, $\mathbf{z}$ and $\mathbf{x}_0$ are well defined. Because $\bar{\mathbf{x}} \in \mathcal{S}_{sol}(0)$, we have $\|\mathbf{x}_0 - \bar{\mathbf{x}} \| \le \|\mathbf{z} - \bar{\mathbf{x}} \| \le \|\mathbf{x}_p - \bar{\mathbf{x}}\|$, and thus $\mathbf{z},\mathbf{x}_0 \in \mathcal{N}$ and $\|\mathbf{x}_p - \mathbf{z}\| \le \kappa_1\|\mathbf{p}\|$.
	
	Since $\mathbf{x}_p \in \mathcal{S}_{sol}(\mathbf{p})$, for any point $\mathbf{y} \in \mathcal{S}_{feas}(\mathbf{p})$, we have
	\[
	F(\mathbf{x}_p) - F(\mathbf{x}_0) = F(\mathbf{x}_p) - F(\mathbf{y}) + F(\mathbf{y}) - F(\mathbf{x}_0) \le L\| \mathbf{y}- \mathbf{x}_0\| .
	\]
	Since $\mathbf{y}$ can be any point in $\mathcal{S}_{feas}(\mathbf{p})$, we have
	\begin{equation}\label{prop3_eq1}
	F(\mathbf{x}_p) - F(\mathbf{x}_0)\le \kappa_2 L\|\mathbf{p}\|.
	\end{equation}
	Next, we have
	\[
	\begin{aligned}
	F(\mathbf{x}_p) - F(\mathbf{x}_0) &\ge F(\mathbf{z}) - F(\mathbf{x}_0) - |F(\mathbf{x}_p) - F(\mathbf{z})| \\
	&\ge c \|\mathbf{z} - \mathbf{x}_0\|^2 - L\|\mathbf{x}_p -\mathbf{z}\| \\
	& \ge c (\|\mathbf{x}_p-\mathbf{x}_0\| - \|\mathbf{z}-\mathbf{x}_p\|)^2 - \kappa_1L\|\mathbf{p}\| \\
	& \ge c(\|\mathbf{x}_p-\mathbf{x}_0\| - \kappa_1\|\mathbf{p}\|)^2 - \kappa_1L\|\mathbf{p}\|.
	\end{aligned}
	\]
	Combining with Eq.~\eqref{prop3_eq1}, we get
	\[
	\kappa_2 L\|\mathbf{p}\| \ge c(\|\mathbf{x}_p-\mathbf{x}_0\| - \kappa_1\|\mathbf{p}\|)^2 - \kappa_1L\|\mathbf{p}\|,
	\]
	and thus
	\[
	d(\mathbf{x}_p, \mathcal{S}_{sol}(0)) \le \|\mathbf{x}_p - \mathbf{x}_0\| \le \kappa_1\|\mathbf{p}\| + \sqrt{\frac{(\kappa_1+\kappa_2)L}{c}\|\mathbf{p}\|}.
	\]
\end{proof}

\subsection{Proof of Theorem~3}
\begin{proof}
	As stated in our manuscript, according to the results proved in Proposition~\ref{prop3}, together with the arguments given in the proof of Theorem \ref{th1}, we can directly have the stability guarantees in this theorem.
\end{proof}



\ifCLASSOPTIONcaptionsoff
  \newpage
\fi



%
\bibliographystyle{IEEEtran}
\bibliography{references}
	
\begin{IEEEbiography}[{\includegraphics[width=1in,height=1.25in,clip,keepaspectratio]{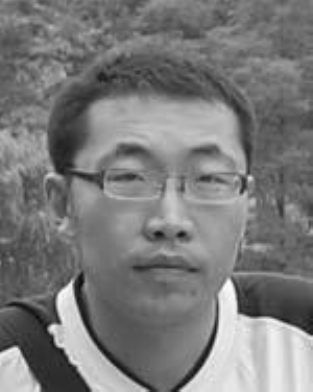}}]{Risheng Liu} (M'12-) received the BSc and PhD degrees both in mathematics from the Dalian University of Technology in 2007 and 2012, respectively. He was a visiting scholar in the Robotic Institute of Carnegie Mellon University from 2010 to 2012. He served as Hong Kong Scholar Research Fellow at the Hong Kong Polytechnic University from 2016 to 2017. He is currently a professor with the International School of Information Science \& Engineering, Dalian University of Technology. His research interests include machine learning, optimization, computer vision and multimedia. He was a co-recipient of the IEEE ICME Best Student Paper Award in both 2014 and 2015. Two papers were also selected as Finalist of the Best Paper Award in ICME 2017. He is a member of the IEEE and ACM.		
\end{IEEEbiography}

\begin{IEEEbiography}[{\includegraphics[width=1in,height=1.25in,clip,keepaspectratio]{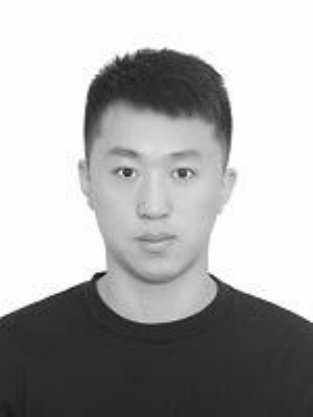}}]{Long Ma} received the M.S. degree in software engineering at Dalian University of Technology, Dalian, China, in 2019. He is currently pursuing the Ph. D. degree in software engineering at Dalian University of Technology, Dalian, China. His research interests include computer vision, image enhancement and machine learning. He is a reviewer for CVPR, ICCV, AAAI, ACCV,  IEEE TCSVT, and Neurocomputing.
\end{IEEEbiography}

\begin{IEEEbiography}[{\includegraphics[width=1in,height=1.25in,clip,keepaspectratio]{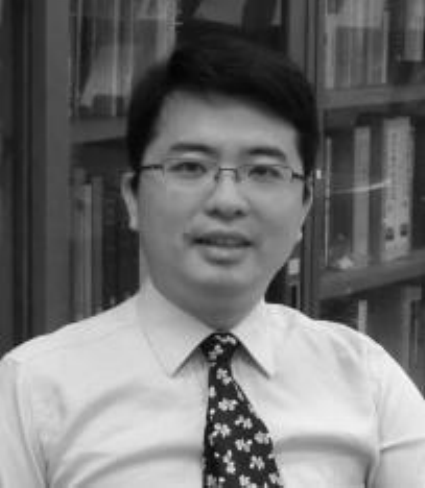}}]{Xiaoming Yuan} is Professor at Department of Mathematics, The University of Hong Kong. His main research interests include numerical optimization, scientific computing and optimal control. Recently, he is particularly interested in optimization problems in various AI and cloud computing areas.
\end{IEEEbiography}

\begin{IEEEbiography}[{\includegraphics[width=1in,height=1.25in,clip,keepaspectratio]{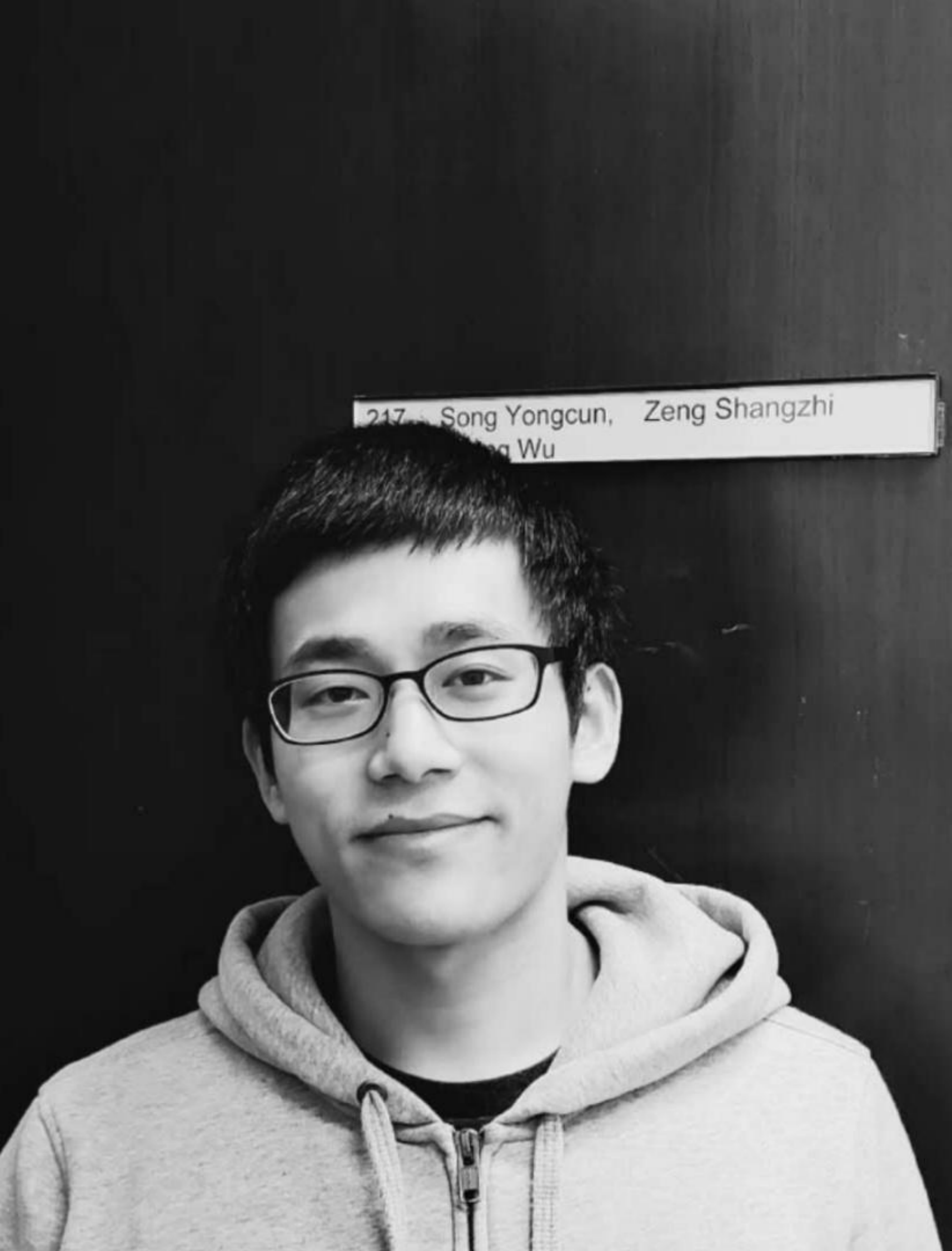}}]{Shangzhi Zeng} received the B.Sc. degree in Mathematics and Applied Mathematics from Wuhan University in 2015, the M.Phil. degree from Hong Kong Baptist University in 2017, and the Ph.D. degree from the University of Hong Kong in 2021. He is currently a PIMS postdoctoral fellow in the Department of Mathematics and Statistics at University of Victoria. His current research interests include variational analysis and bilevel optimization.
\end{IEEEbiography}

\begin{IEEEbiography}[{\includegraphics[width=1in,height=1.25in,clip,keepaspectratio]{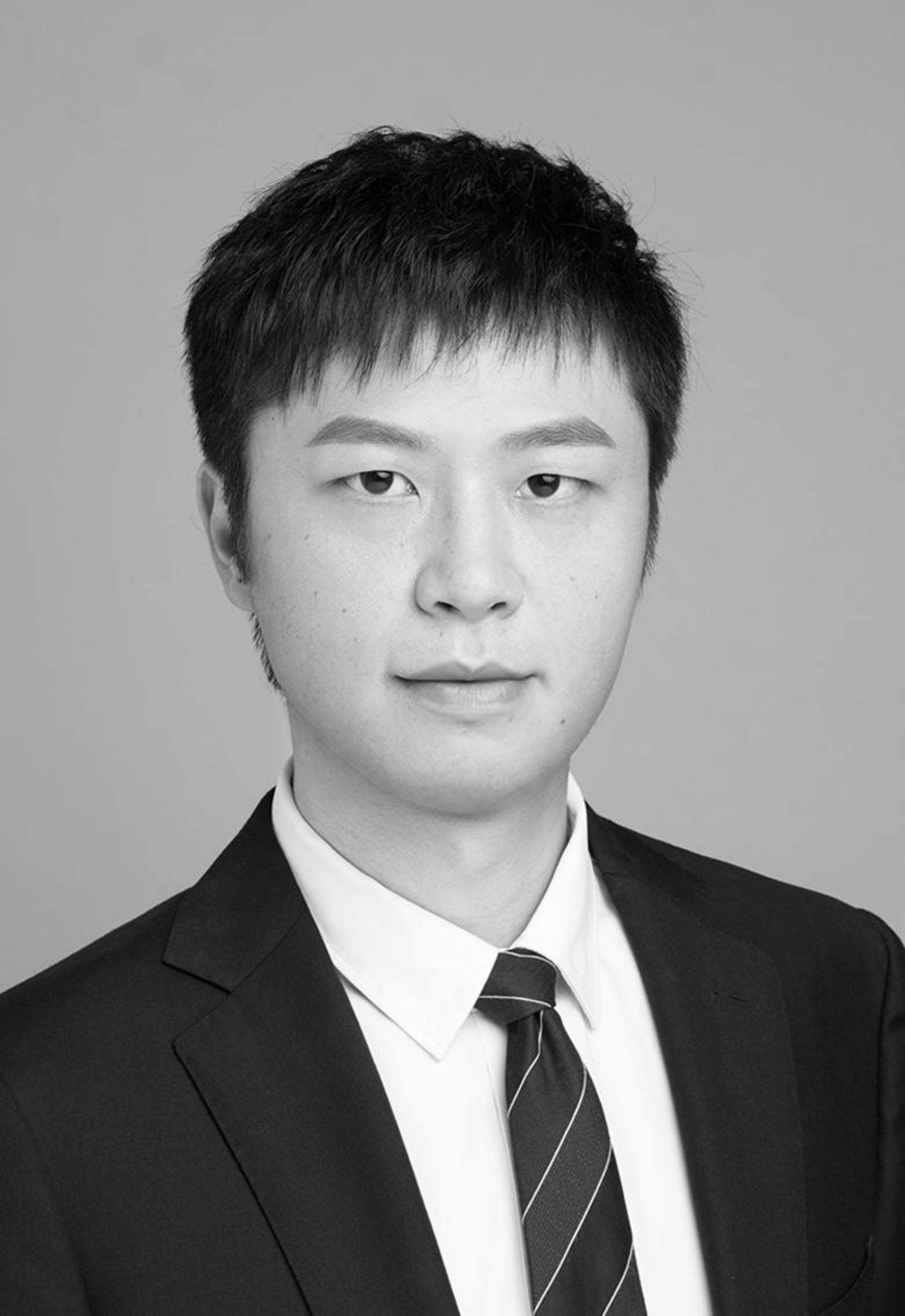}}]{Jin Zhang} received the B.A. degree in Journalism from the Dalian University of Technology in 2007. He pursued a degree in mathematics and received the M.S. degree in Operational Research and Cybernetics from the Dalian University of Technology, China, in 2010, and the Ph. D. degree in Applied Mathematics from University of Victoria, Canada, in 2015. After working in Hong Kong Baptist University for 3 years, he joined Southern University of Science and Technology as a tenure-track assistant professor in the department of mathematics. His broad research area is comprised of optimization, variational analysis and their applications in economics, engineering and data science.
\end{IEEEbiography}

%




\end{document}